\documentclass{article}

\usepackage{amsmath,amssymb,amsfonts}
\usepackage{graphicx}
\usepackage{booktabs}
\usepackage{multirow}
\usepackage{array}
\usepackage{subcaption}
\usepackage{hyperref}
\usepackage{makecell}
\usepackage{xcolor}
\usepackage{float}

\DeclareMathOperator*{\argmax}{arg\,max}

\usepackage[margin=1in]{geometry}
\title{Learning Coverage- and Power-Optimal Transmitter Placement from \textcolor{black}{City Maps}: A Comparative Study of Direct and Indirect Neural Approaches}

\author{\c{C}a{\u{g}}kan Yapar}

\begin{document}

\maketitle

\begin{abstract}
Optimal wireless transmitter placement is a central task in radio-network planning, yet exhaustive search becomes prohibitively expensive at scale. This paper studies the single-transmitter setting under a fixed learned propagation model, enabling exhaustive per-pixel assessment at dataset scale in a regime where measurement-based exhaustive labeling is infeasible and ray-tracing-based exhaustive labeling is computationally out of reach.
We introduce a dataset of 167{,}525 urban scenarios (\emph{RadioMapSeer-}\emph{Deployment}) with dual ground-truth labels for coverage-optimal and power-optimal transmitter locations. Benchmark analysis reveals an asymmetric coverage--power trade-off: coverage-optimal placement sacrifices $13.86\%$ of received-power, whereas power-optimal placement sacrifices only $5.50\%$ of coverage; the best achievable balanced placement lies at $\bar{d}=2.60$ from the ideal point $(100\%,100\%)$.
We evaluate two learning formulations: indirect heatmap-based models predicting received-power radio maps, and direct score-map models predicting the objective landscape over feasible transmitter locations. Within the heatmap family, discriminative models deliver one-shot predictions $1350$--$2400\times$ faster than exhaustive search, while diffusion models additionally support multi-sample inference that improves single-objective performance and, by reusing the same sample pool under a balanced criterion, recovers strong balanced placements without explicit multi-objective training. Dual score-map strategies that combine power and coverage score-maps match the exhaustive balanced optimum ($\bar{d}=2.60$) and remain close to it across smaller candidate budgets, at $14$--$22\times$ speedups including the cost of evaluating shortlisted candidates.
Both formulations admit very fast one-shot inference. Dual score-map methods are strongest overall, whereas heatmap formulations remain attractive for their physically meaningful intermediate maps and, in the diffusion setting, for inference-time search.
\end{abstract}

\section{Introduction}

Choosing where to place a wireless transmitter is a foundational task in radio-network planning. Two natural deployment objectives are to maximize the average received-power over the region of interest and to maximize the fraction of that region covered above a required signal threshold. Under either objective, identifying a good location requires evaluating many candidate placements, each coupled to a propagation prediction over the environment. At realistic spatial resolution, this quickly becomes prohibitively expensive when repeated across many candidate locations and many environments.

Learning-based methods offer a way around this cost. A neural network trained on solved instances can produce either a transmitter placement directly or a prediction structure from which one can be recovered in a single forward pass. In the recent literature, two broad formulations have emerged. The \emph{indirect} route predicts a radio map and then recovers a transmitter location from that map. The \emph{direct} route predicts a scalar objective value at each feasible candidate pixel and selects the best-scoring one. Both formulations have been studied, but a controlled comparison under a shared dataset, shared architectures, and shared evaluation protocol has been largely missing. 

This paper provides such a comparison. \textcolor{black}{We focus on the single-transmitter case in outdoor urban environments. The single-transmitter case is the fundamental case of the placement problem and the only regime in which the direct (score-map) approach studied here applies straightforwardly: the score-map is a 2D image of per-pixel quality scores --- a natural CNN output target. The natural multi-transmitter analog is a tensor of joint-placement scores indexed by $n$-tuples of pixel coordinates, with approximately $150^{2n}$ entries per environment at the feasible-region size used here ($\sim$$5\times10^8$ for $n=2$, $\sim$$10^{13}$ for $n=3$); such tensors are not natural outputs of standard CNNs and would require $\sim$$340$~TB at FP32 to store at $n=2$ dataset scale. Exhaustive learned propagation model evaluation in the single-transmitter case defines benchmark ground truth: for each scenario, the globally optimal placement under each objective is known, and the best achievable balanced placement is also known. This makes it possible to compare different learning formulations against benchmark optima rather than only against one another. Two-transmitter exhaustive labeling, under the same pipeline, costs approximately $25\times$ the per-environment compute of single-transmitter labeling --- in reach in principle but very demanding; three-transmitter (and higher) exhaustive labeling is out of reach with current computational budgets.}

Concretely, we release \emph{RadioMapSeer-Deployment}, a dataset of 167,525 urban scenarios with dual ground-truth labels for coverage-optimal and power-optimal transmitter locations, derived by exhaustive per-pixel evaluation under a fixed learned propagation model. We train and evaluate two matched families of models on this benchmark. The heatmap family predicts received-power radio maps and then recovers the transmitter location from them; within this family, deterministic models produce single-shot predictions, while a diffusion model additionally supports multi-sample inference, in which different noise realizations yield different candidate maps and the best can be selected under a chosen criterion. The score-map family directly predicts the objective landscape for power or coverage over feasible transmitter locations and extends naturally to dual-score-map strategies that combine both objectives for balanced placement.

Across power-only, coverage-only, and balanced evaluation, these two families occupy complementary parts of the accuracy--cost space. Score-map formulations provide the strongest balanced performance on this benchmark; both discriminative heatmap and discriminative score-map models support equally fast one-shot inference; and diffusion-based multi-sample inference contributes a distinct search-over-samples mechanism that improves placement quality at higher computational cost. \textcolor{black}{Two takeaways follow: practically, direct (score-map) approaches achieve the strongest performance on this benchmark, and for balanced placement in particular dual-score-map union selection matches the exhaustive-search optimum; conceptually, both formulation families offer inference-time multi-candidate evaluation as a runtime/accuracy trade-off, top-$K$ candidate shortlisting from a predicted score-map in the score-map family and best-of-$N$ selection across stochastic samples in the diffusion family.} The remainder of the paper develops these findings quantitatively and analyzes what they imply about the underlying geometry of the transmitter-placement trade-off.''

\subsection{Related Work}
\label{sec:relatedwork}

\subsubsection{Classical and Heuristic Transmitter Placement}

Wireless transmitter and base-station placement has traditionally been studied as combinatorial search over a set of candidate locations, with each candidate evaluated via an empirical or deterministic propagation model, typically a ray tracing or dominant-path simulation. Early WLAN planning frameworks combined simulated annealing, genetic algorithms, and pattern search to jointly optimize access-point placement and channel assignment over coverage, capacity, and received-power objectives~\cite{stosic_wlan}. More recent work has focused both on stronger optimization procedures and on embedding candidate evaluation in more realistic site-specific propagation models: Chen~et~al.\ propose an improved Hooke--Jeeves method for indoor mmWave base-station deployment, using ray tracing-derived pathloss and SINR estimates directly inside the optimization loop, and show that better initialization significantly reduces convergence time~\cite{chen_hookjeeves}. Palizban~et~al.\ formulate outdoor mmWave BS planning as a binary integer linear program (BILP) combined with greedy refinement to maximize coverage~\cite{palizban_mmwave}. Li~et~al.\ propose a quantum-inspired genetic algorithm (DAQGA) for 5G BS deployment in a dense urban area, using 3D ray tracing to score coverage and cost~\cite{li_daqga}. Recent work has also combined high-fidelity Sionna ray tracing with interference-aware submodular greedy optimization for sparse multi-transmitter placement, supporting exclusion zones and incremental densification in realistic 3D urban environments while providing theoretical approximation guarantees~\cite{taus2026optimal}. Across these approaches, the practical bottleneck remains the cost of candidate evaluation under explicit propagation modeling, which limits the number of configurations that can be explored in practice.

A separate line of work formulates deployment as an explicit multi-objective problem. Isabona~et~al.\ optimize coverage, capacity, and power consumption jointly for 4G LTE base stations using NSGA-II, producing a set of Pareto-optimal solutions rather than a single deterministic outcome~\cite{isabona_nsga2}. Their formulation is the closest antecedent in the classical literature to the balanced coverage--power analysis of the present paper.

\subsubsection{Learned Radio Propagation as a Planning Enabler}

The idea of replacing classical propagation solvers with neural surrogates gained traction as deep convolutional architectures showed that spatial radio maps can be predicted directly from geometry and transmitter inputs at inference speeds far exceeding ray tracing. Foundational contributions include RadioUNet~\cite{radioUNet}, which formulated outdoor received-power radio map prediction as an image-to-image translation problem and simultaneously introduced the initial version of the RadioMapSeer dataset~\cite{radiomapseer}---\textcolor{black}{city map}- and Tx-location-conditioned radio maps generated with the Dominant Path Model~\cite{DPM} and Intelligent Ray Tracing (IRT)~\cite{IRT} on a shared set of urban \textcolor{black}{city maps}, with the propagation simulations performed using the WinProp software from Altair~\cite{WinPropFEKO}. Concurrently, Zhang~et~al.\ introduced PLNet~\cite{zhang_plnet}, framing pathloss prediction as an image-regression problem solved with a deep convolutional network trained on both simulated and real field data, and reporting significantly lower RMSE than both empirical (SPM) and deterministic (commercial ray tracer) baselines when evaluated against field measurements. FadeNet~\cite{fadenet} extended this line to millimeter-wave large-scale channel fading, reporting approximately 5.6~dB RMSE together with 40--1000$\times$ speedups over ray tracing and demonstrating applicability to transmitter-site selection. EM~DeepRay~\cite{emdeepray} trained a convolutional encoder-decoder to reproduce indoor ray tracer outputs across multiple geometries and frequency bands, with predictions in milliseconds and transfer-learning-based calibration toward real measurements. Jaensch~et~al.\ extend the learned-surrogate idea to beam-index prediction: a UNet trained on Wireless InSite ray tracing labels across 609 Berlin city environments predicts spatial maps of optimal beamforming indices from geospatial data alone, with a single forward pass serving the entire coverage area and achieving a 32$\times$ reduction in beam-training overhead at near-optimal throughput~\cite{jaensch_beam}. Surveys of this body of work document both the breadth of deep-learning propagation modeling techniques and their downstream applications in network planning, localization, and RIS optimization~\cite{romeroSurvey,ZengSurvey,challengeoverview,mdpireview,empowering,wangtutorial}.

Recent large-scale work by Bakirtzis~et~al.~\cite{bakirtzis_deeplearn} compares DL surrogates with differentiable ray tracing on a commercial MNO dataset spanning 13 cities, more than 10{,}000 antennas, and approximately 300~million crowdsourced RSRP measurements. The study reports that DL surrogates achieve accuracy comparable to the differentiable ray tracer Sionna~\cite{sionna_rt} when both are trained on matched synthetic data, and a clear advantage when the DL model is trained on real measurements, while also calibrating faster to new sites. A complementary measurement-based study by Manukyan et~al.~\cite{manukyan_rt_limits} examines Sionna ray tracing fidelity against real cellular data in central Rome and finds that typical solver hyperparameters (path depth, diffuse/specular/refraction flags) have negligible effect on the correlation between simulated and measured power, whereas antenna location and orientation settings are decisive---so that even carefully configured ray tracing offers a ground truth whose alignment with real-world coverage depends substantially on antenna setup rather than on electromagnetic-solver detail. A related study by Manukyan~et~al.~\cite{manukyan_sim2real} examines sim-to-real transfer for DL-based RF-fingerprinting localization and reports that, while models trained purely on ray traced synthetic data degrade substantially on real measurements, high-fidelity synthetic pretraining still halves the real-world localization error compared with real-only training. A related but operationally different line of work studies radio map prediction under imperfect scene information when sparse measurements from the target environment are available at inference time: Jaensch~et~al. show that perturbation-aware training improves robustness to inaccuracies in geometry, material information, and transmitter position, and can outperform ray tracing and classical interpolation baselines on measured indoor data in that measurement-assisted setting~\cite{jaensch_indoor_robust}. Taken together, this body of evidence indicates that modern DL surrogates are competitive propagation predictors, supporting their use in large-scale planning benchmarks while at the same time motivating care in distinguishing benchmark optimality from real-world performance. This literature also points to two complementary routes for narrowing this gap in the present setting: training learned propagation models on real measurements collected in other urban environments~\cite{zhang_plnet,bakirtzis_deeplearn}, and improving scene fidelity through more accurate geometry, electromagnetic material properties, and antenna information.

The RadioMapSeer dataset was subsequently extended with additional subdatasets and a dedicated dataset paper \cite{radiomapseer}. Its 3D subset, \emph{RadioMap3DSeer}, was then used to organize a benchmark radio map prediction challenge~\cite{FirstChallenge,challengeoverview}, which, together with the subsequent indoor radio map challenges~\cite{FirstIndoorChallenge,sapra} based on the \emph{Indoor Radio Map Dataset}~\cite{IndoorDataSetAlt}, drove architectural development and furthered input-feature engineering~\cite{PMNet,REMUNet,Kehai2023GC,IPP-NET,Split_U_Net,TransPathNet,ResUnet,radionet,2,3,4,5,6,7}. More recent work in this line includes vision-transformer architectures for indoor pathloss radio map prediction using DINO-v2 pretrained encoders~\cite{mkrtchyan_vit}, and the construction of large-scale radio map datasets targeting sixth-generation XL-MIMO systems, extending propagation prediction from the current isotropic small-array regime to configurations with up to 32$\times$32 directional arrays across multiple frequency bands~\cite{li_u6g}. Related work also applies UNet-based heatmap prediction to adjacent wireless inference tasks beyond propagation modelling itself, such as single-BS NLOS device positioning trained on synthetic ray tracing data~\cite{khachatrian_nlos}. The IRT4 subdataset of RadioMapSeer, generated with the Intelligent Ray Tracing engine of WinProp with up to four ray--environment interactions~\cite{IRT} on the same \textcolor{black}{city maps}, is the ground-truth source for SAIPP-Net~\cite{radionet}---a UNet-based DL radio map estimator that ranked first in the most recent pathloss radio map prediction challenge~\cite{sapra}---and also for the present benchmark.

\subsubsection{ML-Assisted Transmitter and Base-Station Placement}

A more direct line of research couples learned propagation surrogates with downstream optimizers to reduce total planning cost. Seretis and Sarris train a U-Net on ray traced received-signal-strength (RSS) maps for 31 indoor geometries and use it as a fast evaluator inside a Hooke--Jeeves optimizer with multiple restarts, achieving a roughly 10$\times$ speedup per geometry while maintaining near-optimal coverage and mean-RSS objectives~\cite{seretis_unet}. Bakirtzis~et~al.\ combine the \mbox{EMDeepRay} indoor propagation surrogate~\cite{emdeepray} with four meta-heuristic optimizers (Bayesian optimization, genetic algorithm, \mbox{PSO}, and differential evolution) for femtocell placement in multi-room buildings, and show that replacing the ray tracer with the learned model reduces end-to-end optimization time from several hours to a few minutes while identifying deployments of equivalent throughput effectiveness~\cite{bakirtzis_iwn}. Cisse~et~al.\ train a conditional GAN (cGAN, E-IRGAN) on WinProp-generated radio maps from 33 floor plans and combine it with simulated annealing and genetic algorithms for indoor Wi-Fi AP placement~\cite{cisse_cgan}. Mallik and Villemaud train a cGAN to jointly predict RSS and EMF exposure maps from a campus environment and feed it into a Deep Q-Network (DQN) for sequential outdoor BS deployment under EMF-compliance constraints~\cite{mallik_gandqn}. Feng~et~al.\ embed a pretrained IPP-Net propagation surrogate into a D3QN-based sequential optimizer for indoor multi-AP deployment, using map-based states and a two-stage reward that first enforces a minimum coverage target and then improves redundancy in coverage; on four scenarios from the Indoor Radio Map Dataset, their RM-D3QN framework reports redundancy gains of up to 21.9\% over heuristic and prior DRL baselines while satisfying the coverage requirement~\cite{feng_rmd3qn_2026}. Lee and Molisch propose AutoBS, which uses PPO-based reinforcement learning with PMNet~\cite{PMNet} as a fast reward oracle to jointly place multiple BSs, achieving approximately 95\% of exhaustive capacity on a campus scenario~\cite{autoBS}. Li~et~al.\ introduce AutoPlan, which applies Bayesian optimization over a Sionna ray tracing digital twin calibrated with real-world RSRP measurements to tune BS deployment parameters on a single campus~\cite{autoplan}. Jaensch~et~al.\ exploit the differentiability of a trained pathloss predictor to perform gradient-based optimization of antenna directivity directly through the surrogate model~\cite{jaensch_aerial}. Romero~et~al.\ estimate radio-tomographic maps from sparse measurements and then solve a convex ADMM placement problem over the recovered maps for aerial BS deployment~\cite{romero_aerial}. A key distinction shared by these methods is that the learned model supports placement through downstream optimization or sequential decision-making at inference time, rather than predicting the final deployment in one shot.

The closest conceptual antecedent to the heatmap-based formulation in the present paper is the DA-cGAN framework of Liu~et~al.~\cite{DA-cGAN}, which reframes indoor radio design as image-to-image translation from a floor plan to an RSRP heatmap, from which radio-dot locations are then recovered by post-processing. DA-cGAN's training pairs are drawn from accumulated indoor radio designs produced in practice by human radio engineers working with RF-planner software: for each floor plan, the dot placement is the one an experienced engineer chose while consulting simulated coverage, and the paired heatmap is the corresponding coverage image from that design pipeline. The present paper's training pairs differ in the source of supervision: the heatmap target is the SAIPP-Net-predicted radio map induced by the best pixel found through exhaustive per-pixel search. The central representational choice, however, is the same in both cases: predicting a radio map and recovering the transmitter location from it, rather than regressing coordinates directly. This choice directly motivates the indirect heatmap-based approach studied here.

Zheng~et~al.\ propose OSSN, an end-to-end Transformer that jointly estimates radio maps and recommends a BS location among a fixed set of candidate points within a user-defined box, trained and evaluated on RadioMapSeer~\cite{ossn}. OSSN explicitly notes that ray traced ground truth was unavailable for arbitrary BS locations on RadioMapSeer, which provides ray traced maps for only 80 BS positions per map, and therefore uses learned-model predictions as ground truth while evaluating recommendation quality on a candidate box of 64 points. The present paper follows the same basic idea---learned propagation prediction enabling label generation at scale---but does so under exhaustive evaluation of approximately 10{,}000 feasible candidate positions per \textcolor{black}{city map} across 167{,}525 \textcolor{black}{city maps}, and provides dual ground-truth labels for both coverage and power objectives.

\paragraph{Direct score-map prediction.}
He and Zheng~\cite{avgmap_baseline_paper} introduce the score-map concept for neural transmitter placement: rather than predicting a radio propagation pattern and extracting a location from it, the model directly predicts a per-pixel optimality score for average received-power across the full candidate region from a \textcolor{black}{city map} alone, and the transmitter is placed at the highest-scoring pixel. This formulation---BSCSM-GAN, trained on 700 \textcolor{black}{city maps} with single-objective average-RSRP labels generated by RadioUNet---is the direct conceptual origin of the score-map approach studied in the present paper. The present work extends it in three directions: it provides dual ground-truth labels for both coverage and power objectives rather than a single power objective; it evaluates the score-map family across multiple architectures, loss functions, and training regimes under matched data conditions and at much larger scale (167{,}525 \textcolor{black}{city maps}); and it introduces multi-objective candidate-selection strategies (minimax ranking and union pooling) that go beyond single-pixel argmax selection.

\subsubsection{Generative Diffusion Models in Wireless Optimization}

Yuan and Cheng propose reward-guided diffusion sampling for indoor multi-AP deployment, in which the diffusion model generates AP coordinate distributions guided by a learned reward function capturing coverage, interference, and throughput~\cite{yuan_diffusion}. Their approach differs from the present work in the diffusion target (coordinate distributions vs.\ radio propagation maps), the multi-objective mechanism (explicit learned reward vs.\ post-hoc selection from a sample pool), and the scale (80 buildings at 50k CPU hours vs.\ 167,525 scenarios).

A parallel and growing body of work applies generative diffusion models to wireless network optimization problems beyond transmitter placement. \mbox{GDSG}~\cite{gdsg} reframes \mbox{MEC} computation offloading as a distribution-learning problem: a graph diffusion model learns the distribution of high-quality solutions from suboptimal training labels and achieves near-optimal cost reduction through parallel sampling, demonstrating that an exclusively optimal dataset is not required for convergence. \mbox{DiffSG}~\cite{diffsg} extends this distribution-learning framing to general network optimization and introduces the exceed-ratio metric to quantify how closely sampled solutions approach the optimum. \mbox{GDPlan}~\cite{gdplan} applies score-based graph diffusion to network-topology planning, achieving an order-of-magnitude speedup over commercial solvers on large-scale backbone networks. Wang~et~al.~\cite{wang_ris3d} apply a conditional \mbox{DDPM} to three-dimensional \mbox{RIS} deployment, treating the optimal surface-placement set as a sample from a distribution conditioned on base-station position, user locations, and obstacle geometry, and demonstrate generalization across unseen environment densities. Uslu~et~al.~\cite{gdmra} address a conceptually distinct aspect: they argue that optimal solutions to non-convex resource-allocation problems (e.g., power control) are inherently stochastic, and train a \mbox{GNN}-parameterized \mbox{GDM} to imitate a near-optimal stochastic expert policy rather than a deterministic one, with near-optimality emerging through sequential execution of sampled allocations~\cite{gdmra}. Liang~et~al.~\cite{liang_diffnet} survey these directions with an extended exploration of diffusion models as network optimizers across heterogeneous problem types. Collectively, this cluster shares the framing of optimal-solution distribution learning with the diffusion-based heatmap approach of the present paper, but operates over abstract resource-allocation vectors or graph topologies rather than spatial radio maps conditioned on \textcolor{black}{city map} geometry; the two formulations are therefore complementary in scope.

\subsection{Contributions}

This paper makes five main contributions:
\begin{enumerate}
\item It introduces \emph{RadioMapSeer-Deployment}, a large-scale benchmark of 167,525 urban scenarios with dual ground-truth labels for coverage-optimal and power-optimal transmitter placement, obtained by exhaustive SAIPP-Net evaluation under a fixed learned propagation model.
\item It provides a controlled comparison between two learning formulations for single-transmitter placement under this benchmark: indirect radio map prediction and direct score-map prediction. The latter was introduced for single-objective power optimization by He and Zheng~\cite{avgmap_baseline_paper}; here it is extended to dual ground-truth labels for both coverage and power objectives.
\item It shows that diffusion multi-sampling improves single-objective power or coverage performance through inference-time search, and that the same sample pool can also be reused for balanced multi-objective placement by post-hoc selection, without explicit multi-objective training.
\item It develops dual-score-map candidate-selection strategies that achieve near-optimal, and in the best settings optimal, balanced performance relative to the balanced benchmark optimum.
\item It quantifies the computational trade-off between discriminative prediction, diffusion-based search, and score-map-based candidate screening.
\end{enumerate}

The remainder of the paper is organized as follows. Section~\ref{sec:dataset} describes the dataset generation process, the dual-ground-truth system, and the resulting performance bounds. Section~\ref{sec:methods} summarizes the neural-network architectures and training procedures. Section~\ref{sec:heatmap} presents the heatmap-based approaches together with diffusion multi-sampling and discriminative-model results. Section~\ref{sec:scoremap} introduces the score-map-based methods and reports both single-objective and dual-score-map results. Section~\ref{sec:computational_cost} analyzes the computational cost of all approaches, and Section~\ref{sec:conclusion} concludes the paper and outlines future directions.

\section[Dataset of Optimal Radio Maps and Deployments (RadioMapSeer-Deployment)]{Dataset of Optimal Radio Maps and Deployments\\ (RadioMapSeer-Deployment)}
\label{sec:dataset}

\subsection{Dataset Generation Process}

We construct \emph{RadioMapSeer-Deployment}, a dataset of 167,525 urban configurations obtained from OpenStreetMap~\cite{OpenStreetMap}, with dual ground-truth labels for transmitter locations. Each \textcolor{black}{city map} is a 256$\times$256 binary image whose pixels indicate either building occupancy or free space. \textcolor{black}{Following the convention of the RadioMapSeer dataset~\cite{radiomapseer}, on whose IRT4 subset SAIPP-Net~\cite{radionet} is trained, candidate transmitter locations are restricted to the set of free space pixels in the center 150$\times$150 region (approximately 10,000 positions per city map) of each city map $\mathbf{B} \in \{0,1\}^{256 \times 256}$:
\begin{equation}\label{eq:FreeSpace}
 \mathcal{F} = \{ (y,x) \in \{54,\dots,203\}^2 : \mathbf{B}[y,x] = 0 \}.
\end{equation}
The RadioMapSeer training data uses transmitters placed only in this central region; SAIPP-Net's predictions are therefore reliable in this region by construction, while predictions outside it would extend the surrogate beyond its training regime and have no comparable reliability guarantee. The restriction is inherited from the underlying propagation-prediction dataset rather than introduced by the present methodology.}

For each \textcolor{black}{city map}, we then exhaustively evaluate all feasible candidates to identify the transmitter locations that optimize average received-power and average coverage, with both objectives computed exclusively over the same central 150$\times$150 free-space region $\mathcal{F}$.

\paragraph{Radio Map Prediction with SAIPP-Net.}
We use the SAIPP-Net model~\cite{radionet}, a UNet-based architecture trained on the IRT4 subset of the RadioMapSeer dataset~\cite{radiomapseer}, to predict received-power radio maps from (\textcolor{black}{city map}, transmitter) input pairs. The IRT4 dataset provides ray traced pathloss radio maps for urban environments, encoded as normalized gray levels $f \in [0,1]$ via $f = \max\{(\mathrm{PL}_{\mathrm{dB}} - \mathrm{PL}_{\mathrm{trnc}}) / (M_1 - \mathrm{PL}_{\mathrm{trnc}}),\; 0\}$, where $M_1 = -47$\,dB and $\mathrm{PL}_{\mathrm{trnc}} = -147$\,dB span a 100\,dB dynamic range. During training, a detection threshold $\tau = 0.4$ is applied: gray levels below $\tau$ are set to zero and the remaining range $[\tau, 1]$ is rescaled to $[0,1]$. This preprocessing concentrates the model's capacity on the 60\,dB range of physically relevant signal levels above the detection floor, improving prediction accuracy compared with training on the full 100\,dB range~\cite{radioUNet}. Since received-power and pathloss are related by $P_{\mathrm{rx}} = \mathrm{PL} + P_{\mathrm{Tx}}$, and the RadioMapSeer dataset uses $P_{\mathrm{Tx}} = 23$\,dBm, each pixel value $p \in \{0, \ldots, 255\}$ of the predicted radio map corresponds to a received-power of
\begin{equation}\label{eq:pixel_power}
P_{\mathrm{rx}}\;[\text{dBm}] = 60 \cdot \frac{p}{255} - 84.
\end{equation}
Thus $p = 0$ corresponds to $P_{\mathrm{rx}} = -84$\,dBm (the detection threshold, consistent with a receiver noise figure of 20\,dB at 10\,MHz bandwidth) and $p = 255$ to $P_{\mathrm{rx}} = -24$\,dBm (maximum signal at the transmitter location).

\begin{enumerate}

\item \textbf{Received-Power Radio Map Prediction and Average Received-Power Computation}: For each candidate position, we use SAIPP-Net~\cite{radionet} to predict the received-power radio map $\mathbf{P} \in [0,255]^{256 \times 256}$. Average received-power for a transmitter at $(y,x)$ is then computed as the mean pixel value over free pixels in the central region:
\begin{align}\label{eq:avgPwr}
\mathcal{A}_{\text{P}}(y,x) &= \frac{1}{|\mathcal{F}|} \sum_{(y',x') \in \mathcal{F}} \mathbf{P}_{(y,x)}[y',x']
\end{align}
where $\mathbf{P}_{(y,x)}$ denotes the SAIPP-Net-predicted received-power radio map induced by placing the transmitter at $(y,x)$. The pixel $(y^*_{\text{P}},x^*_{\text{P}})$ that maximizes average received-power is the power-optimal position:
\begin{equation}\label{eq:optTxPL}
(y^*_{\text{P}},x^*_{\text{P}})=\argmax_{(y',x') \in \mathcal{F}}\mathcal{A}_{\text{P}}(y',x')
\end{equation}

\emph{Remark (Connection to Shannon capacity and geometric mean).}
The average pixel value $\mathcal{A}_{\text{P}}$ is linearly related to average received-power in dBm via~\eqref{eq:pixel_power}. At high SNR, Shannon capacity per pixel satisfies $C = \log_2(1 + \mathrm{SNR}) \approx \log_2(\mathrm{SNR}) = (P_{\mathrm{rx,dBm}} - N'_{\mathrm{dBm}})/(10\log_{10}2)$, which is linear in $P_{\mathrm{rx,dBm}}$ and therefore also linear in the pixel value. Maximizing the average pixel value therefore approximately maximizes average spectral efficiency across the coverage area, which is a standard network-planning objective~\cite{goldsmith2005wireless}. The approximation error is less than 0.07\,bits/Hz for $\mathrm{SNR} > 20$\,dB, which holds for the vast majority of covered pixels. In addition, averaging in the dBm (logarithmic) domain corresponds to computing the geometric mean in the linear-power domain, $\mathcal{A}_{\text{P}} \propto \log\!\big(\prod_{\mathbf{y}} P_{\mathrm{rx,lin}}(\mathbf{y})\big)^{1/|\mathcal{F}|}$. Maximizing $\mathcal{A}_{\text{P}}$ therefore implicitly favors a more uniform power distribution: unlike the arithmetic mean, the geometric mean cannot be inflated by a few very strong pixels and instead rewards broad, fair coverage.

\item \textbf{Coverage Map and Average Coverage Computation}: For each candidate position, we derive a binary coverage map $\mathbf{C}_{(y,x)} \in \{0,1\}^{256 \times 256}$ from the predicted received-power map by assigning 1 to pixels above 0, i.e., to pixels whose received-power exceeds the detection threshold of $-84$\,dBm (cf.~\eqref{eq:pixel_power}):
\begin{equation}\label{eq:Cov}
\mathbf{C}_{(y,x)}[y',x'] = \begin{cases}
1 & \text{if } \mathbf{P}_{(y,x)}[y',x'] > 0 \\
0 & \text{otherwise}
\end{cases}
\end{equation}

Coverage then counts the number of pixels that receive signal above this threshold. Average coverage for a transmitter at $(y,x)$ is computed as the mean coverage over free pixels in the central region:
\begin{align}\label{eq:avgCov}
\mathcal{A}_{\text{Cov}}(y,x) &= \frac{1}{|\mathcal{F}|} \left| \{(y',x') \in \mathcal{F} : \mathbf{C}_{(y,x)}[y',x'] = 1\} \right|
\end{align}

The pixel $(y^*_{\text{Cov}},x^*_{\text{Cov}})$ that maximizes average coverage is the coverage-optimal position:
\begin{equation}\label{eq:optTxCov}
(y^*_{\text{Cov}},x^*_{\text{Cov}})=\argmax_{(y',x') \in \mathcal{F}}\mathcal{A}_{\text{Cov}}(y',x')
\end{equation}
When multiple transmitter positions achieve identical coverage counts, which occurs in approximately 5\% of \textcolor{black}{city maps}, we break ties by selecting the position with higher average received-power~\eqref{eq:avgPwr}.
\end{enumerate}

\subsubsection{Dataset Outputs}

For each \textcolor{black}{city map}, the generation pipeline produces three types of outputs: the two optimal transmitter locations (power-optimal and coverage-optimal), the received-power and coverage radio maps induced by those optimal placements, and the two score-maps that record the per-pixel objective values for power and coverage over the feasible region.

\begin{itemize}
\item \textbf{Ground-Truth Location Labels}: The optimal transmitter locations $(y^*_{\text{P}},x^*_{\text{P}})$ and $(y^*_{\text{Cov}},x^*_{\text{Cov}})$ are stored as one-hot images, yielding two coordinate targets per \textcolor{black}{city map}.
\item \textbf{Received-Power and Coverage Radio Maps of Optimal Deployments}: For both the power-optimal and coverage-optimal placements, we store the corresponding received-power and coverage maps, i.e., $\mathbf{P}_{(y^*_{\text{P}},x^*_{\text{P}})}$, $\mathbf{C}_{(y^*_{\text{P}},x^*_{\text{P}})}$, $\mathbf{P}_{(y^*_{\text{Cov}},x^*_{\text{Cov}})}$, and $\mathbf{C}_{(y^*_{\text{Cov}},x^*_{\text{Cov}})}$ (four maps per \textcolor{black}{city map}).
\item \textbf{Score-Maps (Aggregated Average Maps)}: For each \textcolor{black}{city map}, the average maps in \eqref{eq:avgPwr} and \eqref{eq:avgCov} are min--max normalized over the feasible set:

\begin{equation}
\mathcal{A}^{\text{norm}}(y,x) = \frac{\mathcal{A}(y,x) - \min_{(y',x') \in \mathcal{F}} \mathcal{A}(y',x')}{\max_{(y',x') \in \mathcal{F}} \mathcal{A}(y',x') - \min_{(y',x') \in \mathcal{F}} \mathcal{A}(y',x')}
\label{eq:avgmap_norm}
\end{equation}
These normalized score-maps are saved as PNG files together with the corresponding minimum and maximum values, which allow recovery of the original unnormalized scores in \eqref{eq:avgPwr} and \eqref{eq:avgCov}. The normalized score-map $\mathcal{A}^{\text{norm}} \in [0,1]$ assigns value~1 to the best-performing candidate and value~0 to the worst-performing one; in particular, $\mathcal{A}^{\text{norm}}_{\text{P}}(y^*_{\text{P}},x^*_{\text{P}})=1$ and $\mathcal{A}^{\text{norm}}_{\text{Cov}}(y^*_{\text{Cov}},x^*_{\text{Cov}})=1$. The original scores can be recovered as $\mathcal{A}(y,x) = \mathcal{A}^{\text{norm}}(y,x) \cdot (\mathcal{A}_{\max} - \mathcal{A}_{\min}) + \mathcal{A}_{\min}$ using the stored per-\textcolor{black}{city map} metadata $\mathcal{A}_{\max}$ and $\mathcal{A}_{\min}$ (two score-maps per \textcolor{black}{city map}).
\end{itemize}

\begin{figure*}[tbp]
\vspace{-0.0mm}
\centering
\captionsetup[subfloat]{justification=centering, singlelinecheck=false}
\subfloat[\textcolor{black}{City map}]{\includegraphics[width=0.24\columnwidth]{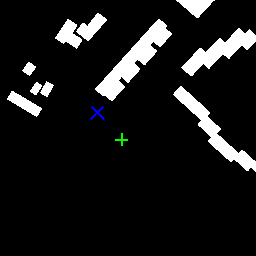}}
\subfloat[Feasible center region]{\includegraphics[width=0.24\columnwidth]{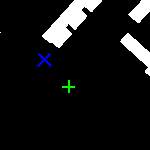}}
\subfloat[Normalized power score-map]{\includegraphics[width=0.24\columnwidth]{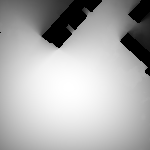}}
\subfloat[Normalized coverage score-map]{\includegraphics[width=0.24\columnwidth]{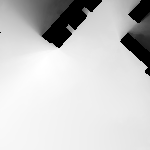}}\\
\subfloat[Power-optimal: received-power map]{\includegraphics[width=0.24\columnwidth]{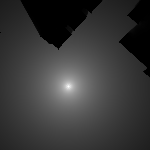}}
\subfloat[Power-optimal: coverage map]{\includegraphics[width=0.24\columnwidth]{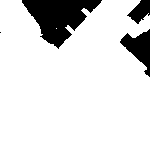}}
\subfloat[Coverage-optimal: received-power map]{\includegraphics[width=0.24\columnwidth]{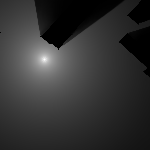}}
\subfloat[Coverage-optimal: coverage map]{\includegraphics[width=0.24\columnwidth]{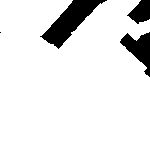}}
\caption{Example \textcolor{black}{city map} 167499 from the test dataset together with the supervision targets derived by exhaustive SAIPP-Net evaluation. The top row shows the \textcolor{black}{city map}, the feasible 150\,$\times$\,150 center region, and the normalized power and coverage score-maps. The green cross marks the ground truth power-optimal location, and the blue cross marks the ground truth coverage-optimal location. The bottom row shows the received-power and coverage maps induced by the power-optimal and coverage-optimal transmitter placements. The figure illustrates the dual ground-truth labels available for each \textcolor{black}{city map} and the spatial difference between the two single-objective optima. }
\label{fig:GTs}
\end{figure*}

Fig.~\ref{fig:GTs} illustrates these supervision targets for one representative test \textcolor{black}{city map} (scenario 167499). The same \textcolor{black}{city map} is used later as the benchmark scenario in Fig.~\ref{fig:Results} for qualitative model comparison, and in Figs.~\ref{fig:pareto}--\ref{fig:paretoZoom} for the objective-space scatter plots.

\subsection{Computational Implementation}
\label{sec:implementation}

The dataset-generation pipeline evaluates approximately 10,000--18,000 candidate transmitter positions per \textcolor{black}{city map} across 167,525 \textcolor{black}{city maps}, resulting in roughly $\sim$1.67 billion SAIPP-Net inference operations in total. The mean number of feasible free-space candidate positions per \textcolor{black}{city map} is $17{,}320$ in the training split, $17{,}354$ in the validation split, and $17{,}322$ in the test split, out of a maximum of $150 \times 150 = 22{,}500$ pixels in the feasible region. At this scale, practical dataset generation depends on batched GPU inference rather than sequential per-candidate evaluation, as summarized below.

\textbf{Memory-Efficient Processing}: For each \textcolor{black}{city map}, the 256$\times$256 map is loaded onto the GPU once and then reused across all candidate evaluations; only the one-hot transmitter layer changes across the batch. This reduces repeated transfers and lowers memory-bandwidth pressure from $O(N \cdot H \cdot W)$ to $O(H \cdot W + N)$ for $H=W=256$.

\textbf{Batch Construction}: For each batch of candidate positions $\{(y_i, x_i)\}_{i=1}^{64}$, the input tensor is formed as
\begin{align}
\mathbf{I}_i = [\mathbf{B}, \mathbf{T}_i] \in \mathbb{R}^{2 \times 256 \times 256},
\end{align}
where $\mathbf{B}$ denotes the \textcolor{black}{city map} and $\mathbf{T}_i$ is the one-hot transmitter map with $\mathbf{T}_i[y_i, x_i]=1$ and zeros elsewhere.

\textbf{Performance Characteristics}: With batch size 64, the implementation achieves about 255 predictions per second, corresponding to a 10--18$\times$ speedup over sequential evaluation. End-to-end exhaustive labeling of one \textcolor{black}{city map} then takes about 0.9--1.2 minutes, depending on \textcolor{black}{city map} density and I/O.

This computational scale---approximately $1.67$ billion SAIPP-Net inference operations in total---is why a learned propagation model is required for exhaustive dataset generation at this density. Running the same $1.67$ billion evaluations with the underlying IRT4 ray tracer~\cite{IRT,radiomapseer}, which takes on the order of $30$--$60$\,seconds per evaluation depending on scene complexity, would raise per-city map labeling time from roughly one minute to several tens of hours and push full dataset generation into millennia of single-GPU compute. Measurement-based \textcolor{black}{evaluation} is even further out of reach in principle, since it would require physically placing a transmitter at every candidate location and collecting measurements across the full region of interest for each placement. SAIPP-Net therefore makes exhaustive per-location benchmark labeling feasible at this scale, while remaining a high-fidelity approximation to the underlying ray tracer. \textcolor{black}{Both supervision and evaluation in this paper are therefore defined under a single fixed learned propagation oracle; the implications for real-world deployment and the route to narrowing the resulting sim-to-real gap are discussed in Section~\ref{sec:relatedwork} and the Conclusions.}

\emph{Remark.} This benchmark construction also clarifies the main routes for reducing the sim-to-real gap in the present problem. A natural extension would be to train SAIPP-Net, or an analogous learned propagation model, on large-scale real measurement data collected in urban environments rather than only on simulation-derived labels, and then use that trained model as the fixed predictor for exhaustive offline candidate evaluation on previously unseen deployment instances. A second and complementary direction is to improve the fidelity of the environment representation itself, for example by incorporating more accurate geometry, electromagnetic material properties, antenna characteristics, or other site-specific scene information. By contrast, approaches that require measurements from the target urban deployment instance are not applicable in the present setting, whether those measurements are used directly as inference-time inputs or for site-specific adaptation, because that would require collecting data for the very deployment instance whose hypothetical transmitter candidates are being ranked.

\subsection{Computational Resources}

Experiments were conducted on NVIDIA Quadro RTX 6000 GPUs with 22.5~GB memory and CUDA~12.8. All computations use FP32 precision for consistency with the exhaustive-search baseline, and runtime measurements include both model inference and SAIPP-Net evaluation of coverage and power.

\subsection{Benchmark Ground-Truth Characteristics and Performance Bounds}
\label{sec:gt_bounds}

Exhaustive evaluation across the test set establishes the theoretical performance limits of the benchmark. For each \textcolor{black}{city map}, we evaluated all candidate transmitter positions (10,000--18,000 candidates) using SAIPP-Net and computed both coverage and power metrics. This exhaustive search yields the following reference points:

\textbf{Power-Optimal Positions}: Locations maximizing power efficiency achieve perfect power (100.00\%) while reducing coverage to $94.50\pm6.53$\% (mean$\pm$standard deviation) of coverage-optimal performance.

\textbf{Coverage-Optimal Positions}: Locations maximizing coverage achieve perfect coverage (100.00\%) while sacrificing power efficiency to $86.14\pm11.40$\% of power-optimal performance.

The average pixel distance between the power- and coverage-optimal positions is $36.18\pm24.12$.

\textbf{Balanced-Optimal Positions}: Locations minimizing $L_2$ distance to the ideal point (100\% coverage, 100\% power) (Section~\ref{sec:l2_metric}) achieve $97.84$\% coverage and $98.55$\% power efficiency, with $\bar{d}=2.60$. This is the best achievable simultaneous performance on both objectives.

These bounds quantify a fundamental trade-off between coverage and power: perfect performance on one objective requires sacrificing the other. The trade-off is asymmetric---coverage-optimal placement sacrifices $13.86\%$ of power, whereas power-optimal placement sacrifices only $5.50\%$ of coverage---and the two optima are typically spatially distinct. The asymmetry is also visible qualitatively in model outputs: coverage-trained models tend to produce broader predicted regions than power-trained models (Fig.~\ref{fig:Results}), consistent with the less regular objective landscape of the coverage objective.

This asymmetry has a partial structural explanation in the differing nature of the two objectives. Both metrics operate on the same per-pixel received-power values, but they reward strong pixels differently. Coverage is a saturating count above the $-84$\,dBm threshold: once a pixel exceeds the threshold, additional signal there yields no further gain. A coverage-optimal placement can therefore profitably extend signal into geometrically harder-to-reach regions where pixels only barely cross threshold; such marginal pixels contribute fully to coverage yet remain weak in dBm and therefore lower the average power. Average power, by contrast, is non-saturating: every additional dB at any pixel improves the score. Because this score is averaged over the region, a power-optimal transmitter tends to sit where it reaches the largest number of pixels efficiently---that is, pixels close to the transmitter and with few intervening buildings, which receive strong signal at low geometric cost. Such placements naturally generate strong signal over broad accessible regions and therefore also deliver wide above-threshold coverage as a by-product. Coverage-optimal placements behave differently: they may shift toward locations that extend signal into distant or heavily shadowed pixels, because each such pixel that is pushed just above threshold contributes fully to coverage while remaining weak in dBm, thereby dragging down average power. Power-optimal placements thus lie naturally close to balanced behavior, whereas coverage-optimal placements generally do not.

\subsection{Dataset Statistics and Split}
The dataset comprises 167,525 urban configurations spanning diverse urban morphologies. We use a fixed-seed (42) 80/10/10 split for reproducibility:

\begin{itemize}
\item \textbf{Training set}: 134,020 \textcolor{black}{city maps}
\item \textbf{Validation set}: 16,752 \textcolor{black}{city maps}
\item \textbf{Test set}: 16,753 \textcolor{black}{city maps}
\end{itemize}

All reported results are computed on the held-out test set.

\textbf{Reproducibility.} \emph{RadioMapSeer-Deployment} is released on IEEE DataPort~\cite{radiomapseerdeployment1TxAlt}, and the code is made publicly available at \texttt{https://github.com/CagkanYapar/Deployment1Tx}.

\subsection{Evaluation Metrics}
\label{sec:l2_metric}

We evaluate every model against both optimization objectives, regardless of its training objective, and we additionally quantify balanced performance across the two.

\textbf{Primary Metric}: This is performance on the training objective---Power\% for power-trained models and Coverage\% for coverage-trained models. Coverage and power percentages are computed relative to the corresponding optimum for each \textcolor{black}{city map}. For example, if a \textcolor{black}{city map}'s coverage-optimal position covers 1000 pixels, then a prediction covering 950 pixels scores $95$\% coverage; similarly, if the power-optimal position achieves an average power of 200, then a prediction achieving 190 scores $95$\% power.

\textbf{Secondary Metric}: This is performance on the alternate objective, i.e., coverage for power-trained models and power for coverage-trained models, and it reveals the trade-off induced by single-objective training.

\textbf{Coordinate Error}: We report Euclidean distance in pixels between predicted and ground-truth coordinates, measured against both the power-optimal and coverage-optimal locations to assess spatial accuracy from both perspectives.

\textcolor{black}{\emph{Remark (Note on metric hierarchy).} The primary quality measures in this paper are the achieved coverage and achieved power percentages relative to the per-environment optimum (the ``Primary Metric'' and ``Secondary Metric'' above). Coordinate (Euclidean) error is reported as a supplementary spatial diagnostic only: spatial displacement of the predicted transmitter from the ground-truth optimal location does not necessarily imply degraded objective performance, since broad regions of the radio environment can be near-equivalent in coverage and power. A predicted transmitter location at non-zero coordinate error from the optimal one can therefore still achieve near-optimal coverage or power, and this case is correctly counted as a near-optimal placement under the primary metrics. Performance comparisons in the remainder of the paper are judged on objective achievement, with coordinate error included as a secondary diagnostic.}

\textbf{Multi-Objective Performance Metric}: To evaluate balanced performance across both objectives, we measure the $L_2$ distance from the ideal point (100\% coverage, 100\% power):
\begin{equation}
\bar{d} = \sqrt{(100 - \bar{C})^2 + (100 - \bar{P})^2}
\label{eq:l2_metric}
\end{equation}
where $\bar{C}$ and $\bar{P}$ denote mean coverage and power percentages over the test set. This metric~\cite{miettinen1999nonlinear}, known as the global criterion method with $p=2$, naturally penalizes imbalanced solutions. For example, a solution achieving $(97\%, 97\%)$ gives $\bar{d}=4.24$, whereas $(100\%, 94\%)$ gives $\bar{d}=6.00$ despite both averaging $97$\%. The $L_2$ metric therefore favors solutions that perform well on \emph{both} objectives simultaneously rather than excelling on one while sacrificing the other.

\textcolor{black}{A natural alternative balance metric is the harmonic mean
$H = 2CP/(C+P)$. For solutions with the same arithmetic mean, writing
$C = m + \delta$ and $P = m - \delta$ gives
\begin{equation*}
  H = m - \frac{\delta^2}{m}, \qquad
  d^2 = 2(100-m)^2 + 2\delta^2.
\end{equation*}
Both criteria penalize imbalance quadratically, but the harmonic-mean penalty is attenuated by $1/m$ and becomes very small in the high-performance regime ($m$ close to $100$), whereas the ideal-point $L_2$ metric retains an undamped quadratic imbalance penalty regardless of $m$. For instance, $(97\%,97\%)$ and $(100\%,94\%)$ both average $97\%$; their harmonic means are $97.00$ and $96.91$ (a $0.09\%$ relative difference), while their $L_2$ distances are $4.24$ and $6.00$ (a $42\%$ relative difference). The $L_2$ metric therefore distinguishes balanced from imbalanced solutions far more sharply in the regime in which the paper's models operate. The two metrics also induce different rankings when solutions differ in their arithmetic means: $(99,91)$ has $d \approx 9.06$ and $H \approx 94.83$, while $(94,94)$ has $d \approx 8.49$ and $H = 94$; the harmonic mean prefers the higher-average $(99,91)$ despite its larger imbalance, while the $L_2$ criterion prefers the balanced $(94,94)$. Since the substantive interest here is in distinguishing placements that perform well on both objectives from placements that win one while sacrificing the other, the $L_2$ criterion is retained as the primary balanced metric throughout the paper.}

For most metrics, we report mean and standard deviation, written as mean$\pm$standard deviation, over the 16,753 test \textcolor{black}{city maps}.

\section{Methods}
\label{sec:methods}

This paper compares two fundamentally different learning formulations for single-transmitter placement. Evaluation uses the metrics defined in Section~\ref{sec:l2_metric}, and all approaches are measured against the benchmark performance bounds established in Section~\ref{sec:gt_bounds}.

The first formulation is \emph{indirect}: given a \textcolor{black}{city map}, the model predicts a received-power radio map associated with an objective-optimal transmitter, and the transmitter location is then recovered by applying argmax to the predicted map. The second formulation is \emph{direct}: given the same \textcolor{black}{city map}, the model predicts the objective value at every feasible transmitter location, so transmitter selection becomes a search over the predicted score landscape. In the present single-transmitter setting, both formulations are trainable because exhaustive evaluation under the fixed learned propagation model provides ground-truth labels for the optimal radio maps (indirect target) and for the full per-location objective maps (direct target).

Heatmap-based (indirect) models, developed in Section~\ref{sec:heatmap}, are trained to predict a received-power radio map induced by an objective-optimal transmitter placement, after which the final coordinate is obtained by pixel-wise argmax on the predicted map. Two model classes are studied in this family: deterministic discriminative networks that produce a single prediction per \textcolor{black}{city map}, and a diffusion model whose stochastic inference supports drawing multiple samples per \textcolor{black}{city map} and selecting among them under a chosen criterion.

Score-map (direct) models, developed in Section~\ref{sec:scoremap}, are trained to predict the power or coverage score at every feasible transmitter location. This formulation was introduced by He and Zheng~\cite{avgmap_baseline_paper} for single-objective average-power optimization; the present paper extends it to dual objectives and evaluates it systematically against the indirect heatmap formulation. Transmitter selection is performed either by argmax on the predicted score-map or by evaluating a top-$K$ shortlist with SAIPP-Net under a single- or multi-objective criterion.

The following sections present the architectures, training setups, and results for each approach.

\section{Heatmap-Based (Indirect) Approaches}
\label{sec:heatmap}

\subsection{Diffusion Model}
\label{ssec:diffusion}

We employ a denoising diffusion model based on the DDPM framework~\cite{ho2020denoising} to learn the distribution of received-power radio maps associated with objective-optimal transmitter placements.

\textbf{Architecture}: The model uses a UNet-style denoising network with standard encoder--decoder skip connections and timestep-conditioned residual blocks in the encoder. At each denoising iteration, the denoising network takes the \textcolor{black}{city map} and the current noisy heatmap as two concatenated input channels. Training uses $T=1000$ diffusion timesteps with a linear $\beta$ schedule ($\beta \in [0.0001, 0.02]$), AdamW with learning rate $10^{-4}$ and weight decay $10^{-5}$, a cosine-annealing learning-rate schedule, and standard training-time data augmentation through random flips and rotations. Full architectural and training hyperparameters, including channel dimensions, timestep embedding dimension, batch size, and number of epochs, are provided in the released code repository.

\textbf{Conditioning}: Conditioning is implemented by concatenating the \textcolor{black}{city map} with the current noisy heatmap at the input of each denoising-network call. Timestep information is injected into encoder residual blocks, so each denoising prediction remains conditioned on both geometry and diffusion timestep throughout the reverse process.

\textbf{Training Objective}: Separate diffusion models are trained for the two optimization objectives. The power-trained model uses received-power radio maps induced by power-optimal transmitter placements, whereas the coverage-trained model uses received-power radio maps induced by coverage-optimal placements. In each case, the model learns to predict the Gaussian noise $\epsilon$ added to the corresponding ground-truth radio maps:
\begin{equation}
\mathcal{L}_{\text{diffusion}} = \mathbb{E}_{t,\epsilon} \left[\|\epsilon - \epsilon_\theta(\mathbf{x}_t, t, \mathbf{B})\|^2\right]
\end{equation}
where $\mathbf{x}_t$ is the noisy heatmap at timestep $t$, $\mathbf{B}$ is the \textcolor{black}{city map}, and $\epsilon_\theta$ is the denoising network.

\textbf{Inference}: Starting from Gaussian noise, the model iteratively denoises over 50 steps to produce a predicted received-power radio map, using a DDIM sampler~\cite{ddim} for substantially faster inference than the 1000-step DDPM trajectory used during training. The estimated transmitter location is then extracted by argmax on the generated map.

\textbf{Multi-Sampling Capability}: Unlike discriminative models, which produce a single deterministic prediction, the diffusion model generates a different radio map sample for each noise realization. Drawing multiple samples per \textcolor{black}{city map} therefore turns inference into a small search procedure over the learned output distribution. Selecting the best sample under the training objective improves single-objective performance, while ranking the same sample set under a balanced criterion yields a post-hoc trade-off placement without changing the training objective.

\textbf{Selection Strategies}: We evaluate $N \in \{1, 5, 10, 20, 50, 100\}$ samples per \textcolor{black}{city map}. When $N>1$, the following selection strategies are applied:
\begin{itemize}
\item \textbf{single}: Use a single sample ($N=1$, no selection).
\item \textbf{best\_power}: Select the sample with the highest average received-power.
\item \textbf{best\_coverage}: Select the sample with the highest coverage.
\item \textbf{best\_l2}: Select the sample with minimum per-instance $L_2$ distance from the ideal point $(100\%, 100\%)$. Similar to the dataset-level $L_2$ metric in \eqref{eq:l2_metric}, the per-instance distance is
\begin{equation}
d_n = \sqrt{(100 - C_n)^2 + (100 - P_n)^2}
\label{eq:l2dist}
\end{equation}
where $C_n$ and $P_n$ are the coverage and power percentages of sample $n$.
\end{itemize}

Selection is performed by evaluating all generated candidates with SAIPP-Net.

\textbf{Computational Implementation}: Diffusion inference involves 50 denoising steps per sample. To accelerate multi-sample generation, we use batched diffusion sampling, in which multiple samples for the same \textcolor{black}{city map} are processed simultaneously during iterative denoising. Concretely, the \textcolor{black}{city map} is replicated across the batch dimension, and different noise initializations are propagated in parallel through the same denoising steps. With batch size 16, generation requires $\lceil N/16 \rceil$ batched forward passes: for example, $N=50$ requires $\lceil 50/16 \rceil = 4$ passes rather than 50 sequential passes, while $N=10$ requires only one batched pass rather than 10 separate ones. Each pass still runs the full 50-step denoising trajectory, so the speedup is moderate rather than ideal, about $1.3$--$2.2\times$ depending on $N$. We evaluated batch sizes of 1, 4, 8, 16, and 32 to determine the best configuration. Detailed timing results are reported in Section~\ref{sec:computational_cost}.

\subsection{Discriminative models}
\label{ssec:discriminative}

We evaluate four discriminative neural architectures for direct heatmap prediction.

\textbf{DeepXL}: \textcolor{black}{DeepXL is a UNet architecture~\cite{UNet} with a symmetric encoder--decoder structure and skip connections. It has the same UNet architecture as that of the diffusion model (channel progression $64\to128\to256\to384\to512\to640\to768$, with matched conv-block definitions, skip connections, and up- and downsampling layers).} Because UNet-style models are widely used for image-to-image translation, DeepXL serves as the primary baseline in this study.

\textbf{PMNet~\cite{PMNet}}: PMNet is a ResNet-based~\cite{He_2016_CVPR} encoder--decoder in which the encoder builds hierarchical features through bottleneck residual blocks with dilated convolutions. An ASPP bottleneck~\cite{ASPP} aggregates multi-scale context over several atrous rates, and the decoder restores spatial resolution using learnable ConvTranspose upsampling with skip connections. This model ranked first in the First Pathloss Radio Map Prediction Challenge~\cite{FirstChallenge}.

\textbf{DC-Net~\cite{DCNet}}: DC-Net is a single U-Net whose encoder stages are augmented with AOT blocks~\cite{AOT}. Each AOT block applies multiple parallel dilated convolutions at increasing dilation rates, fuses their outputs, and adaptively blends the result back into the feature map through a learned sigmoid gate. This gating mechanism allows the network to selectively incorporate long-range context at each spatial location. The decoder mirrors the encoder symmetrically and uses standard skip connections.

\textbf{SIP2Net~\cite{SIP2Net}}: SIP2Net uses the same ResNet encoder--decoder backbone as PMNet, but replaces the standard $3\times3$ bottleneck convolutions in deeper encoder stages with ACNet~\cite{ACNet} asymmetric convolutions, which sum horizontal, vertical, and square-kernel branches to increase structural sensitivity. A reduce layer bridges the standard and ACNet-equipped stages, and the final prediction block also uses ACNet. This model ranked first in the First Indoor Radio Map Prediction Challenge~\cite{FirstIndoorChallenge}. Different from the original implementation, however, we do not use adversarial training in the present study.

All discriminative models share the same training hyperparameters: AdamW with learning rate $0.01$ and weight decay $10^{-5}$, a cosine-annealing learning-rate schedule, batch size 16, 200 training epochs, and standard training-time augmentation via random flips and rotations. Further model-specific settings are provided in the released code repository.

\textcolor{black}{All architectures are adapted from their original sources with the minimal modifications required to process the 150$\times$150 pixel center city maps used here; the trained models and exact configurations are released in the project repository. We do not claim to have found the best per-architecture configuration, and the architecture comparison is reported as a secondary result alongside the main formulation-level study.}

\subsection{Loss Functions}
\label{sec:loss_functions}

We train the discriminative models with multiple loss-function configurations. Let $\mathbf{\hat{H}} \in \mathbb{R}^{H \times W}$ denote the predicted heatmap, $\mathbf{H}_{gt} \in \mathbb{R}^{H \times W}$ the ground-truth heatmap (either a Gaussian centered on the optimal transmitter or the received-power / coverage-induced radio map), and $(x_{pred}, y_{pred})$ and $(x_{gt}, y_{gt})$ the predicted and ground-truth coordinates, respectively.

\subsubsection{Pixel-Level Reconstruction Losses}

\textbf{L2 Loss (Mean Squared Error):} Mean squared error measures pixel-wise differences between predicted and target heatmaps:
\begin{equation}
\mathcal{L}_{\text{L2}} = \frac{1}{HW} \sum_{i=1}^{H} \sum_{j=1}^{W} (\mathbf{\hat{H}}[i,j] - \mathbf{H}_{gt}[i,j])^2
\end{equation}

\textbf{L1 Loss (Mean Absolute Error):} L1 loss provides a more robust alternative by reducing sensitivity to outliers:
\begin{equation}
\mathcal{L}_{\text{L1}} = \frac{1}{HW} \sum_{i=1}^{H} \sum_{j=1}^{W} |\mathbf{\hat{H}}[i,j] - \mathbf{H}_{gt}[i,j]|
\end{equation}

\subsubsection{Structural Similarity Losses}

\textbf{SSIM (Structural Similarity Index):} SSIM measures perceptual similarity by comparing local patterns of luminance, contrast, and structure~\cite{ssim}:
\begin{equation}
\text{SSIM}(\mathbf{x}, \mathbf{y}) = \frac{(2\mu_x\mu_y + C_1)(2\sigma_{xy} + C_2)}{(\mu_x^2 + \mu_y^2 + C_1)(\sigma_x^2 + \sigma_y^2 + C_2)}
\end{equation}
where $\mu_x$ and $\mu_y$ are local means, $\sigma_x^2$ and $\sigma_y^2$ are local variances, $\sigma_{xy}$ is covariance, and $C_1$ and $C_2$ are stability constants. The corresponding loss is:
\begin{equation}
\mathcal{L}_{\text{SSIM}} = 1 - \text{SSIM}(\mathbf{\hat{H}}, \mathbf{H}_{gt})
\end{equation}

\textbf{MS-SSIM (Multi-Scale SSIM):} MS-SSIM extends SSIM by measuring similarity across multiple image scales, thereby capturing both fine detail and coarse spatial structure~\cite{ms-ssim}:
\begin{equation}
\text{MS-SSIM}(\mathbf{x}, \mathbf{y}) = [l_M(\mathbf{x}, \mathbf{y})]^{\alpha_M} \cdot \prod_{j=1}^{M} [c_j(\mathbf{x}, \mathbf{y})]^{\beta_j} [s_j(\mathbf{x}, \mathbf{y})]^{\gamma_j}
\end{equation}
where $l_M$ is luminance comparison at the coarsest scale, $c_j$ and $s_j$ are contrast and structure comparisons at scale $j$, and $\alpha_M$, $\beta_j$, and $\gamma_j$ are weights. The loss is:
\begin{equation}
\mathcal{L}_{\text{MS-SSIM}} = 1 - \text{MS-SSIM}(\mathbf{\hat{H}}, \mathbf{H}_{gt})
\end{equation}

\subsubsection{Gradient-Based Losses}

\textbf{Total Variation (TV) Loss:} TV loss encourages spatial smoothness by penalizing high-frequency variation~\cite{tv_loss}:
\begin{equation}
\mathcal{L}_{\text{TV}} = \frac{1}{HW} \sum_{i=1}^{H-1} \sum_{j=1}^{W-1} \left( |\mathbf{\hat{H}}[i+1,j] - \mathbf{\hat{H}}[i,j]| + |\mathbf{\hat{H}}[i,j+1] - \mathbf{\hat{H}}[i,j]| \right)
\end{equation}

\textbf{Gradient Difference Loss (GDL):} GDL preserves edge sharpness by penalizing differences between the spatial gradients of the prediction and those of the target~\cite{gdl}:
\begin{equation}
\mathcal{L}_{\text{GDL}} = \frac{1}{HW} \sum_{i,j} \left( |\nabla_x \mathbf{\hat{H}}[i,j] - \nabla_x \mathbf{H}_{gt}[i,j]| + |\nabla_y \mathbf{\hat{H}}[i,j] - \nabla_y \mathbf{H}_{gt}[i,j]| \right)
\end{equation}
where $\nabla_x$ and $\nabla_y$ are horizontal and vertical finite-difference gradients.

\textbf{Multi-Scale Gradient Similarity (MS-GSIM):} MS-GSIM measures gradient similarity across multiple scales and is intended to preserve edges at different resolutions:
\begin{equation}
\text{MS-GSIM}(\mathbf{x}, \mathbf{y}) = \prod_{j=1}^{M} \text{GS}_j(\nabla \mathbf{x}, \nabla \mathbf{y})^{\beta_j}
\end{equation}
where $\text{GS}_j$ computes gradient similarity at scale $j$. The loss is:
\begin{equation}
\mathcal{L}_{\text{MS-GSIM}} = 1 - \text{MS-GSIM}(\mathbf{\hat{H}}, \mathbf{H}_{gt})
\end{equation}

\subsubsection{Focal Loss}

\textbf{Focal Loss:} Focal loss addresses class imbalance by down-weighting easy examples and emphasizing harder ones~\cite{focal_loss}. For heatmap regression, we adapt it as:
\begin{equation}
\mathcal{L}_{\text{Focal}} = \frac{1}{HW} \sum_{i,j} -(1 - p_{ij})^\gamma \log(p_{ij})
\end{equation}
where $p_{ij} = 1 - |\mathbf{\hat{H}}[i,j] - \mathbf{H}_{gt}[i,j]|$ represents prediction confidence and $\gamma$ is the focusing parameter (typically $\gamma = 2$). This formulation places greater emphasis on pixels with larger prediction errors.

\subsubsection{Coordinate Supervision Loss}

\textbf{Coordinate Loss:} Direct coordinate supervision uses Euclidean distance between predicted and target transmitter coordinates:
\begin{equation}
\mathcal{L}_{\text{coord}} = \sqrt{(x_{pred} - x_{gt})^2 + (y_{pred} - y_{gt})^2}
\end{equation}
where the predicted coordinates are obtained by soft-argmax so that training remains differentiable. In practice, experiments using only this loss together with a fully connected prediction head did not achieve reasonable accuracy, because training tended to collapse toward trivial center-coordinate solutions.

\subsubsection{Loss Configurations}

We investigate five loss-function configurations built from the components above. Four are used in the main experiments, and the fifth (L2+Coordinate) is reported as a negative result.

\begin{enumerate}
\item \textbf{L2 Only}: A baseline configuration that provides direct pixel-wise supervision via MSE. It is simple and interpretable, but may under-emphasize structural fidelity.
\begin{equation}
\mathcal{L}_{\text{total}} = \lambda_{\text{L2}} \cdot \mathcal{L}_{\text{L2}}
\end{equation}
where $\lambda_{\text{L2}} = 10^7$ provides appropriate scaling for the heatmap values.

\item \textbf{DA-cGAN Stage 2}: This configuration emphasizes structural fidelity through multi-scale structural similarity (MS-SSIM), spatial smoothness (TV), and hard-example focus (Focal)~\cite{DA-cGAN}:
\begin{equation}
\mathcal{L}_{\text{total}} = \lambda_{\text{MS-SSIM}} \cdot \mathcal{L}_{\text{MS-SSIM}} + \lambda_{\text{TV}} \cdot \mathcal{L}_{\text{TV}} + \lambda_{\text{Focal}} \cdot \mathcal{L}_{\text{Focal}}
\end{equation}
with $\lambda_{\text{MS-SSIM}} = 8.4 \times 10^6$, $\lambda_{\text{TV}} = 10^3$, and $\lambda_{\text{Focal}} = 10^7$.

\item \textbf{Hybrid L2+MSGSIM}: This configuration combines pixel-level reconstruction with multi-scale gradient preservation:
\begin{equation}
\mathcal{L}_{\text{total}} = \lambda_{\text{L2}} \cdot \mathcal{L}_{\text{L2}} + \lambda_{\text{MS-GSIM}} \cdot \mathcal{L}_{\text{MS-GSIM}}
\end{equation}
with $\lambda_{\text{L2}} = 6 \times 10^6$ and $\lambda_{\text{MS-GSIM}} = 4 \times 10^6$. The intent is to combine MSE accuracy with edge-aware reconstruction.

\item \textbf{SIP2Net}: This multi-component loss balances pixel accuracy, structural similarity, and edge preservation~\cite{SIP2Net}:
\begin{equation}
\mathcal{L}_{\text{total}} = \lambda_{\text{L1}} \cdot \mathcal{L}_{\text{L1}} + \lambda_{\text{SSIM}} \cdot \mathcal{L}_{\text{SSIM}} + \lambda_{\text{GDL}} \cdot \mathcal{L}_{\text{GDL}}
\end{equation}
with $\lambda_{\text{L1}} = 10^7$, $\lambda_{\text{SSIM}} = 5 \times 10^6$, and $\lambda_{\text{GDL}} = 10^4$.

\item \textbf{L2 + Coordinate}: This configuration combines heatmap reconstruction with direct coordinate supervision:
\begin{equation}
\mathcal{L}_{\text{total}} = \lambda_{\text{L2}} \cdot \mathcal{L}_{\text{L2}} + \lambda_{\text{coord}} \cdot \mathcal{L}_{\text{coord}}
\end{equation}
with weights determined empirically. In practice, however, we observed that coordinate loss conflicts with heatmap-fidelity objectives. A prediction that yields an accurate propagation pattern but is shifted spatially by one or two pixels---which is common under soft-argmax or discretization---can incur large coordinate error despite being physically correct. This conflict degraded training stability. Pure heatmap reconstruction therefore learned better representations, allowing the model to focus on physically correct fields while still recovering accurate transmitter coordinates through the argmax operation at inference time.

\end{enumerate}

\textcolor{black}{The relative weights of individual loss terms within each configuration were chosen by rough trial and error on the validation set rather than by full grid search; the loss-configuration study, like the architecture comparison, is reported as a secondary result, and we do not claim optimality of the chosen weights.} All models use learning rate $\mathrm{LR}=0.01$. The final weights are summarized in Table~\ref{tab:loss_configs}. 

\begin{table}[tbp]
\centering
\caption{Loss Function Configurations and Hyperparameters}
\label{tab:loss_configs}
\small
\resizebox{\textwidth}{!}{%
\begin{tabular}{lcccccccc}
\toprule
\textbf{Configuration} & \textbf{L2} & \textbf{L1} & \textbf{SSIM} & \textbf{MS-SSIM} & \textbf{TV} & \textbf{GDL} & \textbf{MS-GSIM} & \textbf{Focal} \\
\midrule
L2 Only & $10^7$ & 0 & 0 & 0 & 0 & 0 & 0 & 0 \\
\midrule
DA-cGAN Stage 2 & 0 & 0 & 0 & $8.4 \times 10^6$ & $10^3$ & 0 & 0 & $10^7$ \\
\midrule
Hybrid L2+MSGSIM & $6 \times 10^6$ & 0 & 0 & 0 & 0 & 0 & $4 \times 10^6$ & 0 \\
\midrule
SIP2Net & 0 & $10^7$ & $5 \times 10^6$ & 0 & 0 & $10^4$ & 0 & 0 \\
\bottomrule
\end{tabular}
}
\end{table}

\subsection{Training Heatmap Types}
\label{sec:training_targets}

Models are trained with three target representations:

\begin{itemize}
\item \textbf{power}: Received-power radio maps induced by power-optimal positions.
\item \textbf{coverage}: Received-power radio maps induced by coverage-optimal positions.
\item \textbf{Gaussian power/coverage}: Synthetic Gaussian heatmaps centered on the optimal positions with $\sigma=3$ pixels. These are used only for the discriminative baselines. The value $\sigma=3$ was chosen as a practical compromise: smaller values make the target too sharp to regress reliably, whereas larger values produce overly diffuse targets and reduce localization precision.
\end{itemize}

The target representation determines what the network is asked to learn. With received-power radio maps, the model predicts a physically meaningful field induced by \textcolor{black}{city map} geometry and an objective-optimal placement, and the final transmitter coordinate is then extracted by argmax. With Gaussian targets, by contrast, the model is asked to reproduce an artificial positional encoding with no direct physical interpretation. Empirically, radio map targets consistently outperform Gaussian targets, and direct coordinate regression proved less suitable still because training collapsed toward trivial center-of-map solutions.

A natural alternative for the coverage-trained indirect model would be to predict the binary coverage map induced by the coverage-optimal placement rather than its received-power radio map. We evaluated this configuration but found that binary coverage maps are substantially harder regression targets: the step-function character of the coverage threshold removes the smooth spatial-gradient structure that makes received-power maps learnable with standard pixel-level losses, and the resulting models failed to converge to competitive placement accuracy. Received-power radio maps therefore serve as the heatmap target for both objectives. They encode the underlying propagation physics continuously and provide the gradient signal needed for effective training, regardless of which objective is used to select the optimal transmitter position whose received-power radio map is used as the label.

\begin{figure*}[tbp]
\centering
\captionsetup{font=scriptsize}
\subfloat{\includegraphics[width=0.16\columnwidth]{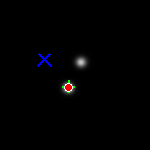}}
\subfloat{\includegraphics[width=0.16\columnwidth]{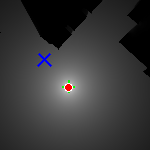}}
\subfloat{\includegraphics[width=0.16\columnwidth]{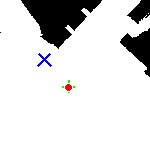}}
\subfloat{\includegraphics[width=0.16\columnwidth]{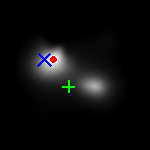}}
\subfloat{\includegraphics[width=0.16\columnwidth]{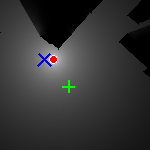}}
\subfloat{\includegraphics[width=0.16\columnwidth]{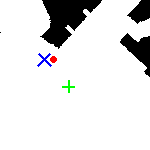}}\\
\subfloat{\includegraphics[width=0.16\columnwidth]{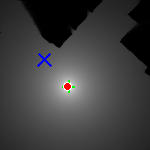}}
\subfloat{\includegraphics[width=0.16\columnwidth]{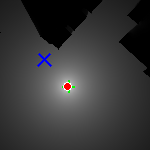}}
\subfloat{\includegraphics[width=0.16\columnwidth]{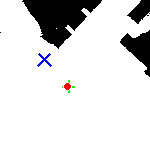}}
\subfloat{\includegraphics[width=0.16\columnwidth]{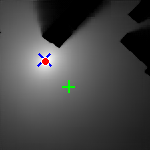}}
\subfloat{\includegraphics[width=0.16\columnwidth]{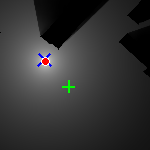}}
\subfloat{\includegraphics[width=0.16\columnwidth]{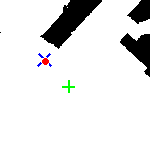}}\\
\subfloat{\includegraphics[width=0.16\columnwidth]{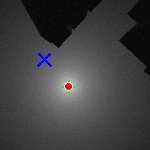}}
\subfloat{\includegraphics[width=0.16\columnwidth]{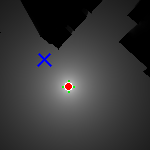}}
\subfloat{\includegraphics[width=0.16\columnwidth]{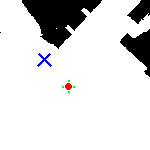}}
\subfloat{\includegraphics[width=0.16\columnwidth]{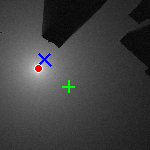}}
\subfloat{\includegraphics[width=0.16\columnwidth]{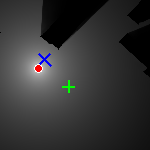}}
\subfloat{\includegraphics[width=0.16\columnwidth]{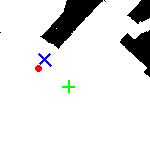}}\\
\subfloat{\includegraphics[width=0.16\columnwidth]{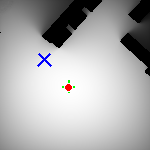}}
\subfloat{\includegraphics[width=0.16\columnwidth]{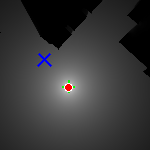}}
\subfloat{\includegraphics[width=0.16\columnwidth]{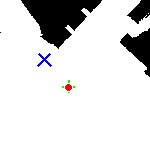}}
\subfloat{\includegraphics[width=0.16\columnwidth]{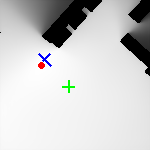}}
\subfloat{\includegraphics[width=0.16\columnwidth]{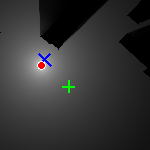}}
\subfloat{\includegraphics[width=0.16\columnwidth]{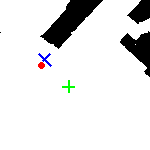}}\\
\subfloat{\includegraphics[width=0.16\columnwidth]{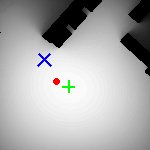}}
\subfloat{\includegraphics[width=0.16\columnwidth]{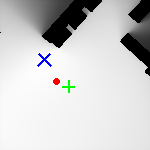}}
\subfloat{\includegraphics[width=0.16\columnwidth]{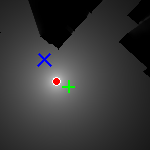}}
\subfloat{\includegraphics[width=0.16\columnwidth]{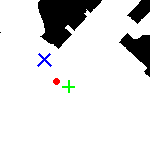}}\\
\subfloat{\includegraphics[width=0.16\columnwidth]{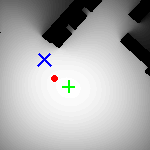}}
\subfloat{\includegraphics[width=0.16\columnwidth]{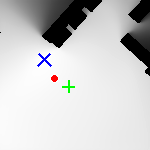}}
\subfloat{\includegraphics[width=0.16\columnwidth]{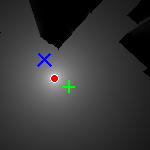}}
\subfloat{\includegraphics[width=0.16\columnwidth]{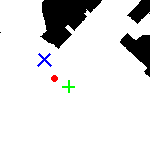}}
\caption{Qualitative comparison on one benchmark \textcolor{black}{city map}
(scenario~167499, the same scenario shown in Fig.~\ref{fig:GTs}).
In all panels, the red dot marks the model-selected
transmitter location, the green cross marks the
ground truth power-optimal location, and the
blue cross marks the ground truth coverage-optimal
location.
\textbf{Rows~1--4} each contain two groups of three panels
(left: power-trained; right: coverage-trained).
Within each group: panel~(i) shows the predicted heatmap
(rows~1--3, indirect formulation) or predicted score-map
(row~4, direct formulation); panel~(ii)~is the SAIPP-Net
reference power map at the model-selected Tx; panel~(iii)
is the corresponding reference coverage map at the same location.
Row~1: indirect, Gaussian target, SIP2Net.
Row~2: indirect, radio map target, DC-Net.
Row~3: indirect, radio map target, Diffusion ($N{=}10$).
Row~4: direct score-map, PMNet, $K{=}1$.
\textbf{Rows~5--6} show four panels each:
(i)~predicted power score-map; (ii)~predicted coverage score-map;
(iii)~SAIPP-Net reference power map at the selected Tx;
(iv)~corresponding reference coverage map.
Row~5: direct score-map, PMNet, union ensemble ($M{=}500$).
Row~6: direct score-map, SIP2Net, minimax ensemble ($K{=}1000$). The objective-space scatter plots in Figs.~\ref{fig:pareto}--\ref{fig:paretoZoom}
show where each selected transmitter falls relative to all feasible candidates
for this city map.}
\label{fig:Results}
\end{figure*}

\subsection{Results}
\label{sec:results}
\begin{table*}[tbp]
\centering
\caption{Diffusion Models --- Power Optimization. Multi-sample best\_power selection reaches 99.97\% power at $N=100$. The table also includes cross-strategy rows where a coverage-trained model is used as a candidate generator under best\_power selection, and rows where a power-trained model is scored by best\_coverage --- these surface favourable coverage--power trade-offs from the same sample pool.}
\label{tab:diffusion_power}
\small
\resizebox{\textwidth}{!}{%
{\color{black}
\begin{tabular}{@{}lcccccccc@{}}
\toprule
N & heatmap & Strategy & Pwr (\%) & Cov (\%) & $\bar{d}$ & ErrPwr (px) & ErrCov (px) & Time (s) \\
\midrule
100 & power & best\_power & 99.97±0.12 & 94.89±6.10 & 5.11 & 0.62±3.135 & 34.02±22.85
 & 20.84 \\
50 & power & best\_power & 99.95±0.15 & 94.37±6.65 & 5.63 & 0.98±3.85 & 35.08±23.42 & 11.94 \\
20 & power & best\_power & 99.91±0.26 & 94.31±6.69 & 5.69 & 1.67±5.42 & 35.12±23.38 & 5.86 \\
10* & power & best\_power & 99.86±0.35 & 94.27±6.72 & 5.73 & 2.32±6.47 & 35.13±23.40 & 3.46 \\
5 & power & best\_power & 99.77±0.55 & 94.18±6.80 & 5.82 & 3.17±7.95 & 35.22±23.35 & 2.09 \\
5 & power & best\_coverage & 99.19±1.82 & 95.31±5.77 & 4.76 & 7.21±13.83 & 32.58±23.67 & 2.09 \\
1** & power & single & 99.12±2.54 & 93.85±7.14 & 6.21 & 7.44±14.84 & 335.61±23.58 & 0.45 \\
10 & power & best\_coverage & 99.05±2.06 & 95.76±5.36 & 4.34 & 8.08±14.72 & 31.54±23.83 & 3.46 \\
100 & coverage & best\_power & 98.83±2.69 & 95.98±4.64 & 4.19 & 8.59±13.22 & 30.41±22.46 & 20.84 \\

\bottomrule
\end{tabular}
}%
}
\\[0.5em]
\footnotesize
N: number of samples; Strategy: selection method; Pwr: received-power \% vs power-optimal GT; Cov: coverage \% vs coverage-optimal GT; $\bar{d}$: $L_2$ distance from ideal (100\%, 100\%); ErrPwr: coordinate error vs power-optimal GT; ErrCov: coordinate error vs coverage-optimal GT; Time: seconds per \textcolor{black}{city map}; *: Shown in Table \ref{tab:computational_summary} and Figures \ref{fig:pareto} and \ref{fig:paretoZoom}; **: Shown in Table \ref{tab:computational_summary}.
\end{table*}

Tables~\ref{tab:diffusion_power}--\ref{tab:fixed_coverage} report heatmap-based results for both optimization objectives. They should be read along two dimensions at once: single-objective accuracy and balanced performance under the $L_2$ metric from Section~\ref{sec:l2_metric}. Several models nearly saturate their training objective while still differing substantially in how much of the alternate objective they preserve. Tables~\ref{tab:diffusion_power}--\ref{tab:fixed_coverage} also include cross-objective rows in which a model trained for one objective is reused as a candidate generator under the other objective's selection rule. For example, a coverage-trained diffusion model paired with best\_power selection becomes an alternative source of candidates for the power objective. These rows make explicit that the same multi-sample pool can be reused across objectives, and they reveal how the training objective interacts with the downstream selection rule.

Fig.~\ref{fig:Results} provides a qualitative comparison on \textcolor{black}{city map} 167499, whose benchmark targets are shown in Fig.~\ref{fig:GTs}. Row~1 shows the Gaussian-target SIP2Net models: both the power-trained and coverage-trained variants correctly localize the corresponding ground-truth optima, but their outputs also exhibit secondary blobs beyond the main peak. Inspection of the ground-truth score-maps shows that these secondary modes do not correspond to genuinely high-performing regions---the score-map for this \textcolor{black}{city map} is unimodal---so these extra blobs are best interpreted as artifacts of regressing toward a fixed-width Gaussian target rather than as evidence of learned near-degenerate solutions. The coverage-trained model also produces a broader predicted region than the power-trained model, which is consistent with the less regular objective landscape of coverage and with the asymmetric trade-off described in Section~\ref{sec:gt_bounds}. Rows~2--6, which show radio map and score-map outputs, are discussed quantitatively below; the objective-space scatter plots in Figs.~\ref{fig:pareto}--\ref{fig:paretoZoom} further contextualize all selected transmitter locations relative to the full feasible candidate set for this \textcolor{black}{city map}.

\subsubsection{Power Optimization Results}

\begin{table*}[tbp]
\centering
\caption{Discriminative Models --- Power Optimization. The strongest configuration (DC-Net, DA-cGAN Stage~2 loss, received-power target) reaches 99.47\% power at $\sim$0.03\,s per \textcolor{black}{city map}. Rows are grouped by target type: received-power radio map (top block), Gaussian power target (middle), Gaussian coverage target (bottom).}
\label{tab:fixed_power}
\small
\resizebox{\textwidth}{!}{%
\begin{tabular}{@{}lcccccccc@{}}
\toprule
Arch. & Loss & Heatmap & Pwr (\%) & Cov (\%) & $\bar{d}$ & ErrPwr (px) & ErrCov (px) & Time (s) \\
\midrule
DC-Net* & DA-cGAN Stage 2 & power & 99.47±1.26 & 94.04±6.83 & 5.98 & 5.17±11.95 & 35.19±23.35 & 0.03 \\
DC-Net & L2+MSGSIM & power & 99.40±1.57 & 93.96±6.95 & 6.07 & 5.32±11.78 & 35.35±23.38 & 0.03 \\
DC-Net & SIP2Net & power & 99.38±1.69 & 94.01±6.89 & 6.02 & 5.15±11.34 & 35.10±23.34 & 0.03 \\
PMNet & SIP2Net & power & 99.37±1.70 & 93.99±6.92 & 6.04 & 5.27±11.68 & 35.26±23.41 & 0.04 \\
DeepXL & DA-cGAN Stage 2 & power & 99.35±1.55 & 94.09±6.78 & 5.95 & 5.32±11.75 & 35.04±23.45 & 0.03 \\
DeepXL & L2+MSGSIM & power & 99.33±1.62 & 94.02±6.85 & 6.02 & 5.36±11.80 & 35.22±23.46 & 0.03 \\
PMNet & DA-cGAN Stage 2 & power & 99.33±2.05 & 93.96±6.95 & 6.08 & 5.88±13.13 & 35.42±23.50 & 0.03 \\
SIP2Net & SIP2Net & power & 99.31±1.97 & 94.00±6.90 & 6.04 & 5.58±12.05 & 35.09±23.33 & 0.04 \\
SIP2Net & DA-cGAN Stage 2 & power & 99.30±2.28 & 94.01±6.93 & 6.03 & 5.62±12.33 & 35.26±23.44 & 0.04 \\
DC-Net & L2 Only & power & 99.29±1.69 & 93.84±7.05 & 6.20 & 5.75±11.87 & 35.52±23.35 & 0.03 \\
PMNet & L2+MSGSIM & power & 99.27±1.77 & 93.85±7.08 & 6.19 & 5.78±12.19 & 35.43±23.42 & 0.04 \\
DeepXL & L2 Only & power & 99.27±1.74 & 93.96±6.90 & 6.08 & 5.53±11.77 & 35.26±23.46 & 0.03 \\
DeepXL & SIP2Net & power & 99.24±2.54 & 93.90±7.14 & 6.15 & 5.49±11.89 & 35.28±23.48 & 0.03 \\
SIP2Net & L2+MSGSIM & power & 99.17±2.29 & 93.71±7.26 & 6.34 & 6.20±12.96 & 35.74±23.38 & 0.04 \\
PMNet & L2 Only & power & 99.16±1.91 & 93.81±7.09 & 6.25 & 6.38±12.89 & 35.48±23.35 & 0.04 \\

SIP2Net*** & L2 Only & Gauss. pwr. & 97.56±11.29 & 92.6±11.68 & 7.79 & 8.11±18.04 & 36.36±24.41 & 0.04 \\

PMNet & L2 Only & Gauss. pwr. & 96.81±13.1 & 91.98±13.26 & 8.63 & 10.18±19.65 & 36.97±24.71 & 0.04 \\

DeepXL & L2 Only & Gauss. pwr. & 95.73±12.0 & 91.95±11.36 & 9.11 & 12.97±19.51 & 36.07±23.8 & 0.03 \\

DC-Net & L2 Only & Gauss. pwr. & 90.83±23.62 & 87.65±20.99 & 15.38 & 15.67±27.93 & 39.42±26.42 & 0.03 \\

PMNet & L2 Only & Gauss. cov. & 86.42±17.44 & 93.61±13.88 & 15.01 & 29.3±23.59 & 27.64±27.71 & 0.04 \\

DeepXL & L2 Only & Gauss. cov. & 85.84±16.69 & 94.00±13.23 & 15.38 & 32.91±24.98 & 26.86±30.9 & 0.03 \\

SIP2Net*** & L2 Only & Gauss. cov. & 84.62±14.92 & 94.55±9.93 & 16.32 & 35.98±25.33 & 26.44±27.20 & 0.04 \\

DC-Net & L2 Only & Gauss. cov. & 77.68±24.3 & 87.92±22.64 & 25.38 & 40.41±27.37 & 36.44±36.8 & 0.03 \\

\bottomrule
\end{tabular}
}
\\[0.5em]
\footnotesize
Architecture: neural network model; Loss: training loss function; Heatmap: ground truth heatmap type; Pwr: received-power \% vs power-optimal GT; Cov: coverage \% vs coverage-optimal GT; $\bar{d}$: $L_2$ distance from ideal (100\%, 100\%); ErrPwr: coordinate error vs power-optimal GT; ErrCov: coordinate error vs coverage-optimal GT; Time: seconds per \textcolor{black}{city map}; *: Shown in Table \ref{tab:computational_summary} and Figures \ref{fig:pareto} and \ref{fig:paretoZoom}; ***: Shown in Figures \ref{fig:pareto} and \ref{fig:paretoZoom}.
\end{table*}

Table~\ref{tab:diffusion_power} shows that diffusion models trained for power optimization approach the exact power optimum and benefit systematically from multi-sampling. The best configuration ($N=100$, best\_power) reaches \textcolor{black}{99.97$\pm$0.12\%} power while maintaining \textcolor{black}{94.89$\pm$6.10\%} coverage, essentially matching the 94.50\% coverage level associated with the power-optimal bound in Section~\ref{sec:gt_bounds}. Repeated sampling together with objective-aware selection therefore improves the single-objective result relative to single-sample diffusion inference, showing that multi-sampling remains useful even when the objective is purely power-oriented.

Table~\ref{tab:fixed_power} shows that discriminative heatmap predictors are also very strong on this objective. The best discriminative configuration---DC-Net with DA-cGAN Stage~2 loss and a received-power target---achieves 99.47$\pm$1.26\% power with competitive balanced performance at real-time inference cost, whereas the Gaussian-target baselines are clearly weaker. Across architectures, the power-trained setting is comparatively favorable because maximizing power already preserves coverage near 94.5\%, so high single-objective performance often translates into reasonably good balanced behavior as well.

Coordinate errors are lower for diffusion models with multi-sampling than for discriminative models, but this comes at substantially higher computational cost. Section~\ref{sec:computational_cost} therefore clarifies an important trade-off that is already visible here: discriminative models are the fastest option, whereas diffusion adds an inference-time search capability that can improve solution quality.

\subsubsection{Coverage Optimization Results}

\begin{table*}[tbp]
\centering
\caption{Diffusion Models --- Coverage Optimization. Multi-sample best\_coverage selection reaches \textcolor{black}{99.76}\% coverage at $N=100$. The table also includes cross-strategy rows where a power-trained model is used as a candidate generator under best\_coverage selection, and rows where a coverage-trained model is scored by best\_power --- these surface favourable coverage--power trade-offs from the same sample pool.}
\label{tab:diffusion_coverage}
\small
\resizebox{\textwidth}{!}{%
{\color{black}
\begin{tabular}{@{}lcccccccc@{}}
\toprule
N & heatmap & Strategy & Cov (\%) & Pwr (\%) & $\bar{d}$ & ErrCov (px) & ErrPwr (px) & Time (s) \\
\midrule
100 & cov & best\_coverage & 99.76±0.49 & 87.01±10.68 & 12.99 & 7.30±14.96 & 34.04±23.02 & 20.84 \\
50 & cov & best\_coverage & 99.61±0.69 & 87.21±10.77 & 12.8 & 8.91±17.12 & 34.71±23.42 & 11.94 \\
20 & cov & best\_coverage & 99.32±1.08 & 87.54±10.62 & 12.48 & 11.26±18.68 & 34.17±23.23 & 5.86 \\
10* & cov & best\_coverage & 99.01±1.43 & 87.84±10.44 & 12.2 & 13.34±19.39 & 33.74±23.09 & 3.46 \\
5 & cov & best\_coverage & 98.54±1.97 & 88.06±10.35 & 12.02 & 16.25±21.28 & 33.17±22.86 & 2.09 \\
100 & pwr & best\_coverage & 96.89±4.31 & 98.59±2.87 & 3.42 & 29.13±23.62 & 9.80±15.35 & 20.84 \\
5 & cov & best\_power & 96.70±3.92 & 94.40±6.68 & 6.50 & 25.43±23.18 & 22.21±19.84 & 2.09 \\
50 & pwr & best\_coverage & 96.55±4.61 & 98.58±2.97 & 3.73 & 29.41±23.91 & 10.54±17.02 & 11.94 \\
10 & cov & best\_power & 96.51±4.12 & 96.01±5.41 & 5.30 & 26.95±23.17 & 18.59±18.93 & 3.46 \\
1** & cov & best\_coverage & 95.85±5.80 & 86.32±12.12 & 14.30 & 26.33±26.28 & 34.71±23.34 & 0.45 \\
\bottomrule
\end{tabular}
}%
}
\\[0.5em]
\footnotesize
N: number of samples; Strategy: selection method; Cov: coverage \% vs coverage-optimal GT; Pwr: received-power \% vs power-optimal GT; $\bar{d}$: $L_2$ distance from ideal (100\%, 100\%); Err\_Cov: coordinate error vs coverage-optimal GT; Err\_Pow: coordinate error vs power-optimal GT; Time: seconds per \textcolor{black}{city map}; *: Shown in Table \ref{tab:computational_summary} and Figures \ref{fig:pareto} and \ref{fig:paretoZoom}; **: Shown in Table \ref{tab:computational_summary}.
\end{table*}

\begin{table*}[tbp]
\centering
\caption{Discriminative Models --- Coverage Optimization. The strongest configuration (DC-Net, DA-cGAN Stage~2 loss, received-power target) reaches 97.65\% coverage at $\sim$0.03\,s per \textcolor{black}{city map}. Rows are grouped by target type: received-power radio map (top block), Gaussian coverage target (middle), Gaussian power target (bottom).}
\label{tab:fixed_coverage}
\small
\resizebox{\textwidth}{!}{%
\begin{tabular}{@{}lcccccccc@{}}
\toprule
Arch. & Loss & Heatmap & Cov (\%) & Pwr (\%) & $\bar{d}$ & ErrCov (px) & ErrPwr (px) & Time (s) \\
\midrule
DC-Net* & DA-cGAN Stage 2 & coverage & 97.65±3.34 & 87.10±11.20 & 13.11 & 17.71±22.35 & 34.44±23.64 & 0.03 \\
SIP2Net & DA-cGAN Stage 2 & coverage & 97.65±3.38 & 87.00±11.02 & 13.21 & 17.85±22.41 & 34.99±23.90 & 0.04 \\
DeepXL & DA-cGAN Stage 2 & coverage & 97.57±3.46 & 86.88±11.30 & 13.34 & 17.76±22.33 & 34.78±23.96 & 0.03 \\
SIP2Net & SIP2Net & coverage & 97.40±3.74 & 88.44±10.59 & 11.85 & 17.59±21.01 & 32.14±22.59 & 0.04 \\
PMNet & DA-cGAN Stage 2 & coverage & 97.40±3.63 & 87.18±11.22 & 13.08 & 18.34±22.24 & 33.95±23.52 & 0.04 \\
DeepXL & L2+MSGSIM & coverage & 97.39±3.67 & 88.47±10.48 & 11.82 & 17.21±20.37 & 32.43±22.67 & 0.03 \\
DC-Net & SIP2Net & coverage & 97.35±3.79 & 87.98±10.66 & 12.31 & 17.90±21.33 & 33.15±22.86 & 0.03 \\
DeepXL & L2 Only & coverage & 97.35±3.71 & 87.66±10.98 & 12.62 & 17.90±21.53 & 33.38±23.51 & 0.03 \\
PMNet & SIP2Net & coverage & 97.33±3.90 & 88.75±10.28 & 11.56 & 17.69±20.86 & 31.90±22.41 & 0.04 \\
DC-Net & L2+MSGSIM & coverage & 97.31±3.79 & 88.41±10.39 & 11.90 & 17.60±20.40 & 32.56±22.40 & 0.03 \\
DC-Net & L2 Only & coverage & 97.26±3.86 & 87.66±10.84 & 12.64 & 18.20±21.46 & 33.36±23.21 & 0.03 \\
SIP2Net & L2+MSGSIM & coverage & 97.17±3.88 & 88.67±10.27 & 11.68 & 18.73±21.23 & 31.68±22.25 & 0.04 \\
PMNet & L2 Only & coverage & 97.17±3.94 & 88.24±10.52 & 12.10 & 19.09±22.35 & 33.44±23.49 & 0.04 \\
DeepXL & SIP2Net & coverage & 97.11±4.14 & 87.72±10.99 & 12.62 & 17.44±20.29 & 33.15±22.60 & 0.03 \\
SIP2Net & L2 Only & coverage & 96.98±4.14 & 87.50±10.93 & 12.86 & 19.50±22.41 & 34.26±23.27 & 0.04 \\

SIP2Net*** & L2 Only & Gauss. cov. & 94.55±9.93 & 84.62±14.92 & 16.32 & 26.44±27.20 & 35.98±25.33 & 0.04 \\

DeepXL & L2 Only & Gauss. cov. & 94.00±13.23 & 85.84±16.69 & 15.38 & 26.86±30.9 & 32.91±24.98 & 0.03 \\

PMNet & L2 Only & Gauss. cov. & 93.61±13.88 & 86.42±17.44 & 15.01 & 27.64±27.71 & 29.3±23.59 & 0.04 \\

SIP2Net*** & L2 Only & Gauss. pwr. & 92.6±11.68 & 97.56±11.29 & 7.79 & 36.36±24.41 & 8.11±18.04 & 0.04 \\

PMNet & L2 Only & Gauss. pwr. & 91.98±13.26 & 96.81±13.1 & 8.63 & 36.97±24.71 & 10.18±19.65 & 0.04 \\

DeepXL & L2 Only & Gauss. pwr. & 91.95±11.36 & 95.73±12.0 & 9.11 & 36.07±23.8 & 12.97±19.51 & 0.03 \\

DC-Net & L2 Only & Gauss. cov. & 87.92±22.64 & 77.68±24.3 & 25.38 & 36.44±36.8 & 40.41±27.37 & 0.03 \\

DC-Net & L2 Only & Gauss. pwr. & 87.65±20.99 & 90.83±23.62 & 15.38 & 39.42±26.42 & 15.67±27.93 & 0.03 \\

\bottomrule
\end{tabular}
}
\\[0.5em]
\footnotesize
Architecture: neural network model; Loss: training loss function; Heatmap: ground truth heatmap type; Cov: coverage \% vs coverage-optimal GT; Pwr: received-power \% vs power-optimal GT; $\bar{d}$: $L_2$ distance from ideal (100\%, 100\%); ErrCov: coordinate error vs coverage-optimal GT; ErrPwr: coordinate error vs power-optimal GT; Time: seconds per \textcolor{black}{city map}; *: Shown in Table \ref{tab:computational_summary} and Figures \ref{fig:pareto} and \ref{fig:paretoZoom}; ***: Shown in Figures \ref{fig:pareto} and \ref{fig:paretoZoom}.
\end{table*}

Table~\ref{tab:diffusion_coverage} shows that coverage optimization is qualitatively different from power optimization. The best diffusion configuration ($N=100$, best\_coverage) reaches \textcolor{black}{99.76$\pm$0.49\%} coverage, but the corresponding power remains around \textcolor{black}{87.01$\pm$10.68\%}, closely matching the much steeper trade-off identified by exhaustive ground-truth analysis. High coverage therefore does not imply proximity to the balanced optimum, because the coverage-optimal region lies intrinsically farther from the ideal point than the power-optimal region.

Table~\ref{tab:fixed_coverage} shows the same overall pattern for discriminative models. The best discriminative configurations remain competitive on coverage, but they do not reach the near-optimal coverage accuracy of diffusion with multi-sampling, and their balanced performance remains limited unless an additional selection mechanism is introduced. The reported power values in the high-80\% range are a direct signature of the underlying geometry of the trade-off.

This asymmetry between the two objectives---power-optimized solutions already lie close to balanced behavior, whereas coverage-optimized solutions do not---explains not only the differing patterns in the power and coverage tables here, but also why the later score-map and dual-candidate selection strategies benefit so strongly from power-side candidates.

\begin{table*}[tbp]
\centering
\caption{Diffusion Models --- Best-$L_2$ Multi-Sample Selection. Top 20 configurations, ranked by $\bar{d}$. Each row corresponds to a diffusion model trained under objective \emph{Opt.} (cov or pwr), with a pool of $N$ samples ranked by the \emph{Strategy} column (best\_l2, best\_cov, or best\_pwr). The best balanced configuration is a coverage-trained model with $N=100$ samples and best\_$L_2$ selection, reaching \textcolor{black}{$\bar{d}=3.1$}.}
\label{tab:bestl2_multisample_diff}
\small
\resizebox{\textwidth}{!}{%
{\color{black}
\begin{tabular}{@{}lcccccccc@{}}
\toprule
 N & Opt. & Strategy & $\bar{d}$ & Cov (\%) & Pwr (\%) & ErrCov (px) & ErrPwr (px) & Time (s) \\
\midrule
100 & cov & best\_l2 & 3.10 & 97.67±2.89 & 97.95±3.13 & 26.09±23.16 & 13.13±15.87 & 20.84 \\
 100 & pwr & best\_l2 & 3.40 & 96.69±4.35 & 99.21±1.71 & 29.94±23.37 & 7.58±12.67 & 20.84 \\
 100 & pwr & best\_cov & 3.50 & 96.86±4.28 & 98.44±3.67 & 29.52±24.28 & 9.81±15.15 & 20.84 \\
 50 & cov & best\_l2 & 3.53 & 97.64±2.89 & 97.37±3.91 & 25.46±23.10 & 15.66±17.91 & 11.94 \\
 50 & pwr & best\_l2 & 3.74 & 96.35±4.74 & 99.16±1.66 & 30.43±23.74 & 8.54±15.07 & 11.94 \\
 50 & pwr & best\_cov & 3.75 & 96.52±4.68 & 98.62±2.89 & 29.56±24.01 & 10.33±16.68 & 11.94 \\
  20 & pwr & best\_cov & 4.05 & 96.11±5.07 & 98.86±2.43 & 30.76±24.01 & 9.19±15.96 & 5.86 \\
 20 & pwr & best\_l2 & 4.07 & 96.00±5.10 & 99.24±1.54 & 31.30±23.78 & 7.79±14.52 & 5.86 \\
 20 & cov & best\_l2 & 4.15 & 97.53±2.95 & 96.66±4.52 & 24.95±22.92 & 17.65±18.40 & 5.86 \\
 100 & cov & best\_pwr & 4.24 & 95.93±4.69 & 98.80±2.65 & 30.56±22.45 & 8.26±13.026 & 20.84 \\
 10 & pwr & best\_cov & 4.35 & 95.75±5.36 & 99.06±2.00 & 31.63±23.78 & 8.03±14.67 & 3.46 \\
 10* & pwr & best\_l2 & 4.37 & 95.68±5.37 & 99.32±1.44 & 31.96±23.66 & 7.02±13.59 & 3.46 \\
 50 & cov & best\_pwr & 4.44 & 95.97±4.65 & 98.14±3.53 & 29.93±23.14 & 11.64±15.92 & 11.94 \\
  20 & cov & best\_pwr & 4.64 & 96.29±4.27 & 97.21±4.32 & 28.27±23.05 & 15.15±17.59 & 5.86 \\
 5 & pwr & best\_cov & 4.73 & 95.34±5.71 & 99.19±1.81 & 32.64±23.75 & 7.12±13.84 & 2.09 \\
5 & pwr & best\_l2 & 4.75 & 95.30±5.71 & 99.35±1.41 & 32.82±23.70 & 6.48±13.03 & 2.09 \\
 10* & cov & best\_l2 & 5.03 & 97.46±3.02 & 95.65±5.54 & 24.26±22.80 & 19.93±19.03 & 3.46 \\
 100 & pwr & best\_pwr & 5.16 & 94.84±6.21 & 99.98±0.08 & 34.14±22.82 & 0.53±2.29 & 20.84 \\
 10 & cov & best\_pwr & 5.27 & 96.52±4.10 & 96.05±5.42 & 26.86±23.06 & 18.40±18.78 & 3.46 \\
 50 & pwr & best\_pwr & 5.63 & 94.37±6.64 & 99.96±0.15 & 35.09±23.42 & 0.99±3.94 & 11.94 \\
\bottomrule
\end{tabular}
}%
}
\\[0.5em]
\footnotesize
$\bar{d}$: $L_2$ distance from ideal point (100\%, 100\%), lower is better; N: number of samples evaluated; Opt: optimization objective (cov=coverage, pwr=power); Strategy: selection strategy (best\_l2, best\_cov, best\_pwr); Cov: coverage \% vs coverage-optimal GT; Pwr: received-power \% vs power-optimal GT; ErrCov: coordinate error vs coverage-optimal GT; ErrPwr: coordinate error vs power-optimal GT; Time: total seconds per \textcolor{black}{city map} including diffusion sampling + SAIPP-Net evaluation; *: Shown in Table \ref{tab:computational_summary} and Figures \ref{fig:pareto} and \ref{fig:paretoZoom}.
\end{table*}

\subsubsection{Balanced Multi-Objective Optimization via Diffusion Multi-Sampling}
\label{sec:multisample}
\begin{figure}[tbp]
    \centering
    \begin{subfigure}[b]{0.48\linewidth}
        \centering
        \includegraphics[width=\linewidth]%
            {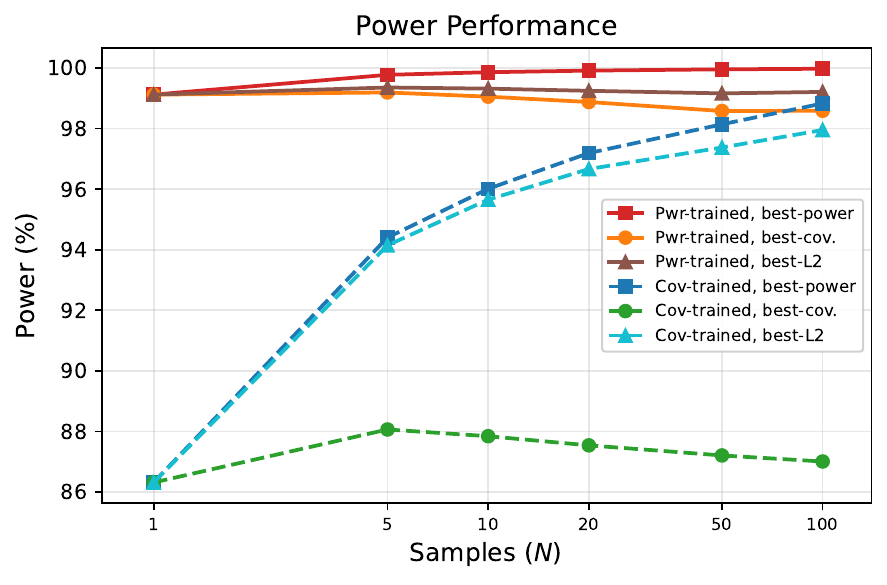}
        \caption{Achieved average power (\%)}
        \label{fig:diffusion_N_power}
    \end{subfigure}
    \begin{subfigure}[b]{0.48\linewidth}
        \centering
        \includegraphics[width=\linewidth]%
            {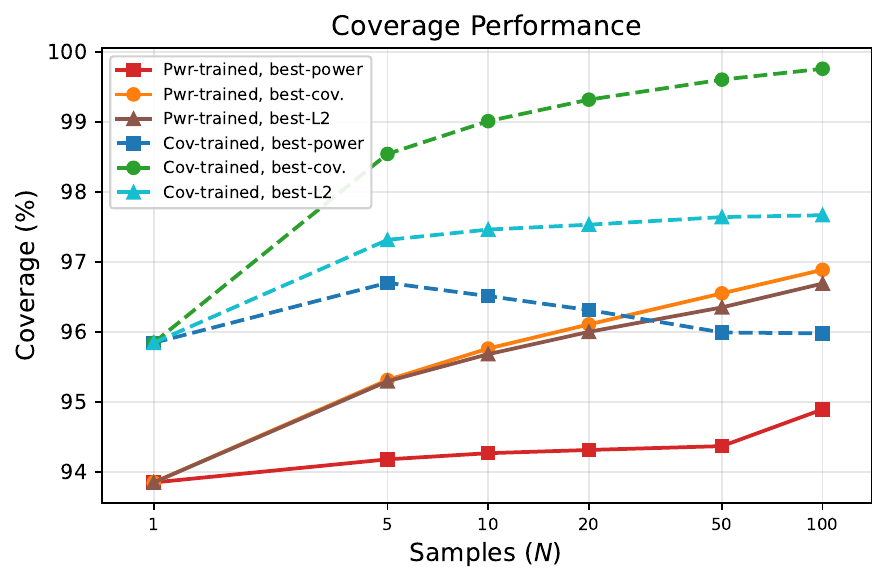}
        \caption{Achieved coverage (\%)}
        \label{fig:diffusion_N_coverage}
    \end{subfigure}
    \\
    \begin{subfigure}[b]{0.48\linewidth}
        \centering
        \includegraphics[width=\linewidth]%
            {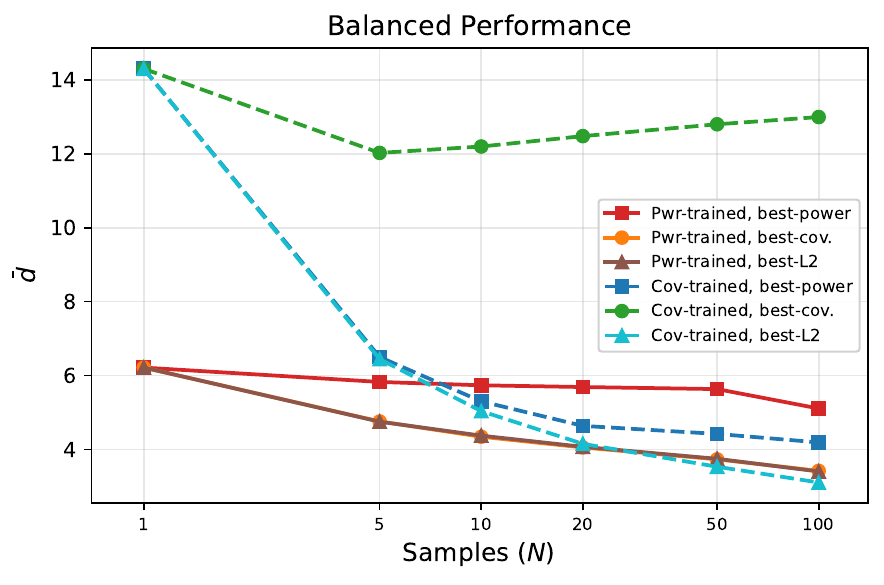}
        \caption{Balanced $\bar{d}$}
        \label{fig:diffusion_N_balanced}
    \end{subfigure}
    \caption{\textcolor{black}{Diffusion heatmap performance vs.\ number of samples $N$. Six curves per panel: two trained diffusion models (\emph{Pwr-trained}, solid; \emph{Cov-trained}, dashed) each evaluated under three selection strategies (\emph{best-power}, \emph{best-coverage}, \emph{best-$L_2$}). At $N=1$ all selection strategies coincide (single sample, no selection). Panels: (a) achieved received-power; (b) achieved coverage; (c) balanced $\bar{d}$ (lower is better).}}
    \label{fig:diffusion_N_sweep}
\end{figure}

Table~\ref{tab:bestl2_multisample_diff} shows that diffusion multi-sampling turns an otherwise single-objective model into a meaningful inference-time search procedure over candidate placements.

The key result is that this search mechanism plays two roles simultaneously. First, it improves single-objective performance: as $N$ increases, the best-power and best-coverage selections improve monotonically because a larger candidate pool raises the probability of drawing a sample close to the corresponding single-objective optimum. Second, it enables post-hoc trade-off selection from the same diffusion sample pool without explicit multi-objective training. A coverage-trained diffusion model generating $N=100$ samples and selecting by best-$L_2$ reaches \textcolor{black}{$\bar{d}=3.1$}, dramatically improving over pure coverage-oriented selection while still preserving \textcolor{black}{97.67\%} coverage. The result is not fully optimal, but it shows that the learned distribution contains high-quality candidates well outside the narrow single-objective mode emphasized during training.

The trend with sample count reinforces this interpretation. As $N$ increases from 10 to 50 to 100, the balanced score improves monotonically, which is consistent with a larger candidate pool probing more of the high-quality tail of the learned radio map distribution. In this sense, diffusion contributes not only a predictor but also a stochastic search layer that can be steered by downstream selection rules.

The success of this post-hoc selection partly reflects the structure of the present objectives. Power-optimal positions already retain $94.50\%$ coverage on average, so the two objectives are not strongly adversarial in this setting. Unlike explicit multi-objective formulations such as the NSGA-II approach of Isabona~et~al.~\cite{isabona_nsga2} or the weighted-reward approach of AutoBS~\cite{autoBS}, the present method does not optimize a predefined Pareto objective set. Favorable trade-offs instead emerge because different samples in the pool score differently on the two objectives, rather than because the trade-off was engineered into the training loss.

This also clarifies the contrast with deterministic discriminative heatmap models. Those models can be highly accurate and much faster, but they emit a single output and therefore do not naturally provide the same search-over-samples mechanism. Diffusion is attractive here not because it dominates every metric, but because it combines physically meaningful radio map generation with a flexible inference-time exploration strategy.

\textcolor{black}{Figure~\ref{fig:diffusion_N_sweep} visualizes how the diffusion sample pool translates into placement performance as $N$ increases. Each curve approaches the per-strategy optimum corresponding to its target objective, and the asymmetric coverage--power trade-off characterized in Section~\ref{sec:gt_bounds} is directly visible in the cross-objective readings. The power-optimal region preserves a relatively high coverage ($94.50\%$ on average across the test set), so the Pwr-trained best-power curve does not show a coverage drop with growing $N$ --- in fact both its power and coverage rise together, since the sample pool concentrates on placements that are simultaneously near power-optimal and near-coverage-good. The Cov-trained best-power curve, by contrast, shows declining coverage with growing $N$. The best-coverage curves show a substantially larger drop in average power with growing $N$, since coverage-optimal placements sacrifice considerably more average power ($86.14\%$) than power-optimal placements sacrifice in coverage. Best-$L_2$ selection lies between the two single-objective extremes and improves with $N$ for both trained models, with the Cov-trained variant providing the best balanced result of the heatmap-based methods at $N=100$ ($\bar{d}=3.17$, Table~\ref{tab:bestl2_multisample_diff}).}

\begin{table*}[tbp]
\centering
\caption{Power Optimization using Score-Maps, $K\in\{1,200\}$. Each architecture predicts a per-pixel score-map; the top-$K$ candidates are evaluated by SAIPP-Net and the highest-scoring one is selected. Rows in the upper block use a power score-map (the training objective matches the test objective); rows in the lower block use a coverage score-map as an alternative candidate source. The strongest configuration is PMNet with $K=200$ from the power score-map, reaching 99.98\% power.}
\label{tab:power_optimization}
\small
\resizebox{\textwidth}{!}{%
\begin{tabular}{@{}lcccccccc@{}}
\toprule
K & Arch. & Score-Map & Pwr (\%) & Cov (\%) & $\bar{d}$ & ErrCov (px) & ErrPwr (px) & Time (s) \\
\midrule
200* & PMNet & pwr & 99.98±0.21 & 94.37±6.69 & 5.63 & 34.97±23.53 & 0.89±6.26 & 0.97 \\
200 & SIP2Net & pwr & 99.98±0.22 & 94.37±6.69 & 5.63 & 34.98±23.54 & 1.00±6.55 & 0.96 \\
200 & DC-Net & pwr & 99.98±0.24 & 94.35±6.70 & 5.65 & 34.98±23.52 & 1.02±6.79 & 0.96 \\
200 & DeepXL & pwr & 99.97±0.28 & 94.32±6.74 & 5.68 & 35.05±23.53 & 1.13±7.02 & 0.96 \\
200* &\color{black} diffusion & pwr &\color{black} 99.95±0.34 &\color{black} 94.33±6.72 &\color{black} 5.67 &\color{black} 35.06±23.56 &\color{black} 1.50±6.74 & 1.38 \\
200 &\color{black} diffusion & cov &\color{black} 95.98±5.80 &\color{black} 97.24±3.35 &\color{black} 4.88 &\color{black} 23.40±21.65 &\color{black} 20.03±21.36 & 1.38 \\
200 & DeepXL & cov & 95.62±5.74 & 97.15±3.43 & 5.22 & 22.81±21.53 & 21.52±21.18 & 0.96\\
200 & DC-Net & cov & 95.28±6.04 & 97.37±3.15 & 5.41 & 21.72±21.21 & 22.54±21.72 & 0.97 \\
200 & PMNet & cov & 94.40±7.13 & 97.47±3.07 & 6.15 & 21.11±21.32 & 24.38±23.39 & 0.97 \\
200 & SIP2Net & cov & 93.74±8.08 & 97.49±3.04 & 6.74 & 21.12±22.16 & 25.52±24.35 & 0.97 \\
1* & PMNet & pwr & 99.63±0.74 & 94.07±6.81 & 5.94 & 35.26±23.52 & 4.90±10.84 & 0.04 \\
1 & SIP2Net & pwr & 99.61±0.76 & 94.06±6.82 & 5.95 & 35.31±23.52 & 5.12±11.10 & 0.04 \\
1 & DC-Net & pwr & 99.57±0.80 & 94.03±6.84 & 5.98 & 35.18±23.45 & 5.23±10.86 & 0.03 \\
1 & DeepXL & pwr & 99.32±1.40 & 93.85±7.03 & 6.18 & 35.57±23.53 & 6.02±10.97 & 0.03 \\
1* &\color{black} diffusion & pwr &\color{black} 99.07±1.40 &\color{black} 93.68±7.19 &\color{black} 6.39 &\color{black} 36.03±23.66 &\color{black} 8.72±12.95 & 0.45 \\
1* & PMNet & cov & 87.39±10.70 & 98.43±2.06 & 12.71 & 17.01±22.61 & 35.37±24.34 & 0.04 \\
1 & DC-Net & cov & 87.02±11.45 & 98.20±2.28 & 13.11 & 18.90±24.57 & 35.77±25.15 & 0.03 \\
1 & SIP2Net & cov & 86.23±11.67 & 98.40±2.09 & 13.87 & 17.78±24.05 & 36.80±25.54 & 0.04 \\
1* &\color{black} diffusion & cov &\color{black} 84.74±12.64 &\color{black} 97.75±3.73 &\color{black} 15.43 &\color{black} 23.56±30.49 &\color{black} 36.92±25.42 & 0.45 \\
1 & DeepXL & cov & 82.61±12.37 & 96.80±3.59 & 17.68 & 23.96±26.73 & 38.33±24.68 & 0.03 \\
\bottomrule
\end{tabular}
}
\\[0.5em]
\footnotesize
K: candidates evaluated with SAIPP-Net; Arch: architecture (diffusion for diffusion models); Score-Map: model training objective; Pwr: received-power \% vs power-optimal GT; Cov: coverage \% vs coverage-optimal GT; $\bar{d}$: $L_2$ distance from ideal (100\%, 100\%); ErrCov/ErrPwr: coordinate errors; Time: seconds per \textcolor{black}{city map}; *: Shown in Table \ref{tab:computational_summary} and Figures \ref{fig:pareto} and \ref{fig:paretoZoom}.
\end{table*}

At the same time, the computational price of this strategy grows with the number of samples. The score-map methods in the next section ultimately achieve better balanced performance at lower runtime, so diffusion multi-sampling should be read as a strong and conceptually interesting indirect alternative rather than as the overall winner on the present benchmark.

\section{Score-Map Approaches: Direct Optimality Scoring}
\label{sec:scoremap}

\subsection{From Indirect to Direct Prediction}

The score-map formulation was introduced by He and Zheng~\cite{avgmap_baseline_paper} for end-to-end prediction of per-pixel average-received-power optimality scores from \textcolor{black}{city maps}. Here we extend that formulation to dual ground-truth labels, i.e., power and coverage, and to multi-objective candidate-selection strategies.

\begin{table*}[tbp]
\centering
\caption{Coverage Optimization using Score-Maps, $K\in\{1,200\}$. Each architecture predicts a per-pixel score-map; the top-$K$ candidates are evaluated by SAIPP-Net and the highest-scoring one is selected. Rows in the upper block use a coverage score-map (the training objective matches the test objective); rows in the lower block use a power score-map as an alternative candidate source. The strongest configuration is SIP2Net (tied with PMNet) with $K=200$ from the coverage score-map, reaching 99.76\% coverage.}
\label{tab:coverage_optimization}
\small
\resizebox{\textwidth}{!}{%
\begin{tabular}{@{}lcccccccc@{}}
\toprule
K & Arch. & Score-Map & Cov (\%) & Pwr (\%) & $\bar{d}$ & ErrCov (px) & ErrPwr (px) & Time (s) \\
\midrule
200* & PMNet & cov & 99.76±0.76 & 87.18±10.70 & 12.82 & 9.90±20.42 & 34.93±23.86 & 0.97 \\
200 & SIP2Net & cov & 99.76±0.77 & 86.51±11.28 & 13.49 & 10.24±21.48 & 35.90±24.70 & 0.97 \\
200 & DC-Net & cov & 99.72±0.85 & 87.52±10.70 & 12.48 & 10.33±20.29 & 34.43±23.62 & 0.96 \\
200* &\color{black} diffusion & cov &\color{black} 99.68±1.03 &\color{black} 86.71±11.21 &\color{black} 13.29 &\color{black} 10.18±20.78 &\color{black} 35.37±24.22 & 1.38 \\
200 & DeepXL & cov & 99.62±1.01 & 87.53±10.77 & 12.48 & 11.49±21.21 & 34.27±23.79 & 0.96 \\
200 &\color{black} diffusion & pwr &\color{black} 96.65±5.29 &\color{black} 98.03±2.06 &\color{black} 3.89 &\color{black} 27.75±24.47 &\color{black} 13.04±12.92 & 1.38 \\
200 & PMNet & pwr & 96.20±5.66 & 98.73±1.66 & 4.00 & 30.02±24.61 & 9.24±10.87 & 0.97 \\
200 & DC-Net & pwr & 96.19±5.68 & 98.71±1.67 & 4.02 & 29.95±24.55 & 9.34±10.93 & 0.96 \\
200 & SIP2Net & pwr & 96.18±5.70 & 98.73±1.66 & 4.02 & 30.08±24.60 & 9.24±10.82 & 0.97 \\
200 & DeepXL & pwr & 96.17±5.70 & 98.64±1.84 & 4.07 & 30.02±24.56 & 9.46±11.18 & 0.96 \\

1* & PMNet & cov & 98.43±2.06 & 87.39±10.70 & 12.71 & 17.01±22.61 & 35.37±24.34 & 0.04 \\
1 & SIP2Net & cov & 98.40±2.09 & 86.23±11.67 & 13.87 & 17.78±24.05 & 36.80±25.54 & 0.04 \\
1 & DC-Net & cov & 98.20±2.28 & 87.02±11.45 & 13.11 & 18.90±24.57 & 35.77±25.15 & 0.03 \\
1* &\color{black} diffusion & cov &\color{black} 97.75±3.73 &\color{black} 84.74±12.64 &\color{black} 15.43 &\color{black} 23.56±30.49 &\color{black} 38.00±26.00 & 0.45 \\
1 & DeepXL & cov & 96.80±3.59 & 82.61±12.37 & 17.68 & 23.96±26.73 & 38.33±24.68 & 0.03 \\
1* & PMNet & pwr & 94.07±6.81 & 99.63±0.74 & 5.94 & 35.26±23.52 & 4.90±10.84 & 0.04 \\
1 & SIP2Net & pwr & 94.06±6.82 & 99.61±0.76 & 5.95 & 35.31±23.52 & 5.12±11.10 & 0.04 \\
1 & DC-Net & pwr & 94.03±6.84 & 99.57±0.80 & 5.98 & 35.18±23.45 & 5.23±10.86 & 0.03 \\
1 & DeepXL & pwr & 93.85±7.03 & 99.32±1.40 & 6.18 & 35.57±23.53 & 6.02±10.97 & 0.03 \\
1* &\color{black} diffusion & pwr &\color{black} 93.68±7.19 &\color{black} 99.07±1.40 &\color{black} 6.39 &\color{black} 36.03±23.66 &\color{black} 8.72±12.95 & 0.45 \\
\bottomrule
\end{tabular}
}
\\[0.5em]
\footnotesize
K: candidates evaluated with SAIPP-Net; Arch: architecture (diffusion for diffusion models); Score-Map: model training objective (cov=coverage, pwr=power); Cov: coverage \% vs coverage-optimal GT; Pwr: received-power \% vs power-optimal GT; $\bar{d}$: $L_2$ distance from ideal (100\%, 100\%); ErrCov/ErrPwr: coordinate errors; Time: seconds per \textcolor{black}{city map}; *: Shown in Table \ref{tab:computational_summary} and Figures \ref{fig:pareto} and \ref{fig:paretoZoom}.
\end{table*}

Each score-map pixel in the $150\times150$ feasible region stores the spatially averaged received-power or coverage defined in~\eqref{eq:avgPwr} and~\eqref{eq:avgCov}. Because these maps are built from exhaustive SAIPP-Net evaluation, they preserve the full objective landscape over candidate locations rather than only the best point. Training on score-maps therefore amounts to learning the optimization surface itself.

Given a predicted score-map, transmitter selection can be performed in several ways: by direct pixel-wise argmax, by evaluating the top-$K$ predicted candidates with SAIPP-Net and selecting the best, or---when both a power score-map and a coverage score-map are available---by combining the two predictions through joint ranking or candidate pooling. These selection rules are studied in the subsections below.

\textcolor{black}{The score-map family in this section uses the same diffusion and discriminative model architectures introduced in Section~4 for the heatmap family; only the training target changes (per-pixel objective scores rather than radio maps), so that the heatmap and score-map results are produced by matched models trained under the matched protocol.}

\subsection{Single Score-Map Results}

Tables~\ref{tab:power_optimization} and~\ref{tab:coverage_optimization} report single-objective score-map performance for $K \in \{1,200\}$ using the DA-cGAN Stage~2 loss across all five model architectures.

These results show that score-map prediction is already highly effective for the single-transmitter task. When trained for a given objective, the predicted ranking places near-optimal transmitter pixels near the top of the list, and SAIPP-Net evaluation over the top-$K$ candidates closes most of the remaining gap.

The more informative question, however, is what happens when these same candidate lists are judged by balanced rather than single-objective criteria. Tables~\ref{tab:power_scoremap_bestl2} and~\ref{tab:coverage_scoremap_bestl2} therefore apply best-$L_2$ selection to the same top-$K$ sets, replacing objective-maximizing selection with the candidate that lies closest to the ideal point $(100\%,100\%)$.

\begin{table}[!htbp]
\centering
\caption{Power Score-Map with Best-$L_2$ Selection, $K\in\{500, 1000, 2000\}$. Top-$K$ candidates from the predicted power score-map are evaluated by SAIPP-Net; the candidate minimising per-instance $L_2$ distance from the ideal point $(100\%,100\%)$ is selected. The power score-map alone already yields a strong balanced result (\textcolor{black}{$\bar{d}=2.68$} at $K=2000$) because the power-optimal region of the landscape is itself near-balanced.}
\label{tab:power_scoremap_bestl2}
\small
\begin{tabular}{@{}lcccccc@{}}
\toprule
Arch. & K & Cov (\%) & Pwr (\%) & $\bar{d}$ & Time (s) \\
\midrule
\color{black} diffusion& 2000 &\color{black} 97.70±3.18 &\color{black} 98.62±2.05 &\color{black} 2.68 & 6.89 \\
PMNet & 2000 & 97.69±3.32 & 98.63±1.99 & 2.69 & 6.48 \\
SIP2Net & 2000 & 97.66±3.36 & 98.64±1.96 & 2.70 & 6.48 \\
DC-Net & 2000 & 97.66±3.35 & 98.64±1.98 & 2.71 & 6.47 \\
DeepXL & 2000 & 97.66±3.33 & 98.63±2.00 & 2.71 & 6.47 \\
\midrule
\color{black} diffusion& 1000 &\color{black} 97.49±3.71 &\color{black} 98.72±1.83 &\color{black} 2.82 & 3.67 \\
PMNet & 1000 & 97.38±4.03 & 98.77±1.72 & 2.89 & 3.26 \\
DeepXL & 1000 & 97.37±4.02 & 98.79±1.69 & 2.90 & 3.25 \\
SIP2Net & 1000 & 97.36±4.06 & 98.80±1.64 & 2.90 & 3.26 \\
DC-Net & 1000 & 97.34±4.08 & 98.80±1.65 & 2.92 & 3.25 \\
\midrule
\color{black} diffusion & 500 &\color{black} 97.14±4.23 &\color{black} 98.89±1.46 &\color{black} 3.07 & 2.31 \\
PMNet & 500 & 96.97±4.60 & 99.01±1.31 & 3.19 & 1.90 \\
DC-Net & 500 & 96.95±4.64 & 99.02±1.24 & 3.20 & 1.89 \\
SIP2Net & 500 & 96.93±4.65 & 99.02±1.29 & 3.22 & 1.90 \\
DeepXL & 500 & 96.93±4.66 & 99.02±1.23 & 3.23 & 1.89 \\
\bottomrule
\end{tabular}
\\[0.5em]
\footnotesize
Arch: model architecture; K: number of power score-map candidates evaluated by SAIPP-Net; Cov: coverage \% vs coverage-optimal GT; Pwr: received-power \% vs power-optimal GT; $\bar{d}$: $L_2$ Euclidean distance from ideal point $(100\%, 100\%)$ computed from mean Cov and Pwr; Time: seconds per \textcolor{black}{city map} including score-map prediction + SAIPP-Net evaluation.
\end{table}

\begin{table}[!htbp]
\centering
\caption{Coverage Score-Map with Best-$L_2$ Selection, $K\in\{500, 1000, 2000\}$. Top-$K$ candidates from the predicted coverage score-map are evaluated by SAIPP-Net; the candidate minimising per-instance $L_2$ distance from the ideal point $(100\%,100\%)$ is selected. Balanced performance improves as $K$ grows and the shortlist spills beyond the (more penalty-prone) coverage-optimal region into better trade-off territory.}
\label{tab:coverage_scoremap_bestl2}
\small
\begin{tabular}{@{}lcccccc@{}}
\toprule
Arch. & K & Cov (\%) & Pwr (\%) & $\bar{d}$ & Time (s) \\
\midrule
\color{black} diffusion& 2000 &\color{black} 97.84±2.81 &\color{black} 98.34±2.38 &\color{black} 2.72 & 6.89 \\
DeepXL & 2000 & 97.85±2.80 & 98.17±2.49 & 2.82 & 6.47 \\
DC-Net & 2000 & 97.86±2.80 & 98.08±2.51 & 2.87 & 6.47 \\
\color{black} diffusion& 1000 &\color{black} 97.91±2.73 &\color{black} 97.88±3.25 &\color{black} 2.98 & 3.67 \\
SIP2Net & 2000 & 97.86±2.80 & 97.85±2.92 & 3.03 & 6.48 \\
DeepXL & 1000 & 97.90±2.77 & 97.61±3.22 & 3.18 & 3.25 \\
PMNet & 2000 & 97.87±2.80 & 97.54±3.68 & 3.26 & 6.48 \\
DC-Net & 1000 & 97.91±2.76 & 97.45±3.29 & 3.30 & 3.25 \\
\color{black} diffusion& 500 &\color{black} 97.98±2.63 &\color{black} 97.14±4.38 &\color{black} 3.50 & 2.31 \\
SIP2Net & 1000 & 97.94±2.70 & 96.89±4.45 & 3.73 & 3.26 \\
DeepXL & 500 & 98.03±2.54 & 96.73±4.22 & 3.81 & 1.89 \\
PMNet & 1000 & 97.95±2.67 & 96.77±4.64 & 3.82 & 3.26 \\
DC-Net & 500 & 98.09±2.51 & 96.47±4.43 & 4.02 & 1.89 \\
PMNet & 500 & 98.08±2.57 & 95.69±5.85 & 4.72 & 1.90 \\
SIP2Net & 500 & 98.13±2.44 & 95.51±6.22 & 4.87 & 1.90 \\
\bottomrule
\end{tabular}
\\[0.5em]
\footnotesize
Arch: model architecture; K: number of coverage score-map candidates evaluated by SAIPP-Net; Cov: coverage \% vs coverage-optimal GT; Pwr: received-power \% vs power-optimal GT; $\bar{d}$: $L_2$ Euclidean distance from ideal point $(100\%, 100\%)$ computed from mean Cov and Pwr; Time: seconds per \textcolor{black}{city map} including score-map prediction + SAIPP-Net evaluation.
\end{table}

The contrast between power and coverage score-maps is instructive. For the power score-map, best-$L_2$ selection is already very competitive because the power-optimal region is itself close to balanced. For the coverage score-map, the top candidates remain concentrated in a region that lies intrinsically farther from the balanced optimum, so performance improves mainly when larger $K$ allows the shortlist to spill into better trade-off territory. The balanced objective is therefore not located midway between the two single-objective optima; it lies much closer to the power side of the landscape. Score-map prediction makes this geometry directly visible by exposing the objective surface and allowing the same candidate pool to be interrogated under different downstream criteria.

\subsection{Dual Score-Map Extensions}
\label{sec:avgmap_multiobjective}

The single-score-map results suggest that balanced optimization should not rely on either objective view alone. The natural next step is therefore to combine the power and coverage score-maps so that candidate generation reflects both sides of the trade-off before SAIPP-Net performs the final evaluation. We study two such strategies.

\subsubsection{Minimax Ranking of Candidates}

The first strategy combines both score-maps through a \emph{minimax} ranking rule. Let $r_{\text{Cov}}(i,j)$ and $r_{\text{P}}(i,j)$ denote the rank of pixel $(i,j)$ when free pixels are sorted by descending predicted coverage and power score, respectively. For each candidate pixel, its combined rank is
\begin{equation}
r_{\text{max}}(i,j) = \max\!\left(r_{\text{Cov}}(i,j),\; r_{\text{P}}(i,j)\right)
\label{eq:maxRank}
\end{equation}
The top-$K$ pixels with the lowest $r_{\text{max}}$ are then evaluated by SAIPP-Net, and the final location is selected by best-$L_2$.

Per-map ranks $r_{\text{Cov}}$ and $r_{\text{P}}$ are unique per pixel: equal predicted scores are broken deterministically by the order in which free pixels are enumerated, i.e., raster-scan order within the feasible 150$\times$150 region. The minimax score $r_{\text{max}}$ is therefore unique for nearly all pixels; on the rare occasions when two pixels share the same $r_{\text{max}}$, the same raster-scan order breaks the tie.

Intuitively, minimax favors candidates that are simultaneously strong under both predicted objectives. It is therefore a principled way to bias the evaluation budget toward the intersection of the two high-confidence regions.

\begin{table*}[tbp]
\centering
\caption{\small Dual Score-Map --- Minimax Ranking with Best-$L_2$ Selection, $K\in\{200, 500, 1000, 2000\}$. Each pixel is ranked within both the power score-map and the coverage score-map; candidates are then sorted by the worse of the two ranks, $r_{\max}$ (eq.~\ref{eq:maxRank}). The top-$K$ minimax candidates are evaluated by SAIPP-Net, and the one minimising per-instance $L_2$ distance from the ideal point is selected. Performance improves consistently with $K$, approaching the exhaustive-search balanced optimum of $\bar{d}=2.60$; at $K=2000$ the strongest configuration reaches $\bar{d}=\textcolor{black}{2.62}$.}
\label{tab:multiobjective_minimax_avgmap}
\small
\begin{tabular}{@{}lccccc@{}}
\toprule
Arch. & K & $\bar{d}$ & Cov (\%) & Pwr (\%) & Time (s) \\
\midrule
\color{black} diffusion& 2000 &\color{black} 2.62 &\color{black} 97.83±2.83 &\color{black} 98.54±2.24 & 6.89 \\
DeepXL & 2000 & 2.64 & 97.80±2.94 & 98.55±2.17 & 6.47 \\
SIP2Net & 2000 & 2.66 & 97.82±2.89 & 98.48±2.16 & 6.48 \\
DC-Net & 2000 & 2.66 & 97.82±2.89 & 98.48±2.15 & 6.47 \\
\color{black} diffusion**& 1000 &\color{black} 2.67 &\color{black} 97.77±2.97 &\color{black} 98.53±2.19 & 3.67 \\
DeepXL & 1000 & 2.69 & 97.73±3.13 & 98.55±2.06 & 3.25 \\
PMNet & 2000 & 2.70 & 97.83±2.84 & 98.39±2.23 & 6.48 \\
SIP2Net* & 1000 & 2.72 & 97.77±3.05 & 98.44±2.08 & 3.26 \\
DC-Net & 1000 & 2.73 & 97.74±3.12 & 98.47±2.03 & 3.25 \\
PMNet & 1000 & 2.76 & 97.78±3.03 & 98.36±2.14 & 3.26 \\
\color{black} diffusion& 500 &\color{black} 2.91 &\color{black} 97.56±3.45 &\color{black} 98.42±2.19 & 2.31 \\
DeepXL & 500 & 2.95 & 97.51±3.57 & 98.41±2.09 & 1.89 \\
DC-Net & 500 & 2.99 & 97.55±3.52 & 98.29±2.08 & 1.89 \\
SIP2Net & 500 & 2.99 & 97.56±3.51 & 98.27±2.10 & 1.90 \\
\color{black} diffusion& 200 &\color{black} 3.06 &\color{black} 97.41±3.70 &\color{black} 98.37±2.14 & 1.38 \\
PMNet & 500 & 3.03 & 97.55±3.54 & 98.21±2.14 & 1.90 \\
DeepXL & 200 & 3.15 & 97.31±3.86 & 98.36±1.98 & 0.96 \\
DC-Net & 200 & 3.17 & 97.37±3.81 & 98.24±1.98 & 0.96 \\
SIP2Net & 200 & 3.18 & 97.39±3.80 & 98.18±2.03 & 0.97 \\
PMNet & 200 & 3.20 & 97.38±3.83 & 98.17±2.02 & 0.97 \\
\bottomrule
\end{tabular}
\\[0.5em]
\footnotesize
Arch: model architecture (diffusion = diffusion score-map model); K: number of minimax-ranked candidates evaluated by SAIPP-Net; $\bar{d}$: $L_2$ Euclidean distance from ideal point $(100\%, 100\%)$ computed from mean Cov and Pwr; Cov: coverage \% vs coverage-optimal GT; Pwr: received-power \% vs power-optimal GT; Time: total seconds per \textcolor{black}{city map} including dual score-map prediction + SAIPP-Net evaluation of K candidates; *: Shown in Table \ref{tab:computational_summary} and Figures \ref{fig:pareto} and \ref{fig:paretoZoom}; **: Shown in Table \ref{tab:computational_summary}.
\end{table*}

Table~\ref{tab:multiobjective_minimax_avgmap} reports results for $K \in \{200, 500, 1000, 2000\}$. Performance improves consistently with $K$, approaching the theoretical optimum of $\bar{d}=2.60$ established by exhaustive ground-truth analysis.

\subsubsection{Single-Objective Union}

\begin{table*}[tbp]
\centering
\caption{Dual Score-Map --- Single-Objective Union with Best-$L_2$ Selection, $M\in\{500, 1000, 2000\}$. The top-$M$ candidates from each score-map (power and coverage) are pooled into a single candidate set, evaluated by SAIPP-Net, and the one minimising per-instance $L_2$ distance from the ideal point is selected. This simple pooling matches the exhaustive-search balanced optimum $\bar{d}=2.60$ at $M=2000$ across all architectures, and remains within $\bar{d}\le 2.65$ down to $M=500$.}
\label{tab:singleobj_union_avgmap}
\small
\resizebox{\textwidth}{!}{%
\begin{tabular}{@{}lcccccccc@{}}
\toprule
Arch. & M & $\bar{d}$ & Cov (\%) & Pwr (\%) & Union & Overlap (\%) & Time (s) \\
\midrule
DeepXL & 2000 & 2.60 & 97.84±2.81 & 98.55±2.23 & 3028 & 48.6 & 11.16 \\
PMNet & 2000 & 2.60 & 97.84±2.81 & 98.55±2.23 & 3043 & 47.9 & 11.17 \\
SIP2Net & 2000 & 2.60 & 97.84±2.81 & 98.55±2.23 & 3052 & 47.4 & 11.17 \\
DC-Net & 2000 & 2.60 & 97.84±2.82 & 98.55±2.23 & 3036 & 48.2 & 11.16 \\
\color{black} diffusion& 2000 &\color{black} 2.60 &\color{black} 97.84±2.82 &\color{black} 98.55±2.23 &\color{black} 3087 &\color{black} 45.65&\color{black} 11.6 \\
DeepXL & 1000 & 2.60 & 97.83±2.85 & 98.56±2.27 & 1667 & 33.3 & 6.29 \\
SIP2Net & 1000 & 2.61 & 97.83±2.86 & 98.56±2.25 & 1670 & 33.0 & 6.30 \\
DC-Net & 1000 & 2.61 & 97.83±2.85 & 98.56±2.26 & 1664 & 33.6 & 6.29 \\
PMNet & 1000 & 2.61 & 97.82±2.88 & 98.57±2.23 & 1665 & 33.5 & 6.30 \\
\color{black} diffusion & 1000 &\color{black} 2.61 &\color{black} 97.81±2.91 &\color{black} 98.58±2.18 &\color{black} 1688 &\color{black} 31.2 &\color{black} 6.72 \\
PMNet* & 500 & 2.62 & 97.77±2.97 & 98.63±2.20 & 888 & 22.3 & 3.28 \\
DC-Net & 500 & 2.62 & 97.78±2.91 & 98.60±2.29 & 888 & 22.3 & 3.27 \\
SIP2Net & 500 & 2.62 & 97.76±2.96 & 98.63±2.22 & 891 & 21.8 & 3.28 \\
DeepXL & 500 & 2.62 & 97.79±2.88 & 98.58±2.34 & 891 & 21.8 & 3.27 \\
\color{black} diffusion**& 500 &\color{black} 2.65 &\color{black} 97.81±2.84 &\color{black} 98.57±2.33 &\color{black} 901 &\color{black} 19.8 &\color{black} 3.71 \\
\bottomrule
\end{tabular}
}
\\[0.5em]
\footnotesize
Arch: model architecture; M: number of top candidates per score-map; $\bar{d}$: $L_2$ Euclidean distance from ideal point $(100\%, 100\%)$ computed from mean Cov and Pwr; Cov: coverage \% vs coverage-optimal GT; Pwr: received-power \% vs power-optimal GT; Union: mean number of unique candidates in the combined set; Overlap: fraction of the top-M candidate set shared between the two score-maps, $\text{Overlap} = (2M - |\mathcal{C}_\text{pow} \cup \mathcal{C}_\text{cov}|)/M \times 100\%$; Time: total seconds per \textcolor{black}{city map} including dual score-map prediction + SAIPP-Net evaluation; *: Shown in Table \ref{tab:computational_summary} and Figures \ref{fig:pareto} and \ref{fig:paretoZoom}; **: Shown in Table \ref{tab:computational_summary}.
\end{table*}

The second strategy pools the top-$M$ candidates from the power score-map and the top-$M$ candidates from the coverage score-map into a union set, evaluates all of them with SAIPP-Net, and selects the final location by best-$L_2$. This strategy is simpler than minimax, but empirically very strong because it preserves high-quality candidates from both objective-specific models without forcing them through a stricter joint pre-ranking rule.

Table~\ref{tab:singleobj_union_avgmap} reports results for $M \in \{500, 1000, 2000\}$. Across models, the union strategy reaches near-optimal and often optimal balanced performance, including $\bar{d}=2.60$ in the best configurations. Direct dual-score-map prediction combined with this simple candidate-pooling rule is therefore sufficient to match the balanced optimum established by exhaustive search.

\subsubsection{Comparison}
\begin{figure}[tbp]
    \centering
    \begin{subfigure}[b]{0.48\linewidth}
        \centering
        \includegraphics[width=\linewidth]%
            {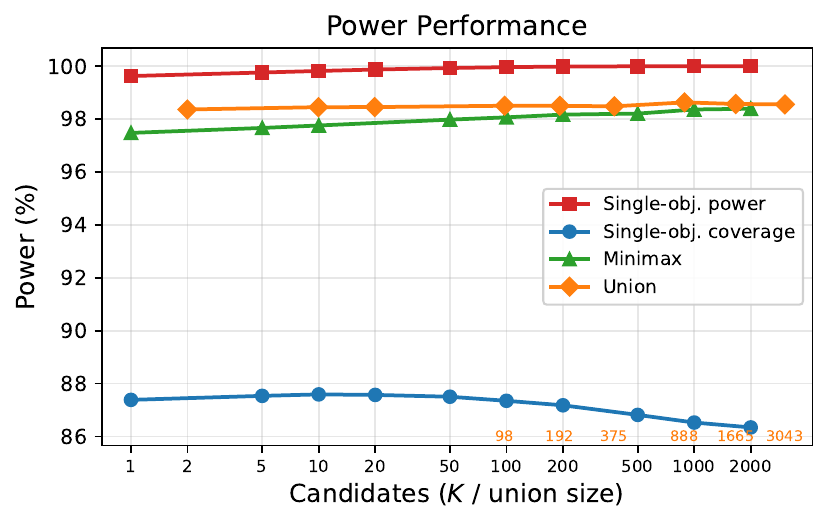}
        \caption{Achieved average power (\%)}
        \label{fig:pmnet_K_power_fair}
    \end{subfigure}
    \begin{subfigure}[b]{0.48\linewidth}
        \centering
        \includegraphics[width=\linewidth]%
            {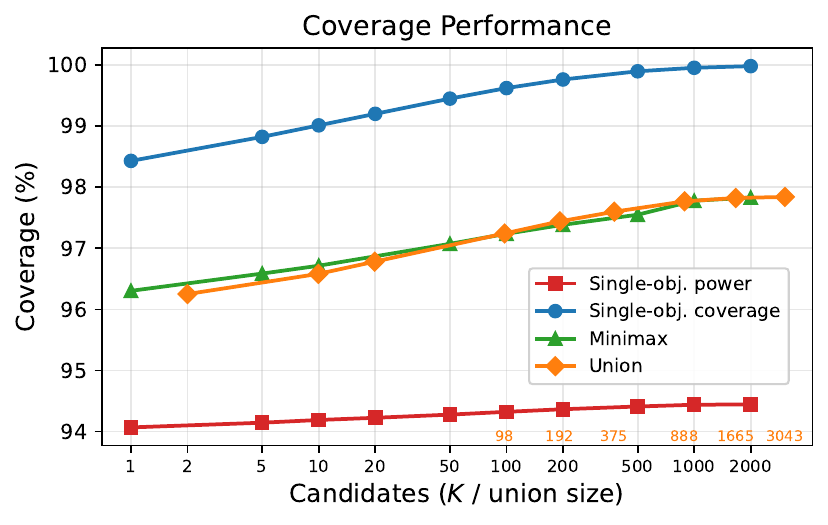}
        \caption{Achieved coverage (\%)}
        \label{fig:pmnet_K_coverage_fair}
    \end{subfigure}
    \\
    \begin{subfigure}[b]{0.48\linewidth}
        \centering
        \includegraphics[width=\linewidth]%
            {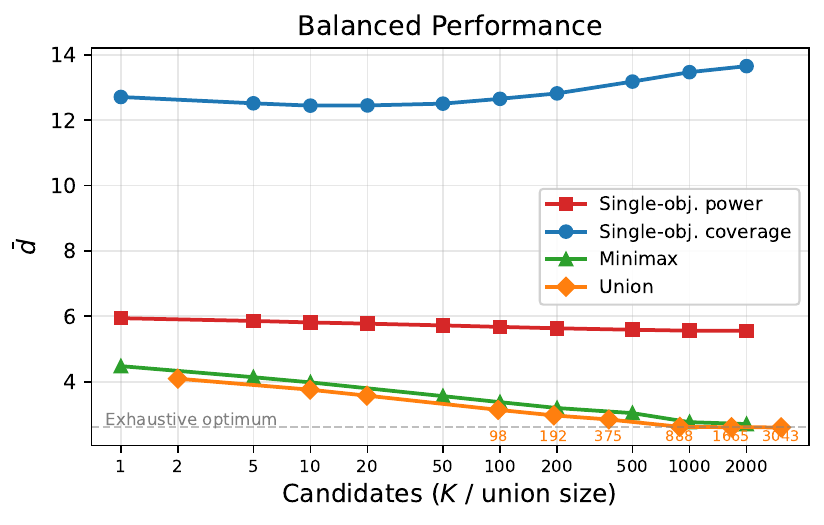}
        \caption{Balanced $\bar{d}$ (dashed: exhaustive optimum $\bar{d}^*=2.60$)}
        \label{fig:pmnet_K_balanced_fair}
    \end{subfigure}
    \caption{\textcolor{black}{Score-map performance vs.\ candidate budget for one representative model (PMNet with DA-cGAN Stage~2 loss), for the four selection strategies introduced in Sections~5.2 and~5.3.1--5.3.2 (\emph{Single-obj.\ power}, \emph{Single-obj.\ coverage}, \emph{Minimax}, \emph{Union}). The x-axis is the actual number of SAIPP-Net evaluations per city map --- $K$ for the single-objective and minimax strategies, and the mean union size (between $K$ and $2K$ after overlap removal) for the union strategy. Panels: (a) achieved power $P_n$; (b) achieved coverage $C_n$; (c) balanced $\bar{d}$, with the exhaustive-search benchmark optimum $\bar{d}^*=2.60$ marked.}}
    \label{fig:pmnet_K_sweep_fair}
\end{figure}

Both dual-map strategies improve substantially over either single score-map used in isolation, and at matched or similar evaluation budgets the union strategy consistently outperforms minimax.

The trade-off geometry is asymmetric: the power-optimal region is already close to the balanced optimum, whereas the coverage-optimal region lies farther away. Minimax can therefore over-penalize candidates near the power-optimal region because of their weaker coverage rank, even when those candidates are in fact close to the best balanced solution. The union strategy avoids this failure mode by allowing the power model's candidates to remain competitive on their own merit while still injecting alternatives from the coverage model when needed. In practice, this simple pooling mechanism proves more faithful to the geometry of the true trade-off than the stricter minimax pre-ranking.

\textcolor{black}{Figure~\ref{fig:pmnet_K_sweep_fair} extends the comparison above across the full candidate-budget range for the representative PMNet configuration, plotted on a cost-fair x-axis (actual SAIPP-Net evaluations per city map rather than the per-map shortlist size, which differs between Minimax and Union). Single-objective strategies saturate tightly on their target metric but remain far from the balanced optimum at every budget. Both dual-map strategies improve monotonically with budget on $\bar{d}$ and approach $\bar{d}^*=2.60$ from above; at matched evaluation cost, Union outperforms Minimax across the budget range, consistent with the asymmetric-geometry argument in the preceding paragraph.}

\begin{figure*}[tbp]
\centering
{\includegraphics[width=0.75\textwidth]{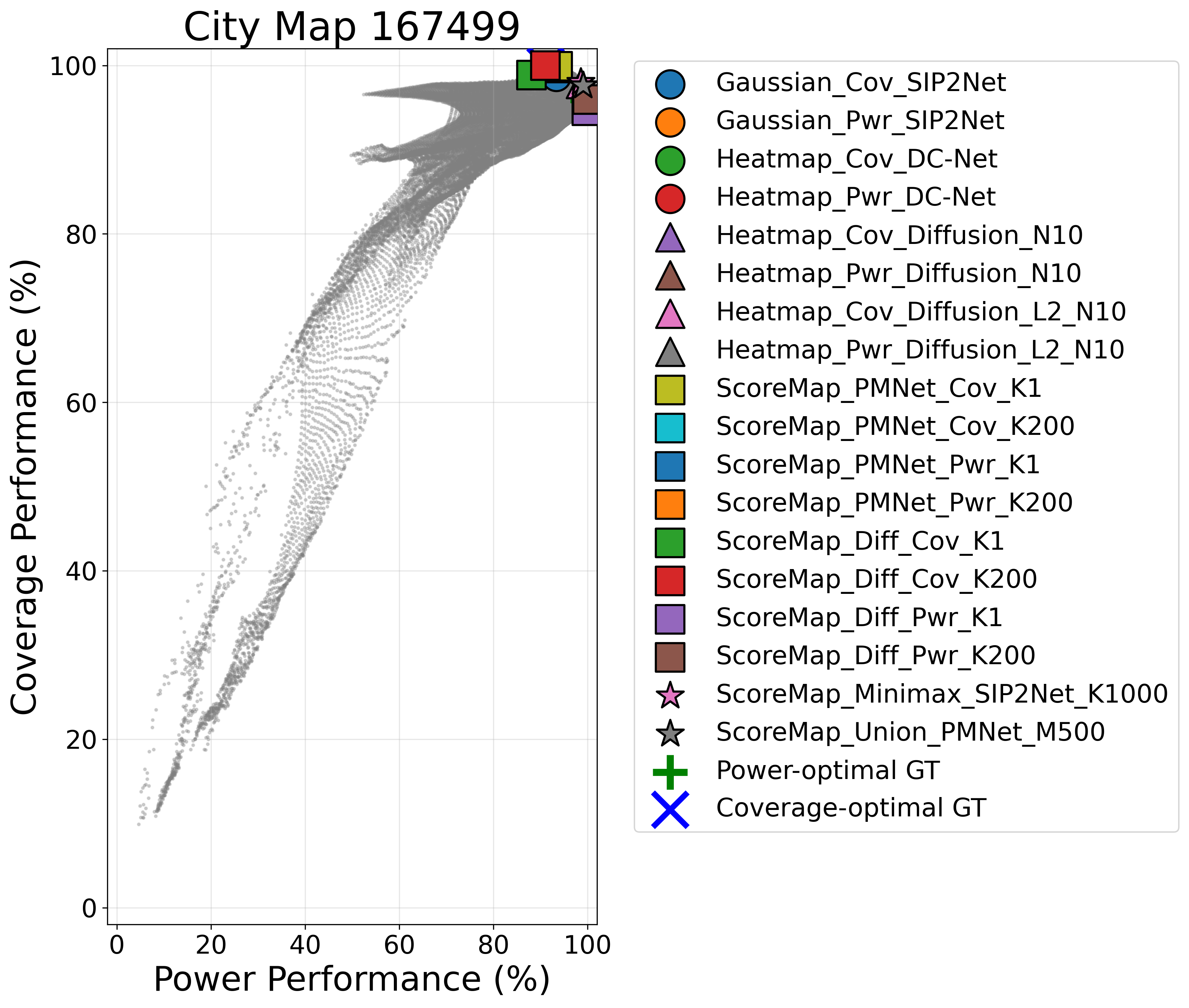}}
\caption{Objective-space scatter plot for \textcolor{black}{city map}~167499 (the benchmark scenario used in Figs.~\ref{fig:GTs} and~\ref{fig:Results}). Each point represents one feasible candidate transmitter location, evaluated exhaustively on both coverage (\%) and average received-power (\%). The upper-right boundary of the point cloud forms the Pareto front of non-dominated solutions. Coloured markers indicate achieved coverage (\%) and average received-power (\%) by each evaluated model.}
\label{fig:pareto}
\end{figure*}

\begin{figure*}[tbp]
\centering
{\includegraphics[width=0.75\textwidth]{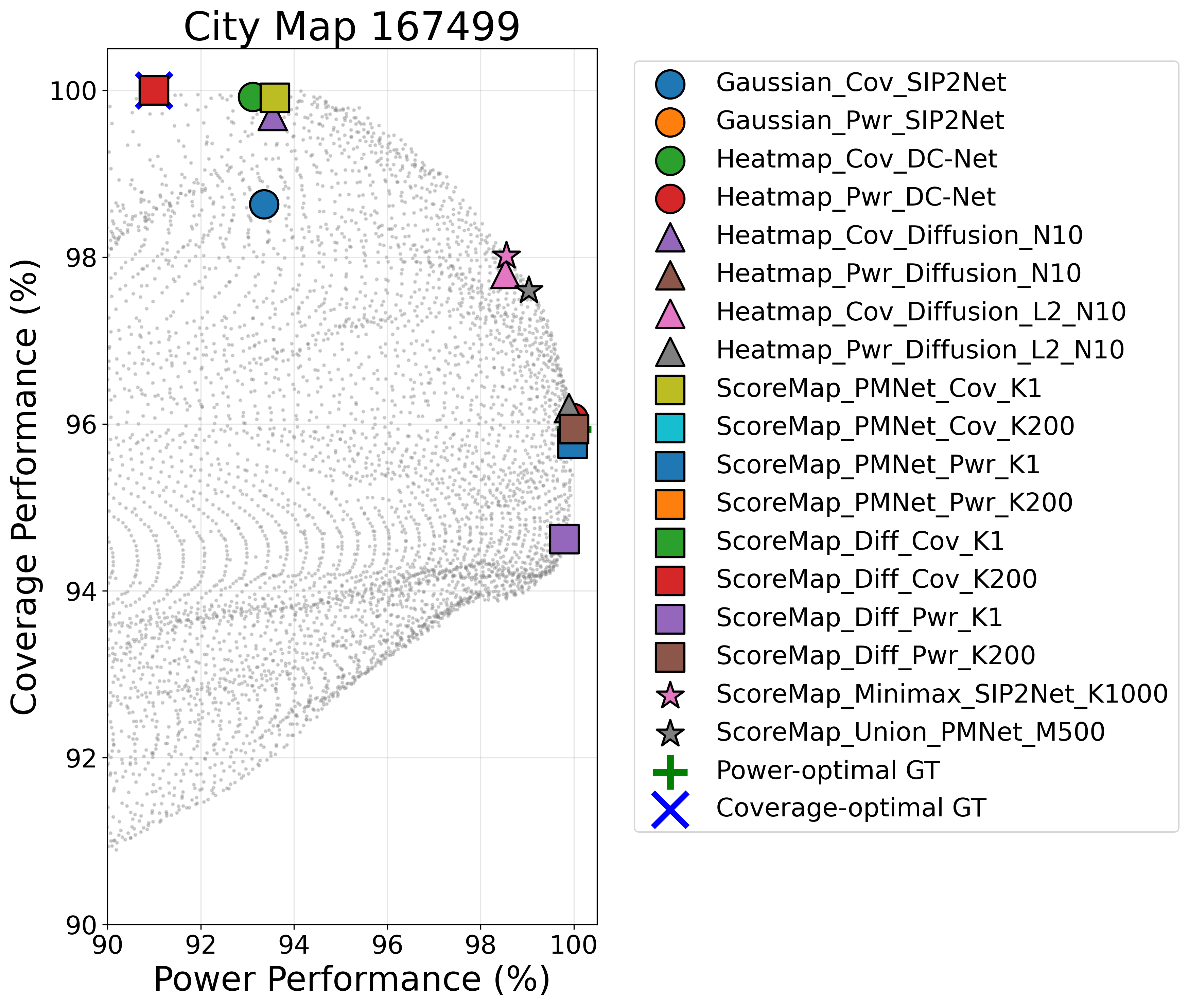}}
\caption{Zoomed version of Fig.~\ref{fig:pareto} (city map~167499), restricted to the $[90\%, 100\%]$ region on both axes. Structure visible at this scale reflects the local landscape near the optimal region; the asymmetry between coverage-optimal and power-optimal clusters is consistent with the dataset-level trade-off reported in Section~\ref{sec:gt_bounds}.}
\label{fig:paretoZoom}
\end{figure*}

\section{Computational Cost Analysis}
\label{sec:computational_cost}

We analyze inference time per \textcolor{black}{city map} for the different neural-network approaches to transmitter placement. All measurements use NVIDIA Quadro RTX 6000 GPUs in FP32 precision. Across the test set, the center 150$\times$150 region contains an average of 17,322 feasible transmitter positions per \textcolor{black}{city map}.

\subsection{Component-Level Timing}

The computational analysis is built around two core components.

\textbf{Model Forward Pass:} The discriminative models (DeepXL, DC-Net, SIP2Net, and PMNet) require roughly 30--40\,ms for a single forward pass. Diffusion models require about 450\,ms for single-sample generation. For multi-sample generation, we use batched diffusion sampling with batch size 16, which yields sublinear scaling with the number of samples.

\textbf{SAIPP-Net Evaluation:} SAIPP-Net evaluates candidate transmitter positions. We use batched evaluation with batch size 64 to improve GPU utilization. From the exhaustive-search baseline (54--72\,s for 17,322 evaluations), each batch requires 0.199--0.266\,s, corresponding to about 3.1--4.2\,ms per candidate when batched. This is approximately a $10\times$ speedup over unbatched processing, which takes about 30--40\,ms per candidate. The remaining gap to the theoretical $64\times$ limit is attributable to GPU transfer overhead and other fixed costs. For candidate-selection strategies with $K=1$ or $N=1$, no SAIPP-Net evaluation is needed, because there is only one candidate to select from.

\subsection{Exhaustive Benchmark Evaluation}

For each \textcolor{black}{city map}, benchmark ground truth is defined by exhaustively evaluating all feasible transmitter candidates under the fixed SAIPP-Net propagation model used in dataset generation. On the test set, this corresponds on average to 17,322 candidate positions, processed in 271 batched SAIPP-Net forward passes and requiring 54--72\,s per \textcolor{black}{city map}. This exhaustive procedure defines the benchmark optima and serves as the runtime reference for all neural approaches.

\subsection{Heatmap-Based Approaches}

\textbf{Discriminative Models (Single Prediction):} The discriminative heatmap models complete inference in 30--40\,ms per \textcolor{black}{city map} using only the model forward pass, with no SAIPP-Net evaluation required. This corresponds to about 1350--2400$\times$ speedup over exhaustive search and enables real-time deployment. The lower bound corresponds to the slowest discriminative pass (40\,ms) compared against the fastest exhaustive runtime (54\,s), and the upper bound corresponds to the fastest discriminative pass (30\,ms) compared against the slowest exhaustive runtime (72\,s).

\textbf{Diffusion Models:} Single-sample generation ($N{=}1$) achieves about 120--160$\times$ speedup with 450\,ms inference time. For multi-sampling, the diffusion model first generates $N$ candidate radio maps through batched denoising and then evaluates each candidate with SAIPP-Net. Using $N{=}10$ as a representative configuration, which is also the setting highlighted in Table~\ref{tab:computational_summary} and Figs.~\ref{fig:pareto}--\ref{fig:paretoZoom}, the runtime decomposes as follows:
\begin{itemize}
\item Batched diffusion sampling (one batch of 10 samples at batch size 16): about 3.2\,s.
\item SAIPP-Net evaluation (one batch of 10 candidates): about 0.2\,s.
\item Total: about 3.46\,s per \textcolor{black}{city map}.
\item Speedup: about 15.6--20.8$\times$ relative to exhaustive search.
\end{itemize}

Although the $N{=}10$ configuration evaluates only 10 candidates rather than all 17,322 exhaustive positions, the runtime is dominated by batched diffusion sampling rather than by SAIPP-Net evaluation. Larger pools such as $N{=}50$ or $N{=}100$ further improve both single-objective and balanced performance (see Tables~\ref{tab:diffusion_power}--\ref{tab:bestl2_multisample_diff}), but they do so at proportionally higher runtime cost.
\begin{table}[tbp]
\centering
\caption{Computational Cost and Performance Trade-offs. Representative configurations from each approach category, grouped by method family. Columns report the number of SAIPP-Net evaluations per \textcolor{black}{city map}, wall-clock time per \textcolor{black}{city map}, balanced performance $\bar{d}$, coverage and power percentages, and speedup relative to the exhaustive-search baseline. Rows marked with explicit $N$, $K$ or $M$ values correspond to the configurations highlighted in Tables~\ref{tab:diffusion_power}--\ref{tab:singleobj_union_avgmap} and in Figs.~\ref{fig:pareto}--\ref{fig:paretoZoom}.}
\label{tab:computational_summary}
\small
\resizebox{\textwidth}{!}{%
\begin{tabular}{@{}lccccccc@{}}
\toprule
Approach & Evals & Time (s) & $\bar{d}$ & Cov (\%) & Pwr (\%) & Speedup \\
\midrule
\multicolumn{7}{l}{\textit{Exhaustive Benchmark Baseline}} \\
Exhaustive $\bar{d}$ & 17,322 & 54--72 & 2.6 & 97.84 & 98.55 & 1.0$\times$ \\
Exhaustive Cov (\%) & 17,322 & 54--72 & 13.86 & 100 & 86.14 & 1.0$\times$ \\
Exhaustive Pwr (\%) & 17,322 & 54--72 & 5.5 & 94.5 & 100 & 1.0$\times$ \\
\midrule
\multicolumn{7}{l}{\textit{Heatmap-Based}} \\
Discriminative Cov (\%) (Table \ref{tab:fixed_coverage}) & 0 & \color{black} 0.03 & 13.11 & 97.65 & 87.10 &\color{black} 1800--2400$\times$ \\
Discriminative Pwr (\%) (Table \ref{tab:fixed_power}) & 0 &\color{black} 0.03 & 5.98 & 94.04 & 99.47 &\color{black} 1800--2400$\times$ \\
\color{black} Diffusion (N=10) Cov (\%) (Table \ref{tab:diffusion_coverage}) & 10 & 3.46 &\color{black} 12.2 &\color{black} 99.01 &\color{black} 87.84 & 16--21$\times$ \\
\color{black} Diffusion (N=1) Cov (\%)(Table \ref{tab:diffusion_coverage}) & 0 & 0.45 &\color{black} 14.30 &\color{black} 95.85 &\color{black} 86.32 & 120--160$\times$ \\
\color{black} Diffusion (N=10) Pwr (\%) (Table \ref{tab:diffusion_power})& 10 & 3.46 &\color{black} 5.73 &\color{black} 94.27 &\color{black} 99.86 & 16--21$\times$ \\
\color{black} Diffusion (N=1) Pwr (\%) (Table \ref{tab:diffusion_power})& 0 & 0.45 &\color{black} 6.21 &\color{black} 93.85 &\color{black} 99.12 & 120--160$\times$ \\
\color{black} Diffusion Cov (N=10) $\bar{d}$ (Table \ref{tab:bestl2_multisample_diff}) & 10 & 3.46 &\color{black} 5.03 &\color{black} 97.46 &\color{black} 95.65 & 16--21$\times$ \\
\color{black} Diffusion Pwr (N=10) $\bar{d}$ (Table \ref{tab:bestl2_multisample_diff}) & 10 & 3.46 &\color{black} 4.37 &\color{black} 95.68  &\color{black} 99.32 & 16--21$\times$ \\
\midrule
\multicolumn{7}{l}{\textit{Score-Map-Based}} \\
Discriminative Cov (\%) (Table \ref{tab:coverage_optimization}) & 0 &\color{black} 0.04 & 12.71 & 98.43 & 87.39 &\color{black} 1350--1800$\times$ \\
Discriminative Cov (\%) (K=200) (Table \ref{tab:coverage_optimization}) & 200 & 0.97 & 13.49 & 99.76 & 86.51 & 56--74$\times$ \\
\color{black} Diffusion Cov (\%) (Table \ref{tab:coverage_optimization}) & 0 & 0.45 &\color{black} 14.43 &\color{black} 97.75 &\color{black} 84.74 & 120--160$\times$ \\
\color{black} Diffusion Cov (\%) (K=200) (Table \ref{tab:coverage_optimization}) & 200 & 1.38 &\color{black} 13.29 &\color{black} 99.68 &\color{black} 86.71 & 39--52$\times$ \\
Discriminative Pwr (\%) (Table \ref{tab:power_optimization}) & 0 & \color{black} 0.04 & 5.94 & 94.07 & 99.63 &\color{black} 1350--1800$\times$ \\
Discriminative Pwr (\%) (K=200) (Table \ref{tab:power_optimization}) & 200 & 0.97 & 5.63 & 94.37 & 99.98 & 56--74$\times$ \\
\color{black} Diffusion Pwr (\%) (Table \ref{tab:power_optimization}) & 0 & 0.45 &\color{black} 6.39 &\color{black} 93.68 &\color{black} 99.07 & 120--160$\times$ \\
\color{black} Diffusion Pwr (\%) (K=200) (Table \ref{tab:power_optimization}) & 200 & 1.38 &\color{black} 5.67 &\color{black} 94.33 &\color{black} 99.95 & 39--52$\times$ \\
Discrim. $\bar{d}$ Minimax (K=1000) (Table \ref{tab:multiobjective_minimax_avgmap})& 1000 &\color{black} 3.2 & 2.69 & 97.73 & 98.55 &\color{black} 17--23$\times$ \\
\color{black} Diffusion $\bar{d}$ Minimax (K=1000) (Table \ref{tab:multiobjective_minimax_avgmap})& 1000 &\color{black} 3.6 &\color{black} 2.67 &\color{black} 97.77 &\color{black} 98.53 &\color{black} 15--20$\times$ \\
Discrim. $\bar{d}$ Union (M=500) (Table \ref{tab:singleobj_union_avgmap})& $\sim$888 & 3.28 & 2.62 & 97.77 & 98.63 & 17--22$\times$ \\
\color{black} Diffusion $\bar{d}$ Union (M=500) (Table \ref{tab:singleobj_union_avgmap})&\color{black} $\sim$901 &\color{black} 3.71 &\color{black} 2.65 &\color{black} 97.81 &\color{black} 98.57 & 14--18$\times$ \\
\bottomrule
\end{tabular}
}
\\[0.5em]
\footnotesize
Evals: number of SAIPP-Net evaluations; Time: seconds per \textcolor{black}{city map}; $\bar{d}$: $L_2$ distance from ideal point (representative values or range); Speedup: vs exhaustive search baseline. 
\end{table}

\subsection{Score-Map-Based Approaches}

Score-map approaches predict optimality scores over all positions and then evaluate only the top-ranked candidates with SAIPP-Net.

\textbf{Discriminative Dual Score-Maps ($K{=}1000$):} Using $K{=}1000$ as a representative setting, which offers a useful compromise between performance and cost, runtime decomposes as follows:
\begin{itemize}
\item Model forward pass for dual predictions: 30--40\,ms.
\item SAIPP-Net evaluation for 1000 candidates (16 batches): 3.2--4.3\,s.
\item Total: 3.2--4.3\,s per \textcolor{black}{city map}.
\item Speedup: 12.5--22.5$\times$ relative to exhaustive search.
\end{itemize}

This configuration evaluates 1000 selected candidates rather than all 17,322. The speedup comes mainly from batching efficiency: exhaustive search requires 271 SAIPP-Net batches, whereas $K{=}1000$ requires only 16. For smaller $K$, the speedup increases further; for example, $K{=}200$ yields 56--74$\times$ speedup with only four batches.

\textbf{Diffusion Dual Score-Maps ($K{=}1000$):} When diffusion is used for score-map generation, the corresponding representative runtime is:
\begin{itemize}
\item Diffusion forward pass (single sample, dual predictions): about 450\,ms.
\item SAIPP-Net evaluation for 1000 candidates (16 batches): 3.2--4.3\,s.
\item Total: 3.6--4.7\,s per \textcolor{black}{city map}.
\item Speedup: 11.5--20.0$\times$ relative to exhaustive search.
\end{itemize}

Similarly, $K{=}200$ achieves 39--52$\times$ speedup for diffusion score-map generation.

\textbf{Union Strategy:} The union approach pools the top-$M$ candidates from both score-maps. For $M{=}1000$, the resulting union contains on average about 1650--1850 candidates over the 16,753 test \textcolor{black}{city maps}, depending on the model. This requires proportionally more SAIPP-Net batches, but it also delivers the strongest balanced performance (Table~\ref{tab:singleobj_union_avgmap}). At $M{=}500$, the discriminative models achieve about 16.5--22$\times$ speedup, whereas the diffusion models achieve about 14--18$\times$ speedup.

\subsection{Performance-Speed Trade-offs}

\textcolor{black}{Table~\ref{tab:computational_summary} summarizes representative configurations from each approach family by reporting, for each method, the number of SAIPP-Net evaluations per city map, a single representative mean runtime, and the corresponding speedup factor against the exhaustive SAIPP-Net baseline (54--72\,s per city map, reported as a range because its runtime depends almost linearly on the number of free pixels per city map). The main pattern is clear: discriminative one-shot  models provide the fastest inference, with speedups of $\sim$$1350$--$2400\times$; diffusion heatmaps trade speed for inference-time search over alternative candidates, with speedups of $\sim$$16$--$160\times$; and score-map shortlist methods trade additional SAIPP-Net evaluation for especially strong balanced performance, with speedups of $\sim$$17$--$23\times$ at the dual-map configurations that approach the exhaustive balanced optimum. In particular, score-map approaches achieve near-optimal balanced performance ($\bar{d}\approx 2.62$--$2.69$) at roughly 3--4\,s per city map, whereas diffusion multi-sampling offers a more expensive route to exploring the objective-space trade-off frontier.}

\section{Conclusions}
\label{sec:conclusion}

We presented a comparative study of neural-network approaches for wireless transmitter placement together with RadioMapSeer-Deployment, a large-scale benchmark of 167,525 urban scenarios with dual ground-truth labels for coverage-optimal and power-optimal transmitter locations under a fixed learned propagation model (SAIPP-Net). This setting makes it possible to compare indirect heatmap-based and direct score-map-based formulations under shared data, shared architectures, and a shared evaluation protocol. 

The benchmark was made possible by learned radio map prediction at scale. Exhaustive per-location labeling with a ray tracer at comparable density would raise the per-\textcolor{black}{city map} cost from roughly one minute to several tens of hours, while exhaustive measurement-based supervision is out of reach in practice. At the present scale of approximately 1.67 billion SAIPP-Net inference operations, SAIPP-Net is therefore the practical mechanism that makes exhaustive benchmark labeling feasible. 

Ground-truth analysis reveals a clear asymmetry in the coverage--power trade-off on this benchmark: coverage-optimal placement sacrifices 13.86\% of received-power, whereas power-optimal placement sacrifices only 5.50\% of coverage. The best achievable balanced placement lies at $\bar{d}=2.60$ from the ideal point $(100\%,100\%)$, which provides a concrete reference for evaluating balanced deployment strategies. 

\textcolor{black}{On this benchmark, score-map approaches are the strongest formulation across all three evaluation surfaces --- power-only, coverage-only, and balanced --- in the present single-transmitter problem. Because they predict the objective landscape over feasible transmitter locations directly, dual-score-map candidate selection reaches near-optimal balanced performance, and Union-based selection in the best configurations attains the balanced benchmark optimum ($\bar{d}=2.60$).}

Heatmap-based approaches nevertheless remain attractive for two reasons. First, they predict physically meaningful intermediate radio maps rather than only a placement score surface; second, in the diffusion setting, multi-sample inference turns prediction into a small search procedure whose sample pool can be reused for both single-objective and balanced placement without explicit multi-objective training. 

From a computational perspective, discriminative one-shot models provide the fastest predictions, diffusion multi-sampling trades speed for inference-time search, and score-map methods deliver the strongest balance between runtime and balanced-placement quality when shortlist evaluation is allowed. The resulting picture is not that one formulation dominates universally, but that the two families occupy different regions of the accuracy--cost space. 

The methodological findings reported here should be interpreted as benchmark-level conclusions about how different learning formulations interact with the transmitter-placement problem under a fixed learned propagation model \textcolor{black}{(SAIPP-Net)}, rather than as direct claims about real-world deployment optimality. \textcolor{black}{Prior work cited in Section~\ref{sec:relatedwork} shows that learned propagation models trained on real measurements can match or outperform ray tracing in measurement-based evaluation~\cite{zhang_plnet,bakirtzis_deeplearn}; the same exhaustive-labeling methodology can therefore be deployed with a measurement-trained successor to SAIPP-Net to simultaneously reduce the synthetic-supervision bias and improve sim-to-real fidelity, through training on real measurements collected in different urban environments and through higher-fidelity scene information such as geometry, electromagnetic material properties, and antenna characteristics. These suggest that the learned-model-based exhaustive-labeling framework used here is not merely a computational convenience, but a practically relevant route.}

Two future directions stand out. A near-term direction is improved transmitter recovery for the heatmap formulation, for example by replacing pixel-wise argmax with a lightweight localization module that maps predicted radio maps, possibly together with the city map, to transmitter coordinates. \textcolor{black}{A broader direction is extension to multi-transmitter deployment. There, the present direct (score-map) formulation does not scale naturally: the natural multi-transmitter analog of the score-map is a tensor of joint-placement scores with approximately $150^{2n}$ entries per environment at the feasible-region size used here ($\sim$$5\times10^8$ for $n=2$, $\sim$$10^{13}$ for $n=3$), which is not a natural CNN output and is also impractical to store at dataset scale. Heatmap-based formulations, by contrast, predict a 2D radio map regardless of the number of transmitters and so remain naturally applicable; the inference-time multi-sampling mechanism of the diffusion model --- useful in the single-transmitter case studied here as a runtime/accuracy trade-off via best-of-$N$ candidate selection --- is potentially of greater value in the multi-transmitter setting.}

\bibliographystyle{IEEEtran}
\bibliography{references}

@article{taus2026optimal,
  title   = {Optimal Transmitter Placement in Realistic Urban Environments},
  author  = {Taus, Lukas and Tsai, Richard and Andrews, Jeffrey G.},
  journal = {arXiv preprint arXiv:2604.28153},
  year    = {2026}
}

@article{jaensch_indoor_robust,
  author={Fabian Jaensch and {\c{C}}a{\u{g}}kan Yapar and Giuseppe Caire and Begüm Demir},
  title        = {Radio Map Prediction from Noisy Environment Information and Sparse Observations},
  journal      = {arXiv preprint arXiv:2602.11950},
  year         = {2026},
  eprint       = {2602.11950},
  archivePrefix= {arXiv}
}

@ARTICLE{feng_rmd3qn_2026,
  author={Feng, Bin and Zheng, Meng and Liang, Wei and Zhang, Lei and Chen, Lin},
  journal={IEEE Wireless Communications Letters}, 
  title={Radio Map-Assisted Transmitter Deployment in Complex Indoor Scenarios via Deep Reinforcement Learning}, 
  year={2026},
  volume={15},
  number={},
  pages={2508-2512},
  doi={10.1109/LWC.2026.3682084}}

@ARTICLE{REMUNet,
  author={Sallouha, Hazem and Sarkar, Shamik and Krijestorac, Enes and Cabric, Danijela},
  journal={IEEE Open Journal of Signal Processing}, 
  title={{REM-U-Net}: Deep Learning Based Agile REM Prediction With Energy-Efficient Cell-Free Use Case}, 
  year={2024},
  volume={5},
  number={},
  pages={750-765},
  doi={10.1109/OJSP.2024.3378591}}

@INPROCEEDINGS{Kehai2023GC,
  author={Qiu, Kehai and Bakirtzis, Stefanos and Song, Hui and Wassell, Ian and Zhang, Jie},
  booktitle={ICASSP 2023 - 2023 IEEE International Conference on Acoustics, Speech and Signal Processing (ICASSP)}, 
  title={Deep Learning-Based Path Loss Prediction for Outdoor Wireless Communication Systems}, 
  year={2023},
  volume={},
  number={},
  pages={1-2},
  doi={10.1109/ICASSP49357.2023.10095501}}

@INPROCEEDINGS{2,
  author={Xing, Tianxiang and Zou, Leyi and Bharadwaj, Tejas and Balaji, Rushabha and Cabric, Danijela},
  booktitle={2025 IEEE 35th International Workshop on Machine Learning for Signal Processing (MLSP)}, 
  title={{U-Net} Based Indoor Radio Map Prediction under Sparse Sampling},
address={Istanbul, Turkiye},
  year={2025}
}

@INPROCEEDINGS{3,
  author={Zheng, Zhihao  and Xiao, Limin and Zhao, Ming and Li, Yunzhou},
  booktitle={2025 IEEE 35th International Workshop on Machine Learning for Signal Processing (MLSP)}, 
  title={Efficient Indoor Radio Map Prediction with Improved {T}ransformers and Active Sampling Strategies},
address={Istanbul, Turkiye},
  year={2025}
}

@INPROCEEDINGS{4,
  author={Chen, Qi and Tan, Haidong and Yang, Jingjing Huang,Ming and Chen,Boyuan},
  booktitle={2025 IEEE 35th International Workshop on Machine Learning for Signal Processing (MLSP)}, 
  title={{IRM-Net}: An Enhanced Attention Networks For Indoor Radio Map
Estimation},
address={Istanbul, Turkiye},
  year={2025}
}

@INPROCEEDINGS{5,
  author={Petrosyan, Khoren and Khachatrian, Hrant and Mkrtchyan, Rafayel and Raptis, Theofanis},
  booktitle={2025 IEEE 35th International Workshop on Machine Learning for Signal Processing (MLSP)}, 
  title={{U-Net} for Indoor Pathloss Prediction from Sparse Measurements with Physics-Informed Features},
address={Istanbul, Turkiye},
  year={2025}
}

@INPROCEEDINGS{6,
  author={Wu, Mengfan and Skocaj, Marco and Boban, Mate},
  booktitle={2025 IEEE 35th International Workshop on Machine Learning for Signal Processing (MLSP)}, 
  title={Radio Map Prediction via Neural Networks with Ground Truth Shortcuts and Selective Sampling},
address={Istanbul, Turkiye},
  year={2025}
}

@INPROCEEDINGS{7,
  author={Kojima, Ryoichi and Ito, Satoshi and Nagao,Tatsuya and Taya, Masato},
  booktitle={2025 IEEE 35th International Workshop on Machine Learning for Signal Processing (MLSP)}, 
  title={Sparse-guided {RadioUNet} With Adaptive Sampling For The {MLSP} 2025 Sampling-assisted Pathloss Radio Map Prediction Data Competition},
address={Istanbul, Turkiye},
  year={2025}
}

@INPROCEEDINGS{IPP-NET,  author={Feng, Bin and Zhen, Meng and Liang, Wei and Zhang, Lei},  booktitle={Proc. IEEE Int. Conf. on Acoustics, Speech and Signal Processing (ICASSP)},   title={ {IPP-NET}: {A} GENERALIZABLE DEEP NEURAL NETWORK MODEL FOR INDOOR
PATHLOSS RADIO MAP PREDICTION},   year={2025}, month={April}, volume={},  number={}}

@INPROCEEDINGS{Split_U_Net,  author={Wu, Mengfan and Skocaj, Marco and Boban, Mate},  booktitle={Proc. IEEE Int. Conf. Acoust., Speech Signal Process. (ICASSP)},   title={ Enhancing Convolutional Models for Indoor Radio Mapping via Ray Marching},   year={2025}, month={April}, volume={},  number={}}

@INPROCEEDINGS{TransPathNet,  author={Li, Xin  and Liu, Ran  and Xu, Saihua and Gulam Razul, Sirajudeen and Yuen, Chau },  booktitle={Proc. IEEE Int. Conf. Acoust., Speech Signal Process. (ICASSP)},   title={  {TransPathNet}: A Novel Two-Stage Framework for Indoor Radio Map Prediction},   year={2025},month={April},  volume={},  number={}}

@INPROCEEDINGS{ResUnet,  author={Cisse, Cheick Tidiani and Baala, Oumaya and Guillet, Valery and Spies, Francois and Caminada, Alexandre},  booktitle={Proc. IEEE Int. Conf. Acoust., Speech Signal Process. (ICASSP)},   title={Generalizable Indoor Path Loss Prediction},   year={2025},  volume={}, month={April}, number={}}

@Article{mkrtchyan_vit,
  author  = {Mkrtchyan, Rafayel and Ghukasyan, Edvard and Petrosyan, Khoren and Khachatrian, Hrant and Raptis, Theofanis P.},
  title   = {Vision Transformers for Efficient Indoor Pathloss Radio Map Prediction},
  journal = {Electronics},
  volume  = {14},
  number  = {10},
  pages   = {1905},
  year    = {2025},
  doi     = {10.3390/electronics14101905}
}

@article{li_u6g,
  author  = {Li, Xiaojie and Han, Yu and Lu, Zhizheng and Jin, Shi and Wen, Chao-Kai},
  title   = {{U6G XL-MIMO} Radiomap Prediction: Multi-Config Dataset and Beam Map Approach},
  journal = {arXiv preprint arXiv:2603.06401},
  year    = {2026}
}

@article{manukyan_rt_limits,
  author  = {Manukyan, Armen and Khachatrian, Hrant and Ghukasyan, Edvard and Raptis, Theofanis P.},
  title   = {On the Limitations of Ray-Tracing for Learning-Based {RF} Tasks in Urban Environments},
  journal = {arXiv preprint arXiv:2507.19653},
  year    = {2025}
}

@article{manukyan_sim2real,
  author  = {Manukyan, Armen and Mkrtchyan, Rafayel and Saribekyan, Ararat and Raptis, Theofanis P. and Khachatrian, Hrant},
  title   = {Bridging the sim-to-real gap in {RF} localization with large-scale synthetic pretraining},
  journal = {Information Fusion},
  year    = {2025},
  doi     = {10.1016/j.inffus.2025.104104}
}

@article{khachatrian_nlos,
  author  = {Khachatrian, Hrant and Mkrtchyan, Rafayel and Raptis, Theofanis P.},
  title   = {Deep learning with synthetic data for wireless {NLOS} positioning with a single base station},
  journal = {Ad Hoc Networks},
  volume  = {167},
  pages   = {103696},
  year    = {2025},
  doi     = {10.1016/j.adhoc.2024.103696}
}

@article{IndoorDataSetAlt,
	title = {Indoor Radio Map Dataset},
 author = {Bakirtzis , Stefanos and Yapar, {\c{C}}a{\u{g}}kan and Qui, Kehai and Wassell, Ian and Zhang, Jie},
 url = {https://dx.doi.org/10.21227/c0ec-cw74},
	howpublished = "\url{https://dx.doi.org/10.21227/c0ec-cw74}",
 journal = {IEEE Dataport},
	year = {2024}, 
	note = "\url{https://dx.doi.org/10.21227/c0ec-cw74}"
}

@article{radiomapseerdeployment1TxAlt,
	title = {Dataset of Optimal Radio Maps and Deployments},
 author = {Yapar, {\c{C}}a{\u{g}}kan},
 url = {https://dx.doi.org/10.21227/wjwa-th03},
 journal = {IEEE Dataport},
	year = {2026}
}

@inproceedings{ddim,
  author    = {Jiaming Song and Chenlin Meng and Stefano Ermon},
  title     = {Denoising Diffusion Implicit Models},
  booktitle = {9th International Conference on Learning Representations, {ICLR} 2021},
  year      = {2021},
  url       = {https://arxiv.org/abs/2010.02502}
}

@book{goldsmith2005wireless,
  title={Wireless communications},
  author={Goldsmith, Andrea},
  year={2005},
  publisher={Cambridge university press}
}

@ARTICLE{liang_diffnet,
  author={Liang, Ruihuai and Yang, Bo and Chen, Pengyu and Li, Xianjin and Xue, Yifan and Yu, Zhiwen and Cao, Xuelin and Zhang, Yan and Debbah, Mérouane and Vincent Poor, H. and Yuen, Chau},
  journal={IEEE Internet of Things Journal}, 
  title={Diffusion Models as Network Optimizers: Explorations and Analysis}, 
  year={2025},
  volume={12},
  number={10},
  pages={13183-13193},
  doi={10.1109/JIOT.2025.3528955}}

@ARTICLE{diffsg,
  author={Liang, Ruihuai and Yang, Bo and Yu, Zhiwen and Guo, Bin and Cao, Xuelin and Debbah, Mérouane and Poor, H. Vincent and Yuen, Chau},
  journal={IEEE Communications Magazine}, 
  title={{DiffSG}: A Generative Solver for Network Optimization with Diffusion Model}, 
  year={2025},
  volume={63},
  number={6},
  pages={16-24},
  doi={10.1109/MCOM.001.2400428}}

@ARTICLE{gdplan,
  author={Kan, Nuowen and Yan, Sa and Zou, Junni and Dai, Wenrui and Gao, Xing and Li, Chenglin and Xiong, Hongkai},
  journal={IEEE Transactions on Networking}, 
  title={{GDPlan}: Generative Network Planning via Graph Diffusion Model}, 
  year={2025},
  volume={33},
  number={4},
  pages={1422-1437},
  doi={10.1109/TON.2025.3535518}}

@misc{wang_ris3d,
      title={Optimization for Massive {3D-RIS} Deployment: A Generative Diffusion Model-Based Approach}, 
      author={Kaining Wang and Bo Yang and Zhiwen Yu and Xuelin Cao and Mérouane Debbah and Chau Yuen},
      year={2025},
      eprint={2509.11969},
      archivePrefix={arXiv},
      primaryClass={cs.NI},
      url={https://arxiv.org/abs/2509.11969}, 
}

@ARTICLE{gdsg,
  author={Liang, Ruihuai and Yang, Bo and Chen, Pengyu and Cao, Xuelin and Yu, Zhiwen and Debbah, Mérouane and Niyato, Dusit and Poor, H. Vincent and Yuen, Chau},
  journal={IEEE Transactions on Mobile Computing}, 
  title={{GDSG}: Graph Diffusion-Based Solution Generator for Optimization Problems in {MEC} Networks}, 
  year={2025},
  volume={24},
  number={10},
  pages={10264-10277},
  doi={10.1109/TMC.2025.3568248}}

@INPROCEEDINGS{gdmra,
  author={Uslu, Yiğit Berkay and Hadou, Samar and Bidokhti, Shirin Saeedi and Ribeiro, Alejandro},
  booktitle={2025 IEEE 10th International Workshop on Computational Advances in Multi-Sensor Adaptive Processing (CAMSAP)}, 
  title={Generative Diffusion Models for Resource Allocation in Wireless Networks}, 
  year={2025},
  volume={},
  number={},
  pages={201-205},
  doi={10.1109/CAMSAP66162.2025.11423961}}

@ARTICLE{bakirtzis_iwn,
  author={Bakirtzis, Stefanos and Wassell, Ian and Fiore, Marco and Zhang, Jie},
  journal={IEEE Network}, 
  title={{AI}-Assisted Indoor Wireless Network Planning With Data-Driven Propagation Models}, 
  year={2024},
  volume={38},
  number={6},
  pages={451-458},
  doi={10.1109/MNET.2024.3397801}}

@INPROCEEDINGS{sionna_rt,
  author={Hoydis, Jakob and Aoudia, Faycal Ait and Cammerer, Sebastian and Nimier-David, Merlin and Binder, Nikolaus and Marcus, Guillermo and Keller, Alexander},
  booktitle={2023 IEEE Globecom Workshops (GC Wkshps)}, 
  title={Sionna {RT}: Differentiable Ray Tracing for Radio Propagation Modeling}, 
  year={2023},
  volume={},
  number={},
  pages={317-321},
  doi={10.1109/GCWkshps58843.2023.10465179}}

@ARTICLE{fadenet,
  author={Ratnam, Vishnu V. and Chen, Hao and Pawar, Sameer and Zhang, Bingwen and Zhang, Charlie Jianzhong and Kim, Young-Jin and Lee, Soonyoung and Cho, Minsung and Yoon, Sung-Rok},
  journal={IEEE Access}, 
  title={{FadeNet}: Deep Learning-Based mm-Wave Large-Scale Channel Fading Prediction and its Applications}, 
  year={2021},
  volume={9},
  number={},
  pages={3278-3290},
  doi={10.1109/ACCESS.2020.3048583}}

@ARTICLE{emdeepray,
  author={Bakirtzis, Stefanos and Chen, Jiming and Qiu, Kehai and Zhang, Jie and Wassell, Ian},
  journal={IEEE Transactions on Antennas and Propagation}, 
  title={{EM DeepRay}: An Expedient, Generalizable, and Realistic Data-Driven Indoor Propagation Model}, 
  year={2022},
  volume={70},
  number={6},
  pages={4140-4154},
  doi={10.1109/TAP.2022.3172221}}

@misc{bakirtzis_deeplearn,
      title={Radio Propagation Modelling: To Differentiate or To Deep Learn, That Is The Question}, 
      author={Stefanos Bakirtzis and Paul Almasan and José Suárez-Varela and Gabriel O. Ferreira and Michail Kalntis and André Felipe Zanella and Ian Wassell and Andra Lutu},
      year={2025},
      eprint={2509.19337},
      archivePrefix={arXiv},
      primaryClass={cs.NI},
      url={https://arxiv.org/abs/2509.19337}, 
}

@ARTICLE{challengeoverview,
  author={Yapar, {\c{C}}a{\u{g}}kan and Jaensch, Fabian and Levie, Ron and Kutyniok, Gitta and Caire, Giuseppe},
  journal={IEEE Open Journal of Signal Processing}, 
  title={Overview of the First Pathloss Radio Map Prediction Challenge}, 
  year={2024},
  volume={5},
  number={},
  pages={948-963},
  doi={10.1109/OJSP.2024.3419563}}

@ARTICLE{romeroSurvey,
  author={Romero, Daniel and Kim, Seung-Jun},
  journal={IEEE Signal Processing Magazine}, 
  title={Radio Map Estimation: A data-driven approach to spectrum cartography}, 
  year={2022},
  volume={39},
  number={6},
  pages={53-72},
  doi={10.1109/MSP.2022.3200175}}

@ARTICLE{ZengSurvey,
  author={Zeng, Yong and Chen, Junting and Xu, Jie and Wu, Di and Xu, Xiaoli and Jin, Shi and Gao, Xiqi and Gesbert, David and Cui, Shuguang and Zhang, Rui},
  journal={IEEE Communications Surveys \&  Tutorials}, 
  title={A Tutorial on Environment-Aware Communications via Channel Knowledge Map for {6G}}, 
  year={2024},
  volume={26},
  number={3},
  pages={1478-1519},
  doi={10.1109/COMST.2024.3364508}}

@misc{wangtutorial,
      title={A Tutorial on Learning-Based Radio Map Construction: Data, Paradigms, and Physics-Awarenes}, 
      author={Xiucheng Wang and Yuhao Pan and Nan Cheng},
      year={2026},
      eprint={2603.17499},
      archivePrefix={arXiv},
      primaryClass={eess.SY},
      url={https://arxiv.org/abs/2603.17499}, 
}

@Article{mdpireview,
AUTHOR = {Feng, Bin and Zheng, Meng and Liang, Wei and Zhang, Lei},
TITLE = {A Recent Survey on Radio Map Estimation Methods for Wireless Networks},
JOURNAL = {Electronics},
VOLUME = {14},
YEAR = {2025},
NUMBER = {8},
ARTICLE-NUMBER = {1564},
URL = {https://www.mdpi.com/2079-9292/14/8/1564},
ISSN = {2079-9292},
DOI = {10.3390/electronics14081564}
}

@ARTICLE{empowering,
  author={Bakirtzis, Stefanos and Yapar, {\c{C}}a{\u{g}}kan and Fiore, Marco and Zhang, Jie and Wassell, Ian},
  journal={IEEE Wireless Communications}, 
  title={Empowering Wireless Network Applications with Deep Learning-Based Radio Propagation Models}, 
  year={2025},
  volume={32},
  number={4},
  pages={124-131},
  doi={10.1109/MWC.012.2400336}}

@inproceedings{zhang_plnet,
author = {Zhang, Xin and Shu, Xiujun and Zhang, Bingwen and Ren, Jie and Zhou, Lizhou and Chen, Xin},
title = {Cellular Network Radio Propagation Modeling with Deep Convolutional Neural Networks},
year = {2020},
isbn = {9781450379984},
publisher = {Association for Computing Machinery},
address = {New York, NY, USA},

booktitle = {Proceedings of the 26th ACM SIGKDD International Conference on Knowledge Discovery \& Data Mining},
pages = {2378–2386},
numpages = {9},
location = {Virtual Event, CA, USA},
series = {KDD '20}
}

@ARTICLE{jaensch_beam,
  author={Jaensch, Fabian and Caire, Giuseppe and Demir, Begüm},
  journal={IEEE Wireless Communications Letters}, 
  title={Beam Index Map Prediction in Unseen Environments From Geospatial Data}, 
  year={2026},
  volume={15},
  number={},
  pages={1519-1523},
  doi={10.1109/LWC.2026.3657504}}

@ARTICLE{jaensch_aerial,
  author={Jaensch, Fabian and Caire, Giuseppe and Demir, Begüm},
  journal={IEEE Transactions on Wireless Communications}, 
  title={Radio Map Prediction From Aerial Images and Application to Coverage Optimization}, 
  year={2026},
  volume={25},
  number={},
  pages={308-320},
  doi={10.1109/TWC.2025.3583171}}

@ARTICLE{seretis_unet,
  author={Seretis, Aristeidis and Xu, Charley and Sarris, Costas},
  journal={IEEE Transactions on Antennas and Propagation}, 
  title={Fast Selection of Indoor Wireless Transmitter Locations With Generalizable Neural Network Propagation Models}, 
  year={2024},
  volume={72},
  number={10},
  pages={7927-7940},
  doi={10.1109/TAP.2024.3439826}}

@article{isabona_nsga2,
author = {Isabona, Joseph and Imoize, Agbotiname Lucky and Ojo, Stephen and Venkatareddy, Prashanth and Hinga, Simon Karanja and Sánchez-Chero, Manuel and Ancca, Sheda Méndez},
title = {Accurate Base Station Placement in {4G} {LTE} Networks Using Multiobjective Genetic Algorithm Optimization},
journal = {Wireless Communications and Mobile Computing},
volume = {2023},
number = {1},
pages = {7476736},
doi = {https://doi.org/10.1155/2023/7476736},
year = {2023}
}

@ARTICLE{mallik_gandqn,
  author={Mallik, Mohammed and Villemaud, Guillaume},
  journal={IEEE Access}, 
  title={{EMF} Aware Reinforcement Learning for Base Station Deployment Using Conditional {GAN}s}, 
  year={2026},
  volume={14},
  number={},
  pages={3806-3820},
  doi={10.1109/ACCESS.2025.3648583}}

@article{cisse_cgan,
  title={Deep Learning-Driven Combinatorial Optimization for Indoor Radio Planning},
  author={Ciss{\'e}, Cheick Tidiani and Baala, Oumaya and Guillet, Val{\'e}ry and Spies, Fran{\c{c}}ois and Caminada, Alexandre},
  journal={Available at SSRN: https://ssrn.com/abstract=4976575},
 year={2024}
}

@INPROCEEDINGS{chen_hookjeeves,
  author={Chen, Dantong and Yang, Songjiang and Wang, Yinghua and Huang, Jie and Wang, Cheng-Xiang and Zhu, Qiuming},
  booktitle={2023 IEEE/CIC International Conference on Communications in China (ICCC)}, 
  title={Ray-Tracing Based Large Indoor Office Base Station Deployment Optimization at Millimeter Wave Bands}, 
  year={2023},
  volume={},
  number={},
  pages={1-6},
  doi={10.1109/ICCC57788.2023.10233518}}

@misc{autoBS,
      title={{AutoBS}: Autonomous Base Station Deployment with Reinforcement Learning and Digital Network Twins}, 
      author={Ju-Hyung Lee and Andreas F. Molisch},
      year={2025},
      eprint={2502.19647},
      archivePrefix={arXiv},
      primaryClass={cs.IT},
      url={https://arxiv.org/abs/2502.19647}, 
}

@misc{autoplan,
      title={Automatic Network Planning with Digital Radio Twin}, 
      author={Xiaomeng Li and Yuru Zhang and Qiang Liu and Mehmet Can Vuran and Nathan Huynh and Li Zhao and Mizan Rahman and Eren Erman Ozguven},
      year={2026},
      eprint={2509.12441},
      archivePrefix={arXiv},
      primaryClass={cs.NI},
      url={https://arxiv.org/abs/2509.12441}, 
}

@misc{yuan_diffusion,
      title={Planning with Language and Generative Models: Toward General Reward-Guided Wireless Network Design}, 
      author={Chenyang Yuan and Xiaoyuan Cheng},
      year={2026},
      eprint={2602.00357},
      archivePrefix={arXiv},
      primaryClass={cs.LG},
      url={https://arxiv.org/abs/2602.00357}, 
}

@article{li_daqga,
title = {Optimization of {5G} base station deployment based on quantum genetic algorithm in outdoor {3D} map},
journal = {Computer Networks},
volume = {269},
pages = {111431},
year = {2025},
issn = {1389-1286},
doi = {https://doi.org/10.1016/j.comnet.2025.111431},
author = {Jianpo Li and Jinjian Pang and Binfeng Jiang and Qi Xu and Enyuan Zhang}
}

@ARTICLE{palizban_mmwave,
  author={Palizban, Nima and Szyszkowicz, Sebastian and Yanikomeroglu, Halim},
  journal={IEEE Wireless Communications Letters}, 
  title={Automation of Millimeter Wave Network Planning for Outdoor Coverage in Dense Urban Areas Using Wall-Mounted Base Stations}, 
  year={2017},
  volume={6},
  number={2},
  pages={206-209},
  doi={10.1109/LWC.2017.2659732}}

@inproceedings{stosic_wlan,
author = {Stosic, Jovan and Hadzi-Velkov, Zoran and Gavrilovska, Liljana},
year = {2005},
booktitle={International Symposium on Wireless Personal Multimedia Communications (IWS / WPMC)}, 
month = {01},
pages = {},
title = {Planning of Large-Scale {WLAN} Infrastructures}
}

@INPROCEEDINGS{ossn,
  author={Zheng, Yi and Liao, Cunyi and Wang, Ji and Liu, Shouyin},
  booktitle={2024 IEEE International Conference on Acoustics, Speech, and Signal Processing Workshops (ICASSPW)}, 
  title={A Transformer-Based Network for Unifying Radio Map Estimation and Optimized Site Selection}, 
  year={2024},
  volume={},
  number={},
  pages={610-614},
  doi={10.1109/ICASSPW62465.2024.10627516}}

@INPROCEEDINGS{romero_aerial,
  author={Romero, Daniel and Viet, Pham Q. and Leus, Geert},
  booktitle={ICASSP 2022 - 2022 IEEE International Conference on Acoustics, Speech and Signal Processing (ICASSP)}, 
  title={Aerial Base Station Placement Leveraging Radio Tomographic Maps}, 
  year={2022},
  volume={},
  number={},
  pages={5358-5362},
  doi={10.1109/ICASSP43922.2022.9746987}}

@INPROCEEDINGS{WinPropFEKO, 
	author={R. {Hoppe} and G. {W{\"o}lfle} and U. {Jakobus}}, 
	booktitle={Proc. Int. Appl. Computational Electromagnetics Society Symp. - Italy (ACES)}, 
	title={Wave propagation and radio network planning software {W}in{P}rop added to the electromagnetic solver package {FEKO}}, 
	year={2017}, 
	volume={}, 
	number={}, 
	address = {Florence, Italy},
	doi={10.23919/ROPACES.2017.7916282}, 
	ISSN={}, 
	month={March},}

@inproceedings{DPM,
  title={Dominant path prediction model for urban scenarios},
  author={Wahl, Ren{\'e} and W{\"o}lfle, Gerd and Wertz, Philipp and Wildbolz, Pascal and Landstorfer, Friedrich},
  booktitle={14th IST Mobile and Wireless Communications Summit},
  pages={1--5},
  year={2005}
}

@inproceedings{IRT,
	title={Fast {3-D} Ray Tracing for the Planning of Microcells by Intelligent Preprocessing of the Data Base},
	author={R. Hoppe and G. W{\"o}lfle and F. Landstorfer},
	booktitle={3rd European Personal and Mobile Communications Conference (EPMCC)},
pages={149--154},
	year={1999}
}

@ARTICLE{radioUNet,  author={R. {Levie} and Yapar, {\c{C}}a{\u{g}}kan and G. {Kutyniok} and G. {Caire}},  journal={IEEE Transactions on Wireless Communications},   title={Radio{UN}et: Fast Radio Map Estimation With Convolutional Neural Networks},   year={2021},  volume={20},  number={6},  pages={4001-4015},  doi={10.1109/TWC.2021.3054977}}

@INPROCEEDINGS{FirstIndoorChallenge,
  author={Bakirtzis, Stefanos and Yapar, {\c{C}}a{\u{g}}kan and Qiu, Kehai and Wassell, Ian and Zhang, Jie},
  booktitle={ICASSP 2025 - 2025 IEEE International Conference on Acoustics, Speech and Signal Processing (ICASSP)}, 
  title={The First Indoor Pathloss Radio Map Prediction Challenge}, 
  year={2025},
  volume={},
  number={},
  pages={1-2},
  doi={10.1109/ICASSP49660.2025.10889381}}

@INPROCEEDINGS{FirstChallenge,
  author={Yapar, {\c{C}}a{\u{g}}kan and Jaensch, Fabian and Levie, Ron and Kutyniok, Gitta and Caire, Giuseppe},
  booktitle={ICASSP 2023 - 2023 IEEE International Conference on Acoustics, Speech and Signal Processing (ICASSP)}, 
  title={The First Pathloss Radio Map Prediction Challenge}, 
  year={2023},
  volume={},
  number={},
  pages={1-2},
  doi={10.1109/ICASSP49357.2023.10433928}}

@InProceedings{ACNet,
author = {Ding, Xiaohan and Guo, Yuchen and Ding, Guiguang and Han, Jungong},
title = {{ACNet}: Strengthening the Kernel Skeletons for Powerful {CNN} via Asymmetric Convolution Blocks},
booktitle = {Proceedings of the IEEE/CVF International Conference on Computer Vision (ICCV)},
month = {October},
year = {2019}
}

@InProceedings{ASPP,
author = {Chen, Liang-Chieh and Zhu, Yukun and Papandreou, George and Schroff, Florian and Adam, Hartwig},
title = {Encoder-Decoder with Atrous Separable Convolution for Semantic Image Segmentation},
booktitle = {Proceedings of the European Conference on Computer Vision (ECCV)},
month = {September},
year = {2018}
}

@ARTICLE{AOT,
  author={Zeng, Yanhong and Fu, Jianlong and Chao, Hongyang and Guo, Baining},
  journal={IEEE Transactions on Visualization and Computer Graphics}, 
  title={Aggregated Contextual Transformations for High-Resolution Image Inpainting}, 
  year={2023},
  volume={29},
  number={7},
  pages={3266-3280},
  doi={10.1109/TVCG.2022.3156949}}

@InProceedings{He_2016_CVPR,
author = {He, Kaiming and Zhang, Xiangyu and Ren, Shaoqing and Sun, Jian},
title = {Deep Residual Learning for Image Recognition},
booktitle = {Proceedings of the IEEE Conference on Computer Vision and Pattern Recognition (CVPR)},
month = {June},
year = {2016}
}

@inproceedings{ho2020denoising,
  title={Denoising diffusion probabilistic models},
  author={Ho, Jonathan and Jain, Ajay and Abbeel, Pieter},
  booktitle={Advances in Neural Information Processing Systems},
  volume={33},
  pages={6840--6851},
  year={2020}
}

@INPROCEEDINGS{DA-cGAN,
  author={Liu, Chun-Hao and Chang, Hun and Park, Taesuh},
  booktitle={2020 IEEE/CVF Conference on Computer Vision and Pattern Recognition Workshops (CVPRW)}, 
  title={{DA-cGAN}: A Framework for Indoor Radio Design Using a Dimension-Aware Conditional Generative Adversarial Network}, 
  year={2020},
  volume={},
  number={},
  pages={2089-2098}}

@INPROCEEDINGS{SIP2Net,
  author={Lu, Wenlihan and Lu, Ziyi and Yan, Jia and Gao, Shijian},
  booktitle={ICASSP 2025 - 2025 IEEE International Conference on Acoustics, Speech and Signal Processing (ICASSP)}, 
  title={{SIP2Net}: Situational-Aware Indoor Pathloss-Map Prediction Network for Radio Map Generation}, 
  year={2025},
  volume={},
  number={},
  pages={1-2}}

@INPROCEEDINGS{PMNet,
  author={Lee, Ju-Hyung and Serbetci, Omer Gokalp and Selvam, Dheeraj Panneer and Molisch, Andreas F.},
  booktitle={GLOBECOM 2023 - 2023 IEEE Global Communications Conference}, 
  title={{PMNet}: Robust Pathloss Map Prediction via Supervised Learning}, 
  year={2023},
  volume={},
  number={},
  pages={4601-4606}}

@article{DCNet,
title = {{DC-Net}: A Distant-range Content Interaction Network for Radio Map construction},
journal = {ICT Express},
volume = {10},
number = {5},
pages = {1145-1150},
year = {2024},
issn = {2405-9595},
doi = {https://doi.org/10.1016/j.icte.2024.08.008},
url = {https://www.sciencedirect.com/science/article/pii/S2405959524001024},
author = {Qi Chen and Ming Huang and JingJing Yang}
}

@InProceedings{UNet,
author="Ronneberger, Olaf
and Fischer, Philipp
and Brox, Thomas",
editor="Navab, Nassir
and Hornegger, Joachim
and Wells, William M.
and Frangi, Alejandro F.",
title="U-Net: Convolutional Networks for Biomedical Image Segmentation",
booktitle="Medical Image Computing and Computer-Assisted Intervention -- MICCAI 2015",
year="2015",
publisher="Springer International Publishing",
address="Cham",
pages="234--241"
}

@book{miettinen1999nonlinear,
  title={Nonlinear multiobjective optimization},
  author={Miettinen, Kaisa},
  volume={12},
  year={1999},
  publisher={Springer Science \& Business Media}
}

@INPROCEEDINGS{avgmap_baseline_paper,
  author={He, Jingyi and Zheng, Yi},
  booktitle={2025 IEEE/CIC International Conference on Communications in China (ICCC Workshops)}, 
  title={{BSCSM-GAN}: End-to-End Generative Prediction of Base Station Coverage Strength Maps for 6{G}-Aware Urban Deployment}, 
  year={2025},
  volume={},
  number={},
  pages={1-5},
  doi={10.1109/ICCCWorkshops67136.2025.11148193}}

@article{radiomapseer,
url = {https://arxiv.org/abs/2212.11777},
journal={arXiv preprint:2212.11777},
author = {Yapar, {\c{C}}a{\u{g}}kan and Levie, Ron and Kutyniok, Gitta and Caire, Giuseppe},
title = {Dataset of Pathloss and {ToA} Radio Maps With Localization Application},
publisher = {arXiv},
year = {2022}
}

@INPROCEEDINGS{radionet,
  author={Feng, Bin and Zheng, Meng and Liang, Wei and Zhang, Lei},
  booktitle={2025 IEEE 35th International Workshop on Machine Learning for Signal Processing (MLSP)}, 
  title={{SAIPP-Net}: A Sampling-Assisted Indoor Pathloss Prediction Method for Wireless Communication Systems}, 
  year={2025}}

@INPROCEEDINGS{sapra,
  author={Yapar, {\c{C}}a{\u{g}}kan and Bakirtzis, Stefanos and Lutu, Andra and Wassell, Ian and Zhang, Jie and Caire, Giuseppe},
  booktitle={2025 IEEE 35th International Workshop on Machine Learning for Signal Processing (MLSP)}, 
  title={The Sampling-Assisted Pathloss Radio Map Prediction Competition}, 
  year={2025},
  volume={},
  number={},
  pages={1-6},
  doi={10.1109/MLSP62443.2025.11204278}}

@misc{OpenStreetMap,
	author = {{OpenStreetMap contributors}},
	title = {{Planet dump retrieved from https://planet.osm.org}},
	howpublished = "\url{https://www.openstreetmap.org}",
	year = {2017},
}

@article{ssim,
  title={Image quality assessment: from error visibility to structural similarity},
  author={Wang, Zhou and Bovik, Alan C and Sheikh, Hamid R and Simoncelli, Eero P},
  journal={IEEE Transactions on Image Processing},
  volume={13},
  number={4},
  pages={600--612},
  year={2004},
  publisher={IEEE}
}

@article{ms-ssim,
  title={Multiscale structural similarity for image quality assessment},
  author={Wang, Zhou and Simoncelli, Eero P and Bovik, Alan C},
  journal={The Thirty-Seventh Asilomar Conference on Signals, Systems \& Computers, 2003},
  volume={2},
  pages={1398--1402},
  year={2003},
  organization={IEEE}
}

@article{focal_loss,
  title={Focal loss for dense object detection},
  author={Lin, Tsung-Yi and Goyal, Priya and Girshick, Ross and He, Kaiming and Doll{\'a}r, Piotr},
  journal={Proceedings of the IEEE International Conference on Computer Vision},
  pages={2980--2988},
  year={2017}
}

@article{tv_loss,
  title={Nonlinear total variation based noise removal algorithms},
  author={Rudin, Leonid I and Osher, Stanley and Fatemi, Emad},
  journal={Physica D: Nonlinear Phenomena},
  volume={60},
  number={1-4},
  pages={259--268},
  year={1992},
  publisher={Elsevier}
}

@inproceedings{gdl,
  title={Video frame synthesis using deep voxel flow},
  author={Liu, Ziwei and Yeh, Raymond A and Tang, Xiaoou and Liu, Yiming and Agarwala, Aseem},
  booktitle={Proceedings of the IEEE International Conference on Computer Vision},
  pages={4463--4471},
  year={2017}
}

\end{document}